\documentclass{article} 
\usepackage[final]{colm2026_conference}

\usepackage{tcolorbox}
\usepackage{array}
\usepackage{subcaption}
\usepackage{float}
\usepackage{adjustbox}
\usepackage{longtable}
\usepackage{multirow}
\usepackage{enumitem}
\usepackage{tikz}
\usepackage{xcolor}
\usepackage{amsmath}
\usepackage{ragged2e}

\definecolor{inpBlue}{RGB}{52,101,164}   
\definecolor{outGreen}{RGB}{38,162,105}  
\definecolor{accOrange}{RGB}{230,97,1}   
\definecolor{softGray}{RGB}{240,240,240} 

\definecolor{checkGreenDark}{RGB}{32,92,69}
\definecolor{checkGreenMid}{RGB}{66,137,105}
\definecolor{checkGreenLight}{RGB}{238,247,242}
\definecolor{checkGreenLine}{RGB}{165,202,184}

\newcolumntype{L}[1]{>{\RaggedRight\arraybackslash}p{#1}}

\usepackage{enumitem}
\tcbuselibrary{breakable, skins}

\definecolor{promptbg}{RGB}{245, 247, 252}
\definecolor{promptframe}{RGB}{180, 190, 210}
\definecolor{headerbg}{RGB}{70, 90, 140}
\definecolor{originalbg}{RGB}{252, 250, 240}
\definecolor{originalframe}{RGB}{210, 195, 150}
\definecolor{surveybg}{RGB}{255, 251, 240}
\definecolor{surveyframe}{RGB}{200, 175, 120}
\definecolor{surveyheader}{RGB}{140, 110, 50}
\definecolor{promptbg}{RGB}{240, 245, 255}
\definecolor{promptframe}{RGB}{150, 170, 210}
\definecolor{promptheader}{RGB}{55, 80, 140}

\usepackage{microtype}
\usepackage{hyperref}
\usepackage{url}
\usepackage{booktabs}

\usepackage{lineno}

\definecolor{darkblue}{rgb}{0, 0, 0.5}
\hypersetup{colorlinks=true, citecolor=darkblue, linkcolor=darkblue, urlcolor=darkblue}

\newtheorem{definition}{Definition}
\newcommand{\pmstd}[2]{#1\,{\footnotesize$\pm$\,#2}} 

\title{From Feelings to Metrics:\\Understanding and Formalizing How Users \textsc{Vibe-Test} LLMs}

\author{
Itay Itzhak \\
Technion -- Israel Institute of Technology \\
The Hebrew University of Jerusalem \\
\texttt{itay1itzhak@gmail.com} \\
\And
Eliya Habba \\
The Hebrew University of Jerusalem \\
\texttt{eliya.habba@mail.huji.ac.il}
\And
Gabriel Stanovsky \\
The Hebrew University of Jerusalem \hspace{1.7cm} \\
\texttt{gabriel.stanovsky@mail.huji.ac.il} \\
\And
Yonatan Belinkov \\
Technion -- Israel Institute of Technology \\
\texttt{belinkov@technion.ac.il} \\
}

\begin{document}
\maketitle

\begin{abstract}
Evaluating LLMs is challenging, as benchmark scores often fail to capture models' real-world usefulness.
Instead, users often rely on \mbox{``vibe-testing'':} informal experience-based evaluation, such as comparing models on coding tasks related to their own workflow.
While prevalent, vibe-testing is often too ad hoc and unstructured to analyze or reproduce at scale.
In this work, we study how vibe-testing works in practice and then formalize it to support systematic analysis.
We first analyze two empirical resources: (1) a survey of user evaluation practices, and (2) a collection of in-the-wild model comparison reports from blogs and social media.
Based on these resources, we formalize vibe-testing as a two-part process: users personalize both \textit{what} they test and \textit{how} they judge responses. 
We then introduce a proof-of-concept evaluation pipeline that follows this formulation by generating personalized prompts and comparing model outputs using user-aware subjective criteria.
In experiments on single-turn coding benchmarks, we find that combining personalized prompts and user-aware evaluation can change which model is preferred, reflecting the role of vibe-testing in practice. 
These findings suggest that formalized vibe-testing can serve as a useful approach for bridging benchmark scores and real-world experience.\footnote{See code and study artifacts at: \url{https://technion-cs-nlp.github.io/vibe-testing-llms}.} 
\end{abstract}

\section{Introduction}

Evaluating LLMs has been a long-standing challenge in NLP research~\citep{laskar-etal-2024-systematic,cao2025toward}, as popular evaluation suites typically report performance as aggregated scores on standardized tasks~\citep{zhang2024helm,ICLR2025_94074dd5}.
However, these scores often miss the usefulness of models in real-world workflows~\citep{kiela-etal-2021-dynabench,mazumder2023dataperf,openaiSycophancyGPT4o}.
In practice, model usefulness often depends on context-dependent criteria, such as clarity, ease of use, or workflow fit~\citep{weidinger2025toward,saad2024lmunit}.
As a result, strong benchmark scores do not necessarily imply a good fit for users' needs in everyday tasks.

\begin{figure*}[t!]
    \centering
    \includegraphics[width=0.98\linewidth,trim=0.0cm 0cm 0.9cm 0cm,clip]{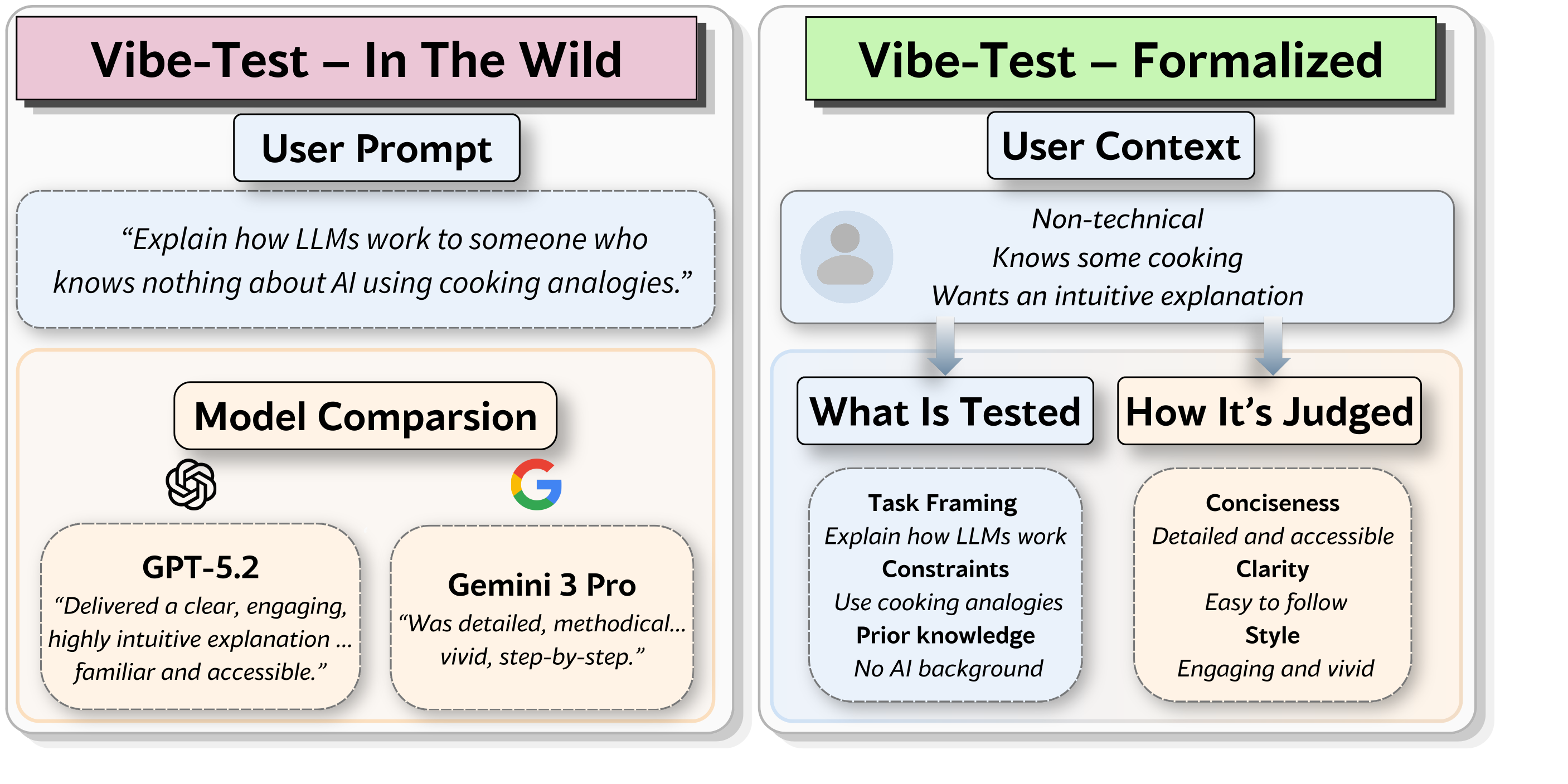}
    \caption{\textbf{Anatomy of a ``vibe-test''.} In practice, users evaluate LLMs by ``vibe-testing'' them: -- writing personalized prompts that test specific behaviors and judging models' responses using personal subjective criteria. We analyze recurring patterns of vibe-testing in real-world user comparisons, formalize them into a two-part structure, and present a proof-of-concept pipeline for automated vibe-testing. Example taken from \href{https://www.tomsguide.com/ai/i-tested-chatgpt-5-2-vs-gemini-3-0-with-7-real-world-prompts-heres-the-winner}{Tom's Guide}.}
    \label{fig:vibe_case_study}
\end{figure*}

In response, many users turn to ``\emph{vibe-testing}'': an informal practice of evaluating models through targeted experiments or extended personal use~\citep{mediumEvaluatingLarge,huggingfaceIntroducingSheets}.
Instead of relying solely on benchmark scores, users compare models on tasks that resemble their own workflows and judge the responses qualitatively.
These comparisons often focus on practical aspects of model behavior, such as writing style, clarity of explanations, ease of use, or how well the output fits a specific workflow (Figure~\ref{fig:vibe_case_study}).
Vibe-testing can be performed by individual users or shared by community members on blogs, forums, and social media.



Users' reliance on vibe-testing implies that it captures valuable aspects of model performance that benchmarks often miss.
However, vibe-testing is inherently informal and subjective -- different users test different tasks and judge responses from their own perspective.
As a result, insights from these evaluations remain scattered and fragmented, making them difficult to compare or transfer across settings. 
This informality leaves a gap between the practical insights of vibe-testing and our ability to study them systematically.
Recent work has suggested assessing model ``vibe'' and personalized evaluation, but none have empirically studied vibe testing itself as a user practice or suggested an evaluation framework grounded in it (Section~\ref{sec:background}).

In this work, we empirically study the practice of vibe-testing, formalize its distinctive patterns, and propose a proof-of-concept evaluation pipeline for systematic analysis. 
We begin by examining vibe testing in practice, drawing on two empirical sources.
First, we conduct a survey on evaluation practices, asking questions such as \textit{``What do you look for when testing a model?''}.
Second, we collect an ``in-the-wild'' corpus of model comparisons from blogs, forums, tech articles, and YouTube reviews.
We annotate these examples to identify recurring patterns in users' test design and response evaluation.
Together, these sources provide real-world evidence of how users vibe-test models: what they test and what they look for when judging responses.

Building on these empirical sources, we formalize vibe-testing as an evaluation practice defined by two recurring types of dimensions.
\emph{Input dimensions} capture what users test and how they construct prompts, while \emph{Output dimensions} capture how users judge model responses.
For example, in coding assistance, input dimensions can include the type of coding task or the amount of context provided (e.g., debugging a codebase).
Output dimensions can include clarity, adherence to constraints, and fit to the user’s workflow (e.g., production-ready code).
This formalization makes vibe-testing easier to compare and analyze, and provides a basis for systematic reproduction.

We leverage this formalization and introduce a proof-of-concept evaluation pipeline that mirrors the two-part structure of vibe-testing.
Given a brief user description, the pipeline first rewrites benchmark prompts to reflect that user's likely context and preferences.
It then compares models head-to-head by judging their responses along user-relevant output dimensions from that user's perspective.
We apply the pipeline to single-turn coding benchmarks and find that personalizing both the prompt and the judgment criteria can change which model is preferred.
In several head-to-head comparisons, the preferred model flips relative to the original benchmark prompts, while non-personal rewrites largely preserve the original ordering.
These results echo the core idea behind vibe-testing: model preferences can change when both the task framing and response judgment are tailored to the user.

Overall, this work takes a first step toward turning vibe-testing from an informal practice into a structured form of user-centered evaluation.

\begin{figure*}[t!]
    \centering
    \subfloat{\includegraphics[width=0.49\linewidth]{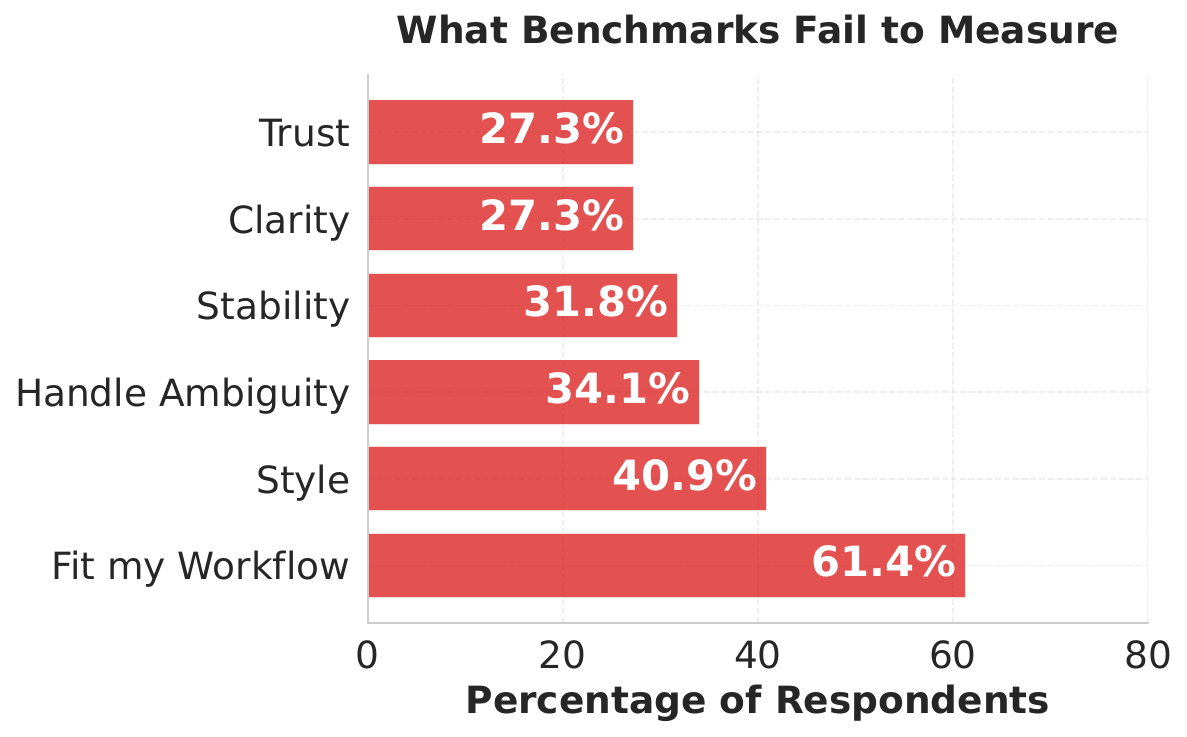}}\label{fig:survey_benchamrk_fails}
    \subfloat{\includegraphics[width=0.49\linewidth]{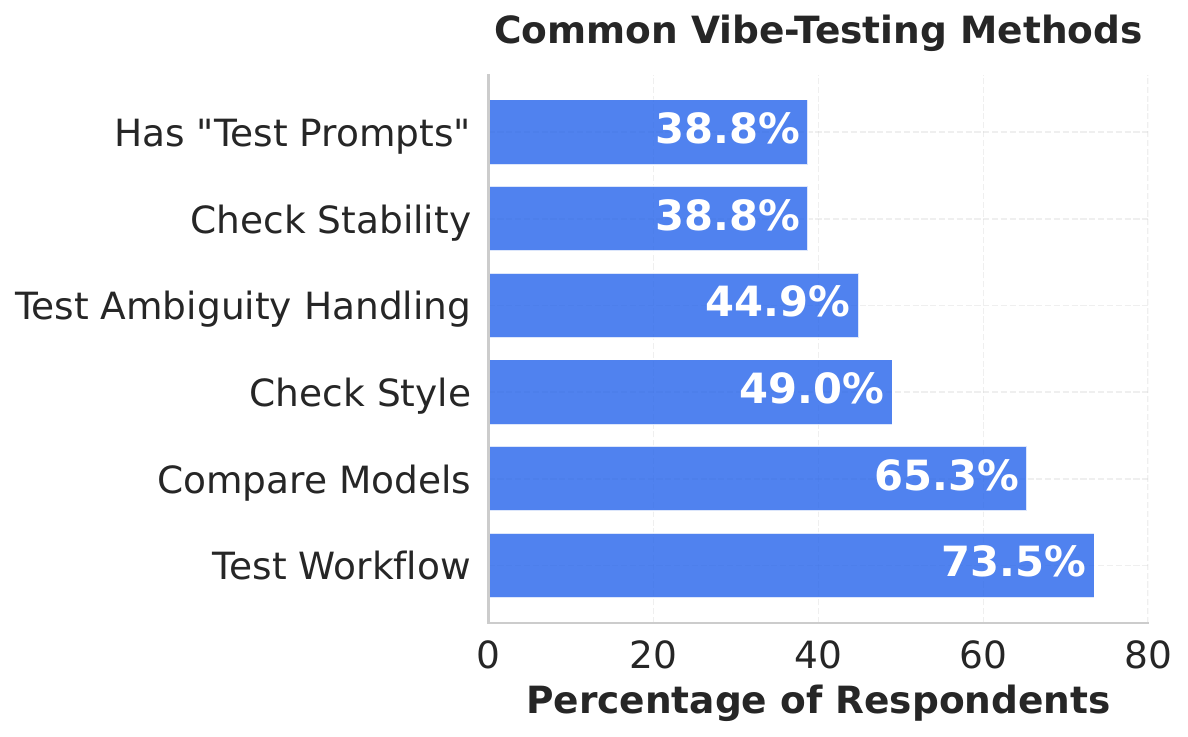}\label{fig:survey_common_vibe_testing_methods}}
    \caption{\textbf{Benchmarks vs.\ vibe-testing in practice.} \textbf{Left: What benchmarks miss.} Survey participants selected real-world qualities that benchmarks fail to capture (multi-select), including workflow and style fit, handling ambiguity, stability, clarity, and trust. \textbf{Right: How users test models.} Common strategies include trying tasks from one’s own workflow, side-by-side comparisons, probing style, stress-testing ambiguity handling, checking repeated runs stability, and using recurring ``test prompts,''. Notably, the qualities participants say benchmarks miss largely match what they explicitly probe during vibe-testing.}\label{fig:survey_results}
\end{figure*}

\section{Background: personal vibe evaluation}\label{sec:background}

Recent work has shown that standard benchmarks often miss aspects of model performance that matter to users in real-world use, motivating more qualitative user-centered evaluation~\citep{cao2025toward,weidinger2025toward}.
However, systematically capturing vibe-testing evaluation requires a framework that is subjective, personalized, and grounded in empirical evidence.
To our knowledge, prior work typically captures only one of these properties.


\paragraph{Vibe-based subjective evaluation.}

A few recent papers study ``vibe''-like aspects of model behavior, but they do so in different ways.
Most relevant is \textsc{VibeCheck}~\citep{dunlap2024vibecheck}, which measures qualitative ``vibe'' differences between models, but at the population level rather than for individual users. 
\textsc{Vibe Checker}~\citep{zhong2025vibe} adds verifiable instruction checks to coding tasks, but covers only automatically checkable traits and leaves out softer subjective dimensions, limiting potential coverage.
Beyond these, HELM Instruct~\citep{zhang2024helm} offers general stylistic evaluation while ChatBench~\citep{chang2025chatbench} focuses on evaluating interactive conversations, both relying on predefined criteria.
\paragraph{User-focused evaluation.}
EvalLM~\citep{kim2024evallm} supports manual rubric customization but does not automate or infer personalization. IQA-Eval~\citep{li2024iqa} adapts evaluation to user personas by simulating interactive correction and questioning, but focuses on the writing style of factual questions. \textsc{EvalAgent}~\citep{wadhwa2025evalagent} mines expert-authored guidance to uncover implicit evaluation criteria, but targets prompt underspecification rather than personalizing the user's input. Complementary work uses LLM-based user simulators for interactive evaluation in task-oriented dialogue~\citep{Luo2024DuetSimBUA,Jia2024SimulBenchELA}, but does not address subjective model comparison across user profiles.
Our work addresses the qualitative user-centered evaluation gap in a different way than previous work. We aim to capture vibe-testing evaluation by empirically examining how users compare models in the real world, formalizing this process, and proposing a modular pipeline that can be extended with other methods.
\section{What is vibe-testing?}\label{sec:understanding_vibe_testing}


To study vibe-testing systematically, we first ask what it looks like in practice.
To answer this, we collect and analyze two complementary empirical resources: a survey of user evaluation practices and an analysis of in-the-wild comparison reports.
Together, these sources provide a clearer and more concrete picture of how users evaluate models in practice.



\subsection{User survey}\label{subsec:user_survey}

We conduct a survey to understand the prevalence of vibe-testing, how users carry it out in practice, what they think benchmarks miss, what vibe-testing helps them assess instead, and whether it's worth automating (full survey results are in Appendix~\ref{appendix:full_survey}).

\paragraph{Demographics.}
We recruit $51$ volunteers via social media platforms (e.g., X/Reddit), including both AI/ML experts ($47\%$) and broader technical practitioners ($47\%$), as well as non-technical users ($6\%$). Respondents reported using AI tools daily ($92\%$) or weekly ($8\%$).

\paragraph{Prevalence of vibe-testing.}
We ask respondents whether they have vibe-tested models, loosely describing vibe-testing as \textit{``Evaluating an AI model through direct interaction, using your own prompts or tasks to judge how the model performs in practice.''}
Most respondents reported ``Yes'' ($82\%$) and that they often experiment with models (mean=$5.31$ on a scale of 1--7).

\paragraph{How users vibe-test models.}
We ask respondents how they test models and how they judge models' outputs.
The most common testing methods were trying tasks from one’s own workflow and comparing models' outputs side by side (Figure~\ref{fig:survey_results}, right).
When asked which criteria they use to judge outputs, the most frequently selected were correctness ($92\%$), clarity ($59\%$), and workflow fit ($41\%$).
These responses suggest that vibe-testing is typically grounded in personal workflows and judged using both correctness and practical, user-relevant criteria.

\paragraph{The benchmark-experience gap.}

We then ask respondents whether they had ever encountered a model that ``felt'' significantly different from what its benchmark scores would suggest.
Most answered ``Yes'' ($86\%$), indicating that many perceive a mismatch between benchmark rankings and real-world experience.
When asked what benchmarks fail to measure, workflow and style fit were the most common selections (Figure~\ref{fig:survey_results}, left).
Notably, these reported gaps align closely with the aspects respondents say they evaluate when vibe-testing, as illustrated side by side in Figure~\ref {fig:survey_results}.
Finally, most respondents ($83\%$) expressed interest in tools that could make the vibe-testing process more structured or automated.


The survey findings suggest that vibe-testing is common among technical users and is built around personal workflow tasks, side-by-side comparisons, and subjective judgments of output. 
They also point to a perceived gap between benchmarks and real-world experience, and to the value that vibe-testing adds.
We next complement these findings by analyzing public in-the-wild model comparison reports to examine recurring patterns in real-world vibe-testing.

\subsection{Analyzing vibe-testing in the wild}

We next turn to ``in the wild'' model-comparison reports to examine how vibe-testing appears in practice.
Unlike benchmark results, these comparisons are typically shared informally across social media, blogs, and community forums.

We semi-automatically construct a carefully curated corpus of $40$ public model comparison reports in four stages:\footnote{Additional details on the corpus are in Appendix~\ref{appendix:full_in_the_wild_analysis}.}

(1) \textbf{Source collection.}
We manually search for public reports that contain concrete, qualitative comparisons of LLMs, drawing from dozens of YouTube reviews, Reddit threads, blog posts, and news articles (source list is in Appendix~\ref{appendix:full_in_the_wild_analysis}). 
We include sources that (i) reference specific models, (ii) describe at least one concrete test input (prompt, task, or scenario), and (iii) include qualitative judgments and subjective claims.
The resulting corpus is a selected collection of naturally occurring vibe-testing examples.

(2) \textbf{Vibe-test instances extraction.}
For each comparison report, we use LLMs\footnote{GPT-5.2 and Gemini 3 Pro were chosen for this task via ``vibe-testing''. Prompts are in Appendix~\ref{appendix:full_in_the_wild_analysis}.} to extract and label \emph{vibe-test instances}, defined as localized cases where a report author evaluates one or more models on a specific input using some qualitative criteria.
Specifically, the LLMs were prompted to return short quoted spans or paraphrased snippets corresponding to vibe-tests, along with the tested task and the stated criteria.
We manually verified extracted \emph{vibe-test instances}, removing false positives and correcting errors. 

(3) \textbf{Attribute annotation.}
We annotate each \emph{vibe-test instance} with a small set of structured attributes, including task type, models compared, and the subjective criteria mentioned in the source (e.g., ``Answer Clarity'').
We perform this annotation with LLM assistance and manual review and refinement to ensure consistency and faithfulness to the original text.

(4) \textbf{Consolidating dimensions.}
To characterize recurring patterns, we ask the LLMs to propose lists of repeated subjective dimensions appearing across \emph{vibe-test instances} by grouping similar criteria under shared labels (e.g., ``Answer Clarity'' and ``Clear Output'' under ``Clarity'').
In parallel, we independently compiled our own lists from manual reviews.
We then iteratively reconcile and refine these lists, using both our judgments and LLM suggestions, to derive a final set of recurring dimensions.
Using this fixed set, we re-annotate \emph{vibe-test instances}, establishing the final dimension labels for the corpus.


The iterative procedure in Stage (4) yields a consolidated list of recurring dimensions spanning both the \emph{input}, what users choose to test and how they frame it, and the \emph{output}, what qualities they attend to when interpreting responses.
Together with our survey results, this list provides the basis for the formalization of vibe-testing presented next.

\section{Formalizing vibe-testing}\label{sec:define_vibe_testing}

Based on our empirical findings in Section~\ref{sec:understanding_vibe_testing}, we now formalize a definition of vibe-testing.

\begin{definition}[Vibe Testing]
Vibe-testing is an interaction-based LLM evaluation practice, in which evaluators adapt both the \textbf{input} they test (via input dimensions) and the \textbf{criteria} used to judge output (via output dimensions). It is intended to capture aspects of practical utility and user experience that standard benchmarks may miss. 
\end{definition}

\paragraph{Dimensions of Vibe-Testing.}
To describe vibe-testing systematically, we introduce vibe dimensions: recurring aspects of what users test and how they judge outputs. Our survey and corpus suggest two broad groups of such dimensions.\footnote{Tables~\ref{tab:vibe_taxonomy_input} and~\ref{tab:vibe_taxonomy_output} in the Appendix list the definitions and illustrative cues for each dimension.}
\begin{itemize}
    \item \textbf{Input-oriented dimensions} consist of \emph{task type}, \emph{task complexity/scope}, \emph{real-world context setting}, \emph{persona-based framing}, \emph{underspecification level}, \emph{constraint tightness}, and \emph{reference material availability}.
    \item \textbf{Output-oriented dimensions} consist of \emph{comparison setup}, \emph{correctness/accuracy}, \emph{clarity and structure}, \emph{cognitive load}, \emph{style/tone fit}, \emph{workflow fit}, \emph{friction/loss of control}, \emph{ambiguity handling}, \emph{reliability/stability}, \emph{trustworthiness/safety behavior}, and \emph{anthropomorphism}.
\end{itemize}

We illustrate these dimensions with the vibe-test example from Figure~\ref{fig:vibe_case_study}.
On the input side, the prompt reflects choices about \emph{real-world context setting} (cooking)  and \emph{persona-based framing} (a non-technical audience), and it sets a moderate \emph{task complexity/scope} (a conceptual explanation rather than a single fact).
An evaluator can further specialize the same test with \emph{constraint tightness} (e.g., ``use exactly three analogies'') or by increasing \emph{underspecification level} (e.g., leaving the target audience implicit).

On the \textbf{output side}, the example setting shows a common \emph{comparison setup} in which the same prompt is run on two models and the responses are read side by side. 
The response judgment reflects multiple dimensions: 
The mention of \textit{``clear...highly intuitive''} points to the \emph{clarity and structure} dimension, while \textit{``detailed, methodical''} indicates their preferred \emph{tone/style} fit, and \textit{``familiar and accessible''} refers to the evaluator's own \emph{workflow fit}.
Different evaluators can run the same prompt with similar judgment criteria in mind, but interpret those criteria differently or assign them different importance, leading to different preferences even when both responses are correct. 
In the next Section, we present an evaluation pipeline that reflects this formulation of vibe-testing -- adapting both the input and the evaluation based on user preferences.

\section{Automating vibe-testing}\label{sec:auto_vibe_testing}

\begin{figure*}[t!]
\centering
  \includegraphics[width=0.99\linewidth,trim=0.2cm 3.7cm 0.2cm 0cm,clip]{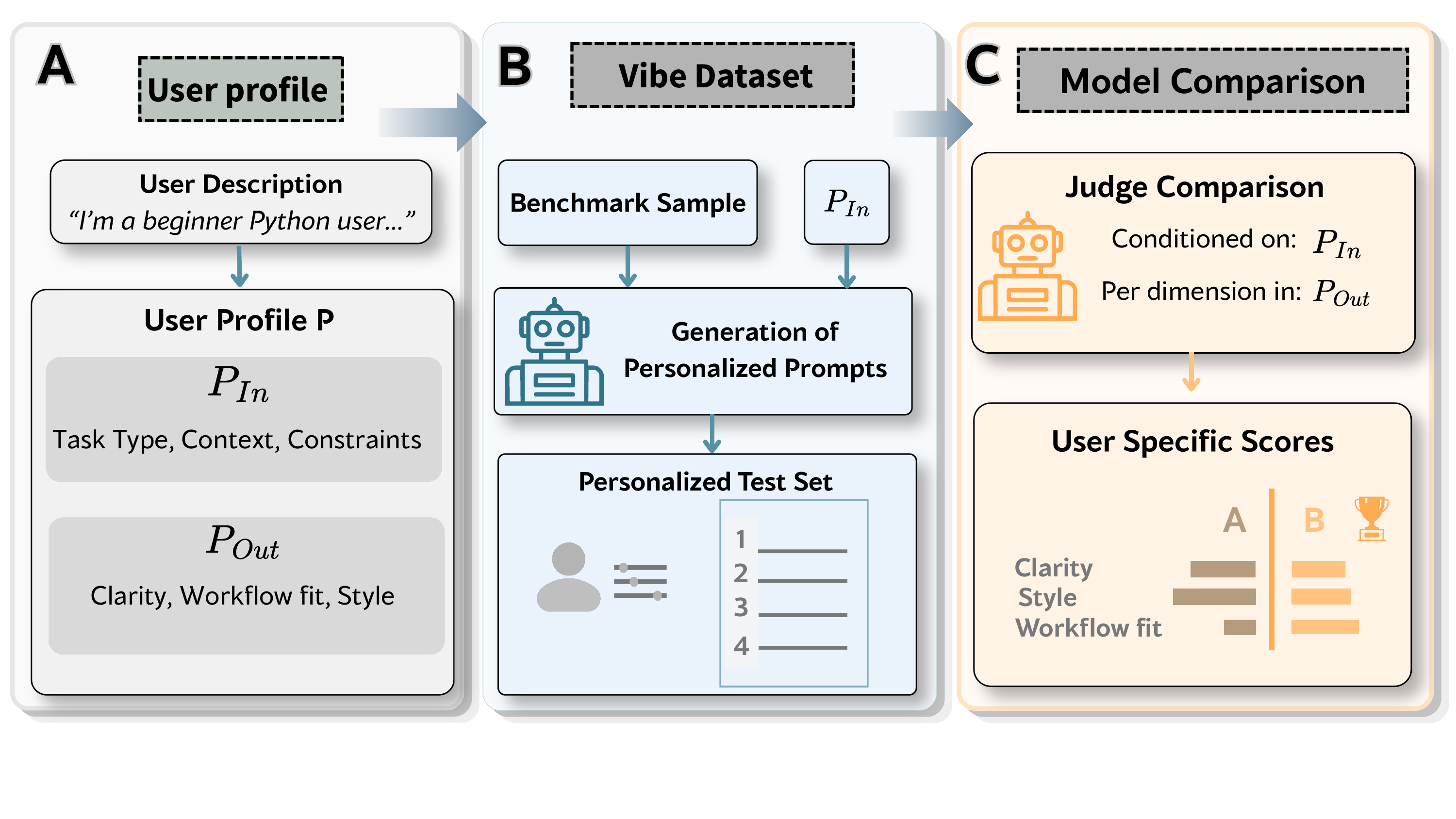}
  \caption{\textbf{Automatic Vibe-Testing Pipeline}: Given a user description, the pipeline \textbf{(A)} constructs a user profile $P$ (composed of input $P_{\text{in}}$ and output $P_{\text{out}}$) preferences, \textbf{(B)} rewrites benchmark samples into a personalized prompt aligned with $P_{\text{in}}$, and \textbf{(C)} compares responses using $P_{\text{out}}$ to produce per-dimension head-to-head model comparisons.}\label{fig:vibe-automation-pipeline} 
\end{figure*}

We now instantiate the formulation above as a modular proof-of-concept pipeline (Figure~\ref{fig:vibe-automation-pipeline}).
Its goal is to test whether such a two-part evaluation method can reveal meaningful preference shifts that benchmarks miss.
We study it in a focused setting using single-turn coding tasks and pairwise comparisons as a concrete testbed.
The resulting pipeline is modular: its stages can be implemented in different ways and expanded upon, using existing or new methods for user profiling, input personalization, and subjective evaluation.




\subsection{Pipeline description}

Given a brief user description, the pipeline first builds a structured profile of the user's input and output preferences.
It then uses that profile to rewrite personalized benchmark prompts and compare candidate models head-to-head from the same user perspective.
The result is a set of user-conditioned comparisons that lets us quantify how model preference changes across users and prompt variants.

\paragraph{(A) User profiling.}  
We begin by converting a natural language user description (e.g., ``I'm a novice Python student'') into a structured user profile $\mathcal{P}$ using an LLM.
The profile includes the user's preferred input dimensions ($\mathcal{P}_{in}$) and output dimensions ($\mathcal{P}_{out}$).
This structured profile is then used to guide both prompt personalization and output evaluation.


\paragraph{(B) Vibe dataset construction.}



To personalize what is being tested, we rewrite benchmark prompts based on the user's input dimensions.
For each benchmark sample $s$ and profile $\mathcal{P}$, we generate $K$ variations of the original prompt.
We do this by first generating a small set of editing options according to $\mathcal{P}_{in}$, such as ``request concise answer'' or ``emphasize efficiency''.
To create a new prompt variant, we sample a combination of these options and apply them to the original prompt. 
We then run a semantic-preservation verification using an LLM to flag variants that are likely to change the task intent (after pipeline refinement, almost all variants pass).
The resulting ``vibe dataset'' pairs each canonical benchmark sample $s$ with a set of $K$ controlled prompt variants conditioned on $\mathcal{P}_{in}$.

\paragraph{(C) Model Comparison.}  

We evaluate both whether a model solves the task and how well its response fits the user's preferences.
To measure correctness, we compute Pass@$1$ on the benchmark tests. 
To measure response preference, an LLM judge compares two model outputs side by side from the same user perspective.
For each output dimension in $\mathcal{P}_{out}$, the judge chooses the preferred response for that user along with a confidence score and rationale.
These pairwise judgments are aggregated into win rates to quantify shifts in model preferences across users.

The resulting pipeline is a personalized evaluation suite that reflects the user’s input and output preferences.
For each user, the pipeline yields a tailored set of test prompts and interpretable pairwise personal judgment scores. 
These judgments can be inspected at the dimension level, comparing models on specific dimensions across samples, or aggregated into an overall preference signal.
In this way, the pipeline captures both which model is preferred and why.
In the following section, we evaluate the pipeline on coding tasks across four profiles and four model matchups.

\subsection{Practitioner's guide}

Practitioners can use the pipeline for adapting evaluations to a target user or community. They can construct a user profile from direct descriptions, surveys, or other available user data, then use its input dimensions to personalize an existing benchmark or task collection and its output dimensions to define the evaluation criteria.
Results can be analyzed by dimension or aggregated across users, and open-ended tasks can use human review or task-specific rubrics.
See Appendix~\ref{appendix:practitioner_guide} for the full Practitioner's guide.

 \section{Experiments}


\subsection{Experimental setup}

\paragraph{Models.}
We study four head-to-head matchups between related models: (1) GPT-5.1~\citep{openai_gpt51_systemcard_2025} vs.\ GPT-OSS-20B~\citep{agarwal2025gpt}, (2) GPT-5.1 vs.\ GPT-4o~\citep{openai_gpt4o_systemcard_2024}, (3) Gemini-3 Pro~\citep{pichai2025new} vs.\ Gemma-3 4B~\citep{Kamath2025Gemma3T}, and (4) Qwen3-32B vs.\ Qwen3-14B~\citep{yang2025qwen3}.
These pairings allow us to test whether personalization reveals finer-grained trade-offs when the models being compared have an expected capability ordering (due to size differences or provider tiers).
Prompt personalization is done using GPT-5.1 and Qwen3-32B, with $k=2$ and $3$ variations, respectively.
Unless stated otherwise, we use GPT-5.1, GPT-OSS-20B, and Qwen3-14B as LLM judges (GPT-5.1 omitted for Gemini and Qwen comparisons due to cost).
We report judge agreement percentages and Cohen's Kappa.\footnote{See Appendix~\ref{appendix:judge_agreements} for more implementation details.}

\paragraph{Data and prompt variants.}
We use the MBPP+ and HumanEval+ Datasets~\citep{evalplus}, sampling $100$ problems from each.
For each persona and problem, we evaluate models under three prompt types: the original prompt, $K$ personalized variants, and $K$ neutral paraphrase controls produced with \textsc{PromptSuite}~\citep{habba-etal-2025-promptsuite}.
The control prompts change the wording without adding persona-specific information (e.g., adding ``Perform the following task:''), which helps isolate the effect of personalized paraphrasing.
We evaluate one generation per prompt due to cost, as partial experiments with GPT-5.1, GPT-OSS-20B, and Qwen models show a consistent win rate.
To check task preservation after personalization, we evaluate the correctness of personalized prompts using the original benchmark tests and report the preservation rate: the percentage of samples solved with both the original prompt and personalized rewrite. This is only a lower bound, since failures may also result from additional constraints or changes in response format. We report this for GPT-5.1 and Gemini-3 Pro and manually inspect 20 prompts that fail this check.

\begin{table*}[t]
  \centering
  \setlength{\tabcolsep}{5pt}
  \renewcommand{\arraystretch}{1.05}
  \begin{tabular}{c l| c c c c}
    \toprule
     &  &
    \multicolumn{4}{c}{\textbf{Win-rate (Tie-rate)}} \\
    \cmidrule(lr){3-6}
    \textbf{Model Pair} & \textbf{Prompt Type} & \textbf{Beginner} & \textbf{Intermediate} & \textbf{Researcher} & \textbf{Advanced} \\
    \midrule
    \multirow{3}{*}{%
      \begin{tabular}[c]{@{}c@{}}\texttt{GPT-5.1} \\\textit{vs.} \\\texttt{GPT-OSS-20B}\end{tabular}%
    }
      & Original
        & 0.03\textsuperscript{*} {\small(0.00)} & 0.01\textsuperscript{*} {\small(0.00)} & 0.08\textsuperscript{*} {\small(0.02)} & \textbf{0.91}\textsuperscript{*} {\small(0.01)} \\
      & Personalized
        & \textbf{0.77}\textsuperscript{*} {\small(0.02)} & \textbf{0.55}\textsuperscript{*} {\small(0.02)} & 0.43\textsuperscript{*} {\small(0.01)} & \textbf{0.58}\textsuperscript{*} {\small(0.03)} \\
      & Control
        & 0.06\textsuperscript{*} {\small(0.00)} & 0.02\textsuperscript{*} {\small(0.00)} & 0.10\textsuperscript{*} {\small(0.02)} & \textbf{0.93}\textsuperscript{*} {\small(0.01)} \\
    \midrule
    \multirow{3}{*}{%
      \begin{tabular}[c]{@{}c@{}}\texttt{GPT-5.1} \\\textit{vs.} \\\texttt{GPT-4o}\end{tabular}%
    }
      & Original
        & 0.09\textsuperscript{*} {\small(0.00)} & 0.16\textsuperscript{*} {\small(0.02)} & \textbf{0.63}\textsuperscript{*} {\small(0.02)} & \textbf{0.88}\textsuperscript{*} {\small(0.00)} \\
      & Personalized
        & \textbf{0.94}\textsuperscript{*} {\small(0.01)} & \textbf{0.77}\textsuperscript{*} {\small(0.02)} & \textbf{0.97}\textsuperscript{*} {\small(0.00)} & \textbf{0.82}\textsuperscript{*} {\small(0.02)} \\
      & Control
        & 0.08\textsuperscript{*} {\small(0.01)} & 0.19\textsuperscript{*} {\small(0.03)} & \textbf{0.70}\textsuperscript{*} {\small(0.02)} & \textbf{0.95}\textsuperscript{*} {\small(0.01)} \\
    \midrule
    \multirow{3}{*}{%
      \begin{tabular}[c]{@{}c@{}}\texttt{Gemini-3-Pro} \\\textit{vs.} \\\texttt{Gemma-3-4B}\end{tabular}%
    }
      & Original
        & 0.48 {\small(0.02)} & 0.53 {\small(0.01)} & 0.48 {\small(0.03)} & \textbf{0.66}\textsuperscript{*} {\small(0.02)} \\
      & Personalized
        & \textbf{0.93}\textsuperscript{*} {\small(0.01)} & \textbf{0.90}\textsuperscript{*} {\small(0.01)} & \textbf{0.97}\textsuperscript{*} {\small(0.01)} & \textbf{0.70}\textsuperscript{*} {\small(0.02)} \\
      & Control
        & 0.44\textsuperscript{*} {\small(0.02)} & 0.45\textsuperscript{*} {\small(0.02)} & 0.48 {\small(0.02)} & \textbf{0.69}\textsuperscript{*} {\small(0.02)} \\
    \midrule
    \multirow{3}{*}{%
      \begin{tabular}[c]{@{}c@{}}\texttt{Qwen3-32B} \\\textit{vs.} \\\texttt{Qwen3-14B}\end{tabular}%
    }
      & Original
        & 0.54 {\small(0.04)} & \textbf{0.62}\textsuperscript{*} {\small(0.03)} & 0.53 {\small(0.04)} & 0.38\textsuperscript{*} {\small(0.05)} \\
      & Personalized
        & 0.54 {\small(0.02)} & \textbf{0.67}\textsuperscript{*} {\small(0.04)} & \textbf{0.62}\textsuperscript{*} {\small(0.03)} & \textbf{0.54}\textsuperscript{*} {\small(0.04)} \\
      & Control
        & 0.54 {\small(0.05)} & \textbf{0.59}\textsuperscript{*} {\small(0.04)} & \textbf{0.55}\textsuperscript{*} {\small(0.06)} & 0.40\textsuperscript{*} {\small(0.05)} \\
    \bottomrule
  \end{tabular}
  \caption{
    \textbf{Personalization shifts win-rates.}
      Per-sample win rates on MBPP+ by user and prompt type, reported for the first model in each pair (tie-rate in parentheses).
      On benchmark prompts, \texttt{GPT-5.1} underperforms for Beginner and Intermediate, but personalized rewrites sharply increase its win rate, as they do for \texttt{Gemini-3-Pro}. 
      Both remain stronger for the Advanced persona in all prompts.
      For \texttt{Qwen3-32B} vs.\ \texttt{Qwen3-14B}, personalization yields a weaker shift toward the larger model. Control shows generic prompt rewrites match the original pattern.
    \textsuperscript{*} denotes statistical significance on a two-sided binomial test.
  }
  \label{tab:results_mbpp_win_rates}
\end{table*}

\paragraph{Personas and vibe dimensions.}
We use four author-written controlled user personas representing varying levels of coding expertise archetypes: 
\emph{Beginner Student}, \emph{Intermediate Learner}, \emph{AI Researcher}, and \emph{Advanced Developer}.
Each persona includes a description specifying both input and output preferences, and assigns importance weights from 1 to 5 to the output dimensions.
We also evaluate two real-life personas derived from public user data as a limited check that the observed effects extend beyond the controlled personas (See Appendix~\ref{appendix:real_personas}).
To control for prompt length, we also test one- to three-word personalized variants for the Beginner and Advanced personas (See construction details and results in Appendix~\ref{appendix:short_personalization}).

Because our experiments focus on single-turn coding, we evaluate the dimensions most relevant to that setting, excluding \emph{Stability} and \emph{Safety}.
We further replace \emph{Friction} and \emph{Ambiguity handling} with two narrower dimensions: \emph{Context awareness}, which captures whether the response respects the task context and constraints, and \emph{Persona consistency}, which captures whether the response fits the intended user role.

\paragraph{Pairwise judging and aggregation.}
For each persona, sample, and output dimension, an LLM judge assigns a pairwise label (\textsc{A wins}/\textsc{B wins}/\textsc{Tie}).
To mitigate position effects, we evaluate each comparison twice with the responses order-swapped following~\citet{jiang2025codejudgebench} and resolve disagreements using the judge's confidence score.
To determine the per-sample winner, we first compare $Pass@1$ correctness: if only one model is correct, it wins.
Otherwise, we aggregate the output dimensions judgments using the persona's weights.
We report per-dimension and overall win and tie rates across all judges.
\footnote{See Appendix~\ref{appendix:exp_details} for experimental details, and Appendix~\ref{appendix:more_results} for robustness analysis and ablations}

\paragraph{Human validation of LLM judgments.}

We validate the automated judging setup with a human preference study.
Six graduate student annotators compare model outputs in the same pairwise, persona-conditioned format used for LLM judges, selecting a winner or tie across dimensions.
This task is demanding -- it requires code understanding and careful comparison of lengthy responses across seven dimensions. We therefore keep the study small, with each annotator completing 12 easier-to-annotate comparisons in about one hour.
The study covers two personas (Beginner, Advanced), two model pairs (GPT-5.1 vs. GPT-4o, Gemini-3-Pro vs. Gemma-3-4B), and 32 sampled comparisons (24 original, 8 personalized), filtered for length and a minimal 2-judge consensus.
This checks whether our persona-conditioned LLM judgments align with human preferences (see Appendix~\ref{appendix:human_validation}).

\subsection{Results}\label{sec:experiments_and_results}

\paragraph{Personalization changes which model is preferred.}

As shown in Table~\ref{tab:results_mbpp_win_rates}, persona-specific rewrites often reverse the pattern seen under the original benchmark prompts.
For GPT-5.1 vs.\ GPT-OSS-20B and GPT-5.1 vs.\ GPT-4o, the original prompts favor the weaker or older model for the Beginner and Intermediate personas, whereas personalized prompts sharply increase GPT-5.1’s win rate.
For the more expert persona, GPT-5.1 is already mostly preferred on the original prompts and remains preferred after personalization.
Gemini-3 Pro vs.\ Gemma-3-4B preference is balanced on original prompts and led by Gemini-3 Pro on personalized prompts, with a substantially weaker shift for Qwen3-32B vs.\ Qwen3-14B.
In contrast, control paraphrases mostly preserve the original ordering, suggesting that the shift is unlikely to stem from generic rephrasing variations.



\begin{figure*}[t!]
  \centering
  \begin{subfigure}{0.49\linewidth}
    \centering
    \includegraphics[width=\linewidth,trim=0cm 0cm 0cm 0cm,clip]{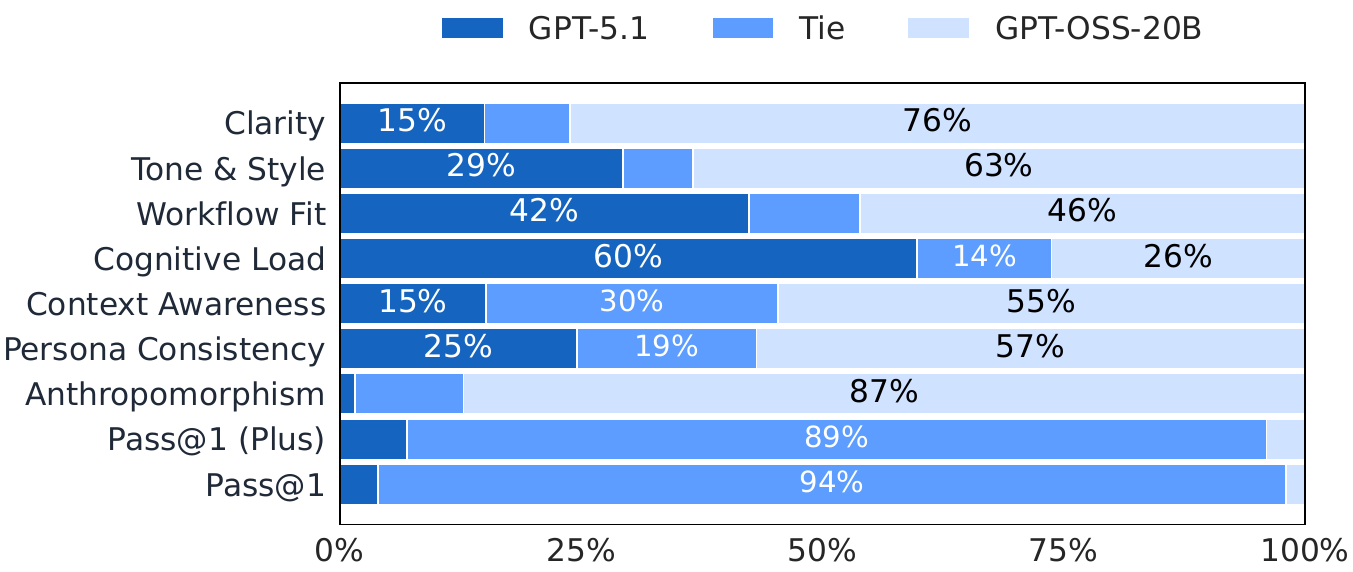}
    \caption{\textbf{Original prompts}}
    \label{fig:pairwise_results_original}
  \end{subfigure}\hfill
  \begin{subfigure}{0.49\linewidth}
    \centering
    \includegraphics[width=\linewidth]{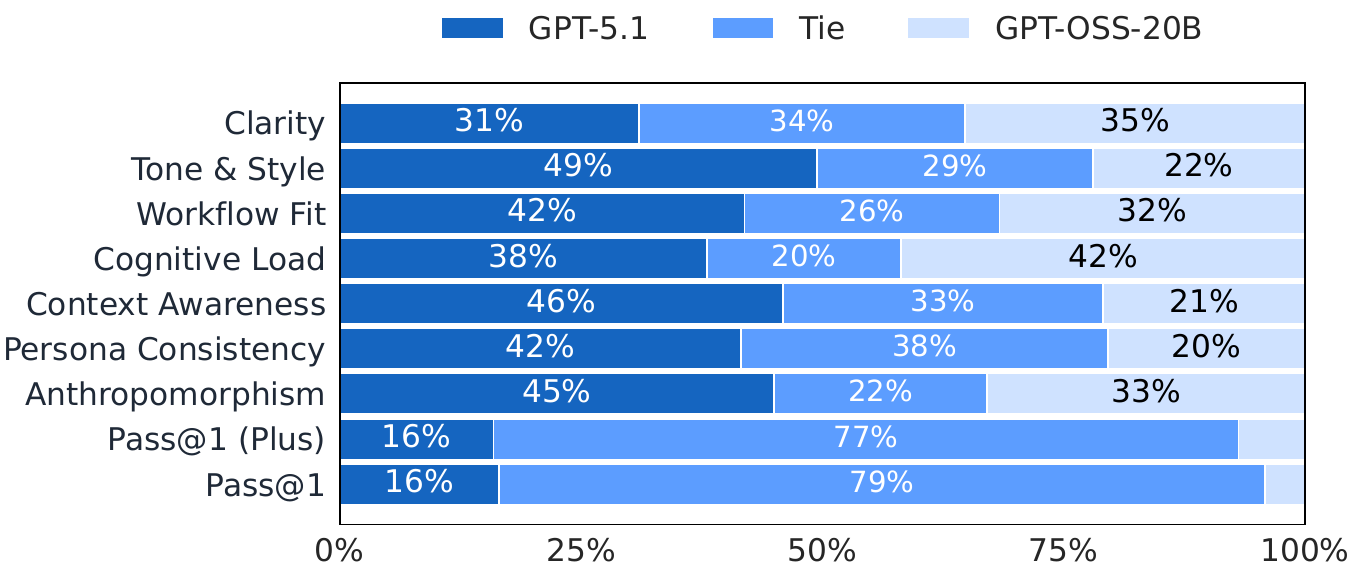}
    \caption{\textbf{Personalized}}
    \label{fig:pairwise_results_personlized}
  \end{subfigure}
   \caption{\textbf{Personalization changes model preferences.} Head-to-head win rates for GPT-5.1 vs.\ GPT-OSS-20B on MBPP+, broken down by dimensions. \textbf{Left:} original benchmark prompts. \textbf{Right:} persona-specific rewrites averaged over four personas. Several dimensions favor different models depending on the prompt form, showing that benchmark prompts can mask user-relevant differences beyond correctness.}
  \label{fig:head_to_head_results}
\end{figure*}

\paragraph{Dimension-level shifts .}

The dimension-level results help explain why overall preferences shift, as shown in Figure~\ref{fig:head_to_head_results}.
On the original prompts, GPT-OSS-20B is often favored over GPT-5.1 on clarity and tone, while GPT-5.1 leads on cognitive load.
After personalization, GPT-5.1 becomes more competitive across these subjective dimensions, reversing several gaps.
Manual inspection suggests that GPT-5.1 often defaults to concise, solution-focused responses, whereas GPT-OSS-20B gives verbose, tutor-like answers.
Responses to benchmark prompts may therefore mask user-relevant trade-offs by favoring one interaction style.


\paragraph{The effect generalizes across evaluation settings.}

We observe mostly similar trends when using Qwen3-32B as a generator, when evaluating on HumanEval+, and varying the winner rule, including unweighted scoring, majority vote, and strict tie-breaking (Appendix~\ref{appendix:more_results}).
The task preservation check shows $87\% \pm 2\%$, indicating personalization mostly preserves the original task ($20$ failed checks were manually verified to also preserve the original task).


\paragraph{The preferences of LLM judges are consistent and align with human judgment.}
Across samples, LLM judges reach reasonably consistent per-sample preferences with a mean agreement of $78\% \pm 13$ and Fleiss's $\kappa=0.39 \pm 0.16$.
Human validation on the original prompts shows high agreement both within humans ($94\% \pm 15$, $\kappa=0.80 \pm 0.39$) and between humans and LLM judges ($89\% \pm 16$, $\kappa=0.78 \pm 0.35$).
Agreement is lower on the small personalized-prompt subset, where longer responses increased annotation difficulty. We therefore treat the original-prompt results as support for the judging protocol, while the personalized results remain inconclusive (Appendix~\ref{appendix:more_results}).
Overall, these results support the reliability of preference judgments across users and samples.
\section{Conclusion}

We study vibe-testing as a real-world evaluation practice and present an empirically grounded formalization.
We turn this formalization into a proof-of-concept pipeline and show that personalization can change model preferences.
More broadly, our findings show that model quality depends not only on what a benchmark measures, but also on how users frame tasks and judge responses.
Formalized vibe-testing provides a practical foundation for future evaluation methods that systematically model user-dependent preferences.

\section{Limitations and future work}

Our empirical grounding is limited by both sources: the in-the-wild analysis relies on LLM-assisted extraction and manual verification, which limits scale and remains subjective, and the survey includes $51$ participants from a highly technical population.

We study a narrow setting: single-turn coding tasks with four author-written personas.
This setting provides an objective anchor for comparing correctness and subjective preference, but does not establish cross-domain generalization. It also excludes central vibe-testing settings in which correctness is partial or difficult to verify.

The pipeline also relies on LLMs for rewriting and judging, so rewrites may drift from the intended user framing or task, and LLM judgments may be biased and not fully match human preferences.
Prompt personalization changes several factors at once: length, context, constraints, and requested style.
Human and LLM judgment agreement on personalized prompts is low and inconclusive, so the observed shifts may partly reflect systematic judge preferences.



Our modular pipeline is simplified by design. Future work can extend it to multi-turn~\citep{liao2024automatic,Lu2025CanLA} and tool-augmented settings~\citep{chi2024amongagents}, strengthen human validation~\citep{kim2024evallm}, replace hand-crafted personas with learned profiles~\citep{davidson2023user,jiang2025know,zhao2025personalens,fu2025pref,Wang2025KnowYF}, and expand beyond coding~\citep{li2024iqa,chang2025chatbench}. 


\section*{Ethics statement}

Our work studies and operationalizes a form of informal evaluation that users already practice. We do not train new language models or deploy a user-facing system; we propose an evaluation pipeline and report controlled experiments.

\paragraph{Human subjects.}
We collected a survey from volunteer respondents recruited via public social media posts and ran a human annotation task with volunteer annotators. Participation in both was optional and uncompensated. We did not collect sensitive personal data beyond coarse self-reported background and usage habits, and all reported results are aggregated or anonymized. We submitted the study for institutional ethics review as required by our relevant institutions.

\paragraph{In-the-wild sources.}
Our in-the-wild corpus is derived from publicly available model comparison reports. We use these sources to study evaluation practices rather than to profile individuals.
We construct two coarse coding profiles from public Reddit and YouTube material. We extract only task and response preferences relevant to coding assistance, do not infer sensitive attributes, and omit unnecessary personal information from the released profiles.
In released artifacts, we retain public source metadata, including names and links to the original materials, and attribute all content to its original authors. All rights to the original materials remain with their respective authors or publishers.

\paragraph{LLM usage and potential harms.}
Our pipeline uses LLMs for prompt rewriting and for automated judging. These components may encode biases, including preferences for certain writing styles or verbosity, and could disadvantage particular user groups or interaction styles if used without validation. We mitigate position bias via swapped-order judging and include controls for generic paraphrasing, but automated judging remains imperfect.
We therefore view the pipeline as a tool for surfacing trade-offs, not a definitive arbiter of model quality.

\section*{Acknowledgments}
This research was supported by the Israel Science Foundation (grant No. 2942/25), by the Israeli Ministry of Science (grant no. 7256), and the European Union (ERC, Control-LM, 101165402). Views and opinions expressed are those of the authors only and do not necessarily reflect those of the European Union or the European Research Council Executive Agency; neither the European Union nor the granting authority can be held responsible for them.
We thank the ``Google Academic Program Award'' for providing access to Gemini.
We would like to express our gratitude to Gili Lior for valuable feedback and thoughtful comments throughout this work, and also thank Dana Arad, Tomer Ashuach, Orian Dabod, Noam Dahan, Shahar Levy, Nir Mazor, and Michael Toker for their assistance and support.

\bibliography{colm2026_conference}
\bibliographystyle{colm2026_conference}

\newpage
\appendix

\section{Survey extended details}\label{appendix:full_survey}

We recruited volunteer respondents via public social media posts aimed at a technical audience (e.g., X and Reddit).
Participation was optional and uncompensated. We collected coarse self-reports of technical background and usage habits, as well as questions about evaluation routines, perceived benchmark gaps, and interest in automation.
Before answering any questions, participants were shown the information and definition as in Figure~\ref{fig:survey-preamble}.
In the released survey results, examples of open-ended prompts and ``golden prompts'' were omitted to preserve respondent privacy.
The full survey questions and answer distributions are in Table~\ref{tab:survey_part_a}, Table~\ref{tab:survey_part_b}, Table~\ref{tab:survey_part_c}, and Table~\ref{tab:survey_part_d}.

Some questions were optional or multi-select. We therefore compute percentages with respect to the number of respondents who answered each question (or selected at least one option in multi-select questions). Likert-style questions are reported as distributions over the 1--7 scale. When reporting means in the main text, we compute them from the raw counts; due to rounding in percentage displays, means computed from the rounded percentages may differ slightly.

Figure~2 in the main paper visualizes two multiple-choice questions from Table~\ref{tab:survey_part_b} and Table~\ref{tab:survey_part_d}: (i) perceived benchmark failures (Q11), and (ii) common vibe-testing methods (Q7).
We plot the same percentages, ordered by frequency, and use the table above as the authoritative reference for exact values.



\begin{figure*}[t]
\centering

\begin{tcolorbox}[
  enhanced,
  breakable,
  width=\textwidth,
  colback=originalbg,
  colframe=originalframe,
  fonttitle=\bfseries\small,
  title={\strut Original Sample (MBPP+)},
  coltitle=black,
  boxrule=0.5pt,
  arc=2pt,
  top=4pt, bottom=4pt, left=6pt, right=6pt,
  after skip=6pt
]
\textit{``Write a python function to check whether the two numbers differ at one bit position only or not.''}
\end{tcolorbox}

\begin{tcolorbox}[
  enhanced,
  breakable,
  width=\textwidth,
  colback=promptbg,
  colframe=promptframe,
  fonttitle=\bfseries\small,
  title={\textcolor{white}{\strut Beginner Personalized Prompt \textnormal{(generated by GPT-5.1)}}},
  colbacktitle=headerbg,
  boxrule=0.4pt,
  arc=2pt,
  top=4pt, bottom=4pt, left=6pt, right=6pt,
  after skip=6pt
]
\footnotesize
``I am \texttt{python\_novice}, a student learning Python for introductory statistics using simple class-survey style examples, and I need to write a clear Python function that checks whether two integers differ at exactly one bit position only (no objects or advanced patterns, just a plain function I can run in a notebook). Please explain everything as if I only know basic Python, prefer readability over optimization, and structure the answer so it can be copied cell-by-cell into a Jupyter notebook, with short paragraphs starting with ``Explanation:'' before each code block describing what the code is about to do. After the main solution, include a brief recap of how each part of the code helps solve the task, connect the idea to simple stats-style situations (like thinking of binary flags in survey responses), and keep the required background limited to basic Python, not any special libraries. Also, after the explanation, suggest a few tiny experiments I can try (such as changing the input numbers or trying different bit patterns) and add a short reflection on what the results show about how the function works.''
\end{tcolorbox}

\begin{tcolorbox}[
  enhanced,
  breakable,
  width=\textwidth,
  colback=promptbg,
  colframe=promptframe,
  fonttitle=\bfseries\small,
  title={\textcolor{white}{\strut Intermediate Personalized Prompt \textnormal{(generated by GPT-5.1)}}},
  colbacktitle=headerbg,
  boxrule=0.4pt,
  arc=2pt,
  top=4pt, bottom=4pt, left=6pt, right=6pt,
  after skip=6pt
]
\footnotesize
``I'm working on a tiny personal script that compares two integer IDs and needs a Python function \texttt{differ\_by\_one\_bit(a, b)} that returns \texttt{True} only when the two numbers differ in exactly one bit position (and \texttt{False} otherwise), and it should run efficiently for lots of checks in a row, without worrying about micro-optimizations. I already have a broken stub:

\medskip
\begin{quote}
\ttfamily
def differ\_by\_one\_bit(a, b):\\
\quad\quad\# TODO: this is incomplete / may be wrong, fix or rewrite it\\
\quad\quad return None
\end{quote}
\smallskip

\noindent I don't need basic syntax explained (like if/else or simple loops); instead, I want the work organized as (1)~read/validate the two inputs; (2)~determine whether they differ at exactly one bit; (3)~return a boolean result, plus a short comment block above the function explaining in 2--3 sentences why the approach works and an `Explanation' section after the code with brief Goal/Approach/Edge Cases notes. If anything about the requirement seems unclear, I want to pause and write down my clarifying questions first, and after my solution passes tests I'll add a short reflection on what I learned and one mistake I might have made without tests.''
\end{tcolorbox}

\begin{tcolorbox}[
  enhanced,
  breakable,
  width=\textwidth,
  colback=promptbg,
  colframe=promptframe,
  fonttitle=\bfseries\small,
  title={\textcolor{white}{\strut AI Researcher Personalized Prompt \textnormal{(generated by GPT-5.1)}}},
  colbacktitle=headerbg,
  boxrule=0.4pt,
  arc=2pt,
  top=4pt, bottom=4pt, left=6pt, right=6pt,
  after skip=6pt
]
\footnotesize
``I want to add a small, reusable utility to my existing research repo: implement a Python function that, given two integer-like values, determines whether they differ in exactly one bit position, designed so it can be reused across different model-training pipelines and experiments. Please structure the code into small, pure functions with straightforward control flow, include explicit type hints for all public interfaces, and keep the design generic (not tied to any specific dataset or model) so it can be easily profiled, benchmarked, and extended by collaborators. In addition to the main boolean result, have the top-level function return a lightweight metadata dictionary (e.g., counts or intermediate values) to support downstream analysis and ablation studies.''
\end{tcolorbox}

\begin{tcolorbox}[
  enhanced,
  breakable,
  width=\textwidth,
  colback=promptbg,
  colframe=promptframe,
  fonttitle=\bfseries\small,
  title={\textcolor{white}{\strut Advanced Personalized Prompt \textnormal{(generated by GPT-5.1)}}},
  colbacktitle=headerbg,
  boxrule=0.4pt,
  arc=2pt,
  top=1pt, bottom=1pt, left=6pt, right=6pt,
  after skip=2pt
]
\footnotesize
``I'm working on performance-sensitive code in a production service and need a clean, reliable helper for bit comparisons. Task: implement a Python function that determines whether two given integers differ at exactly one bit position, suitable for inclusion in our existing codebase and CI pipeline. Provide only the final implementation (with a simple, well-named API) plus a very brief note on the core idea and time/space complexity, written as you would for code review, with no extra commentary.''
\end{tcolorbox}

\caption{\textbf{Example of expertise-level personalized prompts generated by GPT-5.1 for an MBPP+ coding task.} Given a single original problem statement, the pipeline produces personalized prompts for the user profiles. 
Each of them reflects different assumptions about prior knowledge, desired explanation depth, and code style preferences.}
\label{fig:personalized-prompts-example-gpt5}
\end{figure*}

\begin{figure*}[t]
\centering

\begin{tcolorbox}[
  enhanced,
  breakable,
  width=\textwidth,
  colback=originalbg,
  colframe=originalframe,
  fonttitle=\bfseries\small,
  title={\strut Original Sample (MBPP+)},
  coltitle=black,
  boxrule=0.5pt,
  arc=2pt,
  top=4pt, bottom=4pt, left=6pt, right=6pt,
  after skip=6pt
]
\textit{``Write a python function to check whether the two numbers differ at one bit position only or not.''}
\end{tcolorbox}

\begin{tcolorbox}[
  enhanced,
  breakable,
  width=\textwidth,
  colback=promptbg,
  colframe=promptframe,
  fonttitle=\bfseries\small,
  title={\textcolor{white}{\strut Beginner Personalized Prompt \textnormal{(generated by Qwen3-32B)}}},
  colbacktitle=headerbg,
  boxrule=0.4pt,
  arc=2pt,
  top=4pt, bottom=4pt, left=6pt, right=6pt,
  after skip=6pt
]
\footnotesize
``Hi there! I'm trying to learn how to compare numbers in Python using their binary form. Can you help me create a function that checks if two integers have exactly one different bit in their binary representation? I think this might involve some bitwise math, but I'm not sure how to count the differences. Let's make this simple to understand!

\medskip
\noindent Requirements:
\begin{itemize}[nosep, leftmargin=1.2em, topsep=2pt]
\item The function should take two integers as input and return \texttt{True} if they differ at exactly one bit position, \texttt{False} otherwise
\item Please explain each part of the code separately with plain language descriptions
\item It would help if you mentioned how binary representations and bitwise operations relate to statistical concepts like binary differences
\item Show step-by-step why each calculation is done that way
\end{itemize}

\medskip
\noindent For example, 13 (\texttt{1101}) and 9 (\texttt{1001}) should return \texttt{True} because only the second bit differs, but 2 (\texttt{10}) and 4 (\texttt{100}) should return \texttt{False} since they have two differing bits. Can you guide me through building this logic from scratch?''
\end{tcolorbox}

\begin{tcolorbox}[
  enhanced,
  breakable,
  width=\textwidth,
  colback=promptbg,
  colframe=promptframe,
  fonttitle=\bfseries\small,
  title={\textcolor{white}{\strut Intermediate Personalized Prompt \textnormal{(generated by Qwen3-32B)}}},
  colbacktitle=headerbg,
  boxrule=0.4pt,
  arc=2pt,
  top=4pt, bottom=4pt, left=6pt, right=6pt,
  after skip=6pt
]
\footnotesize
``I'm building a utility function for my coding projects to check if two integers differ by exactly one bit position. First, I need a clear plan: explain your approach in plain English, then implement the solution in three stages---input validation, bitwise comparison logic, and final result formatting. Make sure to handle edge cases like maximum integer values, zero, and numbers with only one differing bit in higher positions. Write the function \texttt{differ\_At\_One\_Bit\_Pos(a, b)} that returns \texttt{True} when exactly one bit differs between the inputs, \texttt{False} otherwise. Verify your solution works for both small values like \texttt{(2,\,3)} and large values like \texttt{(1048576,\,2097152)} without performance issues.''
\end{tcolorbox}

\begin{tcolorbox}[
  enhanced,
  breakable,
  width=\textwidth,
  colback=promptbg,
  colframe=promptframe,
  fonttitle=\bfseries\small,
  title={\textcolor{white}{\strut AI Researcher Personalized Prompt \textnormal{(generated by Qwen3-32B)}}},
  colbacktitle=headerbg,
  boxrule=0.4pt,
  arc=2pt,
  top=4pt, bottom=4pt, left=6pt, right=6pt,
  after skip=6pt
]
\footnotesize
``I am developing a bit-level analysis module for a research project. My task is to create a Python function that determines whether two integers differ by exactly one bit position. The solution must be structured with modular components for bitwise operations and result validation. I need reusable functions with domain-specific names like \texttt{compute\_bitwise\_difference} and \texttt{validate\_single\_bit\_variation} to ensure clarity. The implementation should include inline documentation for reproducibility in academic settings and configurable parameters for experimental testing. My code must pass standard test cases verifying single-bit differences across various integer pairs.''
\end{tcolorbox}

\begin{tcolorbox}[
  enhanced,
  breakable,
  width=\textwidth,
  colback=promptbg,
  colframe=promptframe,
  fonttitle=\bfseries\small,
  title={\textcolor{white}{\strut Advanced Personalized Prompt \textnormal{(generated by Qwen3-32B)}}},
  colbacktitle=headerbg,
  boxrule=0.4pt,
  arc=2pt,
  top=4pt, bottom=4pt, left=6pt, right=6pt,
  after skip=6pt
]
\footnotesize
``As a low-level systems programmer optimizing for embedded devices, I need you to implement a memory-efficient Python function with $O(1)$ space complexity to determine if two integers exhibit Hamming distance of exactly 1. The solution must execute within 200ms latency constraints on an 8-core ARMv8 architecture with 1MB heap limit. Your implementation should leverage bitwise operations for optimal performance and include quantitative benchmarks comparing execution time and memory footprint against a naive implementation using bitstring conversion. Debugging must account for race conditions in parallel processing scenarios where multiple threads validate bit differences concurrently.''
\end{tcolorbox}

\caption{\textbf{Example of expertise-level personalized prompts generated by Qwen3-32B for an MBPP+ coding task.} Given a single original problem statement, the pipeline produces personalized prompts for the user profiles: \emph{Beginner}, \emph{Intermediate}, \emph{AI Researcher}, and \emph{Advanced}. Each reflects different assumptions about prior knowledge, desired explanation depth, and code style preferences.}
\label{fig:personalized-prompts-example-qwen}
\end{figure*}

\begin{figure*}[t!]
  \centering
  \begin{subfigure}{0.49\linewidth}
    \centering
    \includegraphics[width=\linewidth,trim=0cm 0cm 0cm 0cm,clip]{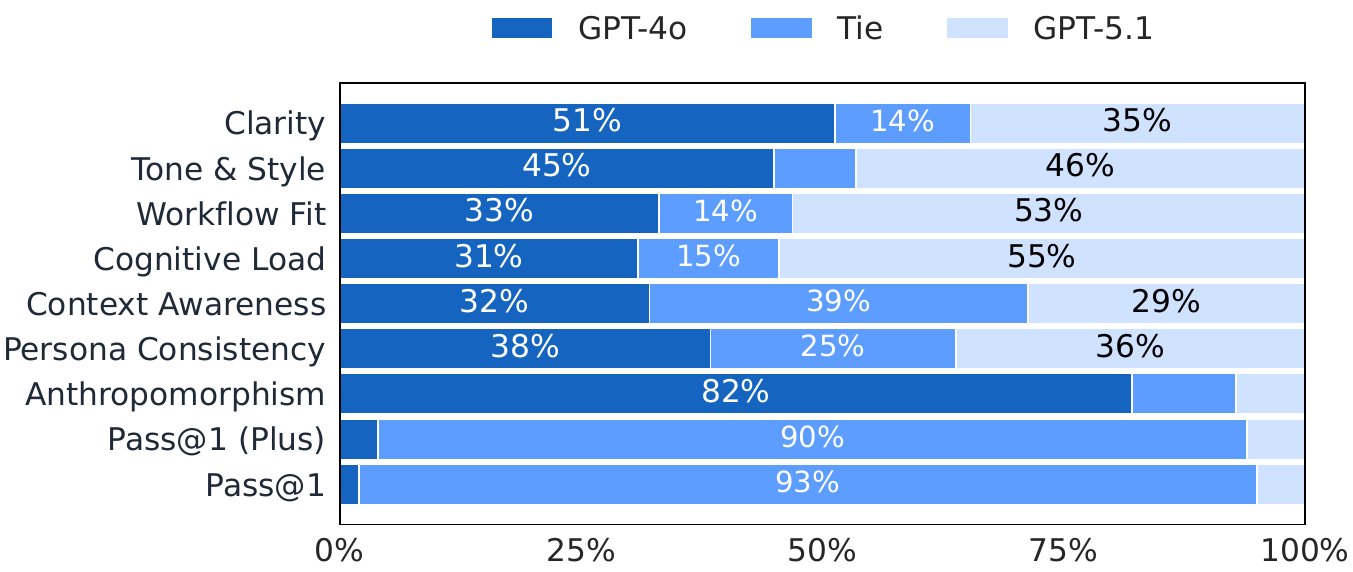}
    \caption{\textbf{Original prompts}}
    \label{fig:pairwise_results_original_gpt4o}
  \end{subfigure}\hfill
  \begin{subfigure}{0.49\linewidth}
    \centering
    \includegraphics[width=\linewidth]{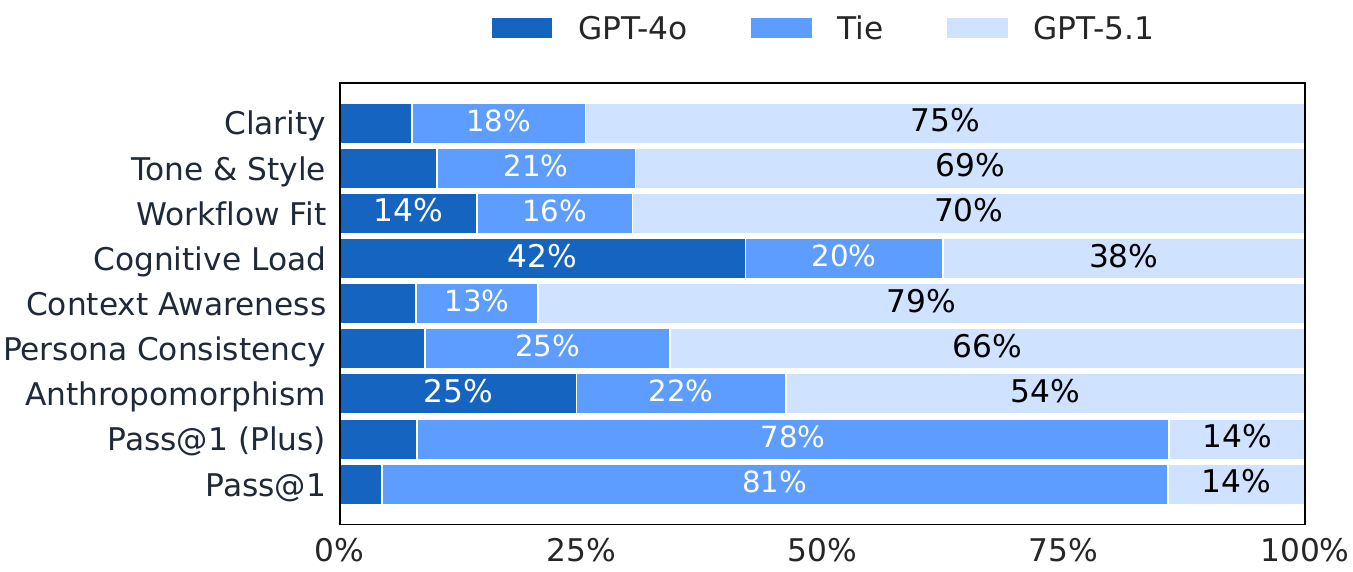}
    \caption{\textbf{Personalized}}
    \label{fig:pairwise_results_personlized_gpt4o}
  \end{subfigure}
   \caption{\textbf{Personalization changes model preferences GPT-5.1 vs.\ GPT-4o.} Head-to-head win rates for on MBPP+, broken down by dimensions. \textbf{Left:} original benchmark prompts. \textbf{Right:} persona-specific rewrites averaged over four personas. Several dimensions favor different models depending on the prompt form, showing that benchmark prompts can mask user-relevant differences beyond correctness.}
  \label{fig:head_to_head_results_gpt4o}
\end{figure*}

\begin{figure*}[t!]
  \centering
  \begin{subfigure}{0.49\linewidth}
    \centering
    \includegraphics[width=\linewidth,trim=0cm 0cm 0cm 0cm,clip]{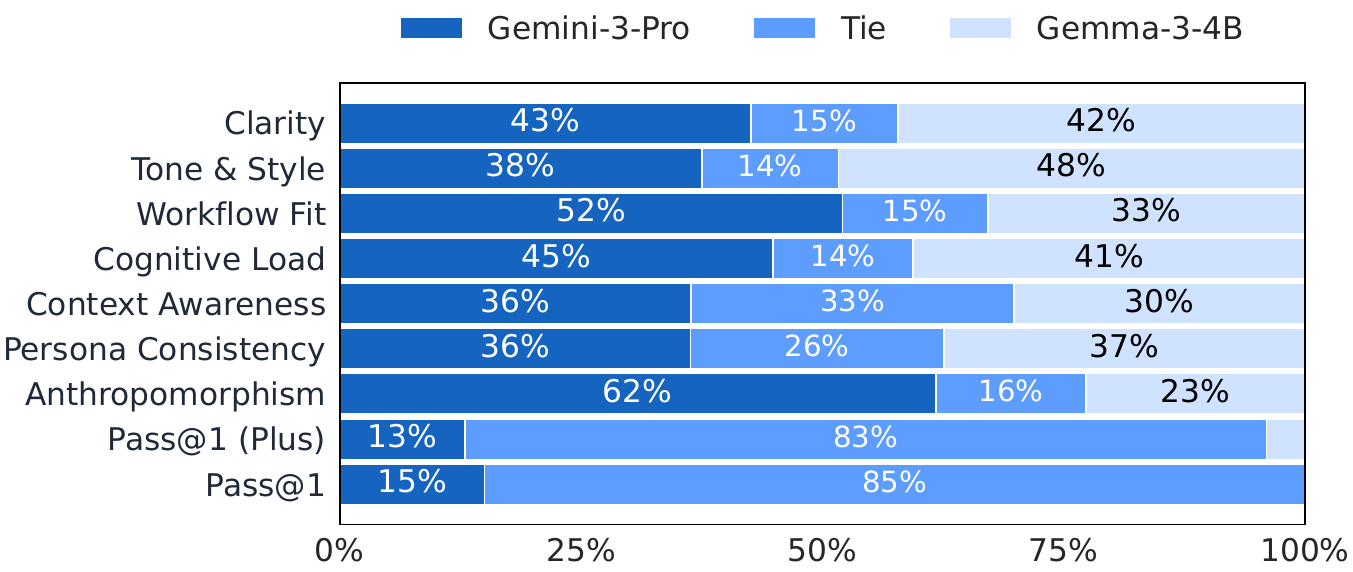}
    \caption{\textbf{Original prompts}}
    \label{fig:pairwise_results_original_gemini}
  \end{subfigure}\hfill
  \begin{subfigure}{0.49\linewidth}
    \centering
    \includegraphics[width=\linewidth]{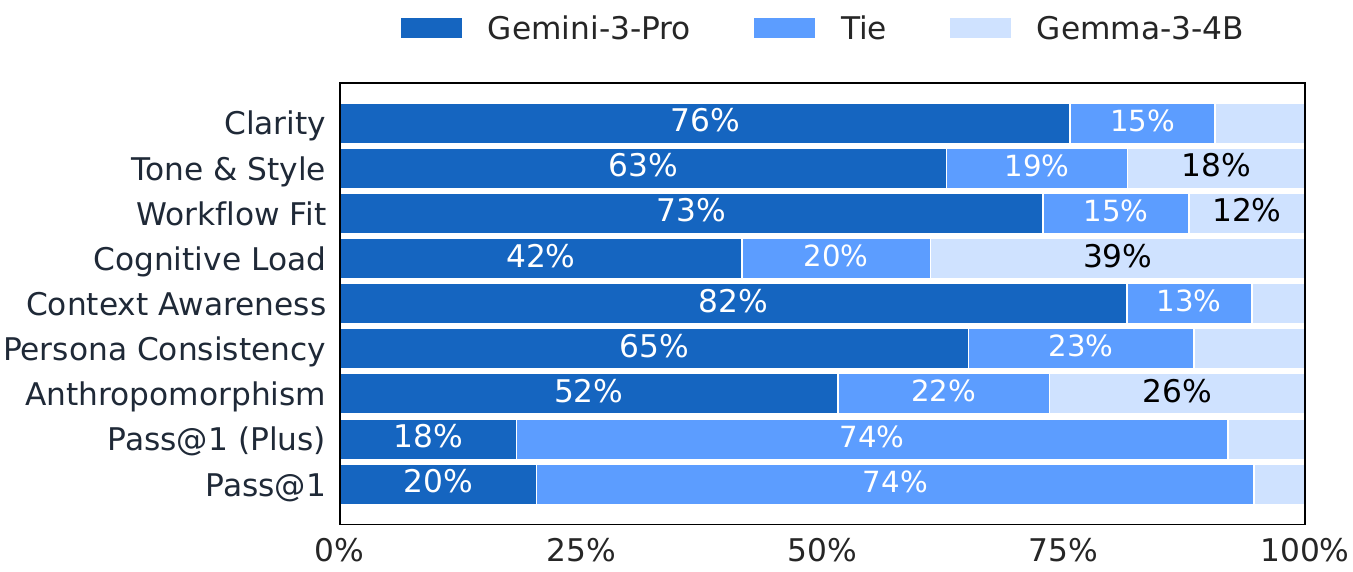}
    \caption{\textbf{Personalized}}
    \label{fig:pairwise_results_personlized_gemini}
  \end{subfigure}
   \caption{\textbf{Personalization changes model preferences for Gemini-3-Pro vs.\ Gemma-3-4B.} Head-to-head win rates on MBPP+, broken down by dimensions. \textbf{Left:} original benchmark prompts. \textbf{Right:} persona-specific rewrites averaged over four personas. Several dimensions favor different models depending on the prompt form, showing that benchmark prompts can mask user-relevant differences beyond correctness.}
  \label{fig:head_to_head_results_gemini}
\end{figure*}

\begin{figure*}[t!]
  \centering
  \begin{subfigure}{0.49\linewidth}
    \centering
    \includegraphics[width=\linewidth,trim=0cm 0cm 0cm 0cm,clip]{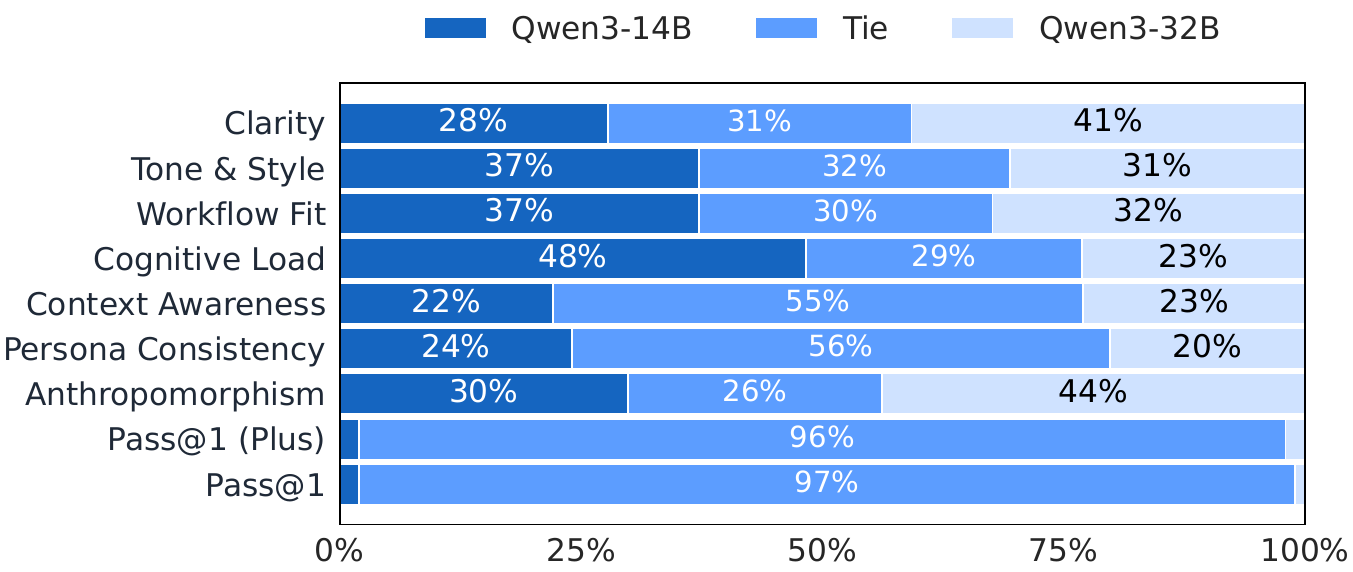}
    \caption{\textbf{Original prompts}}
    \label{fig:pairwise_results_original_qwen}
  \end{subfigure}\hfill
  \begin{subfigure}{0.49\linewidth}
    \centering
    \includegraphics[width=\linewidth]{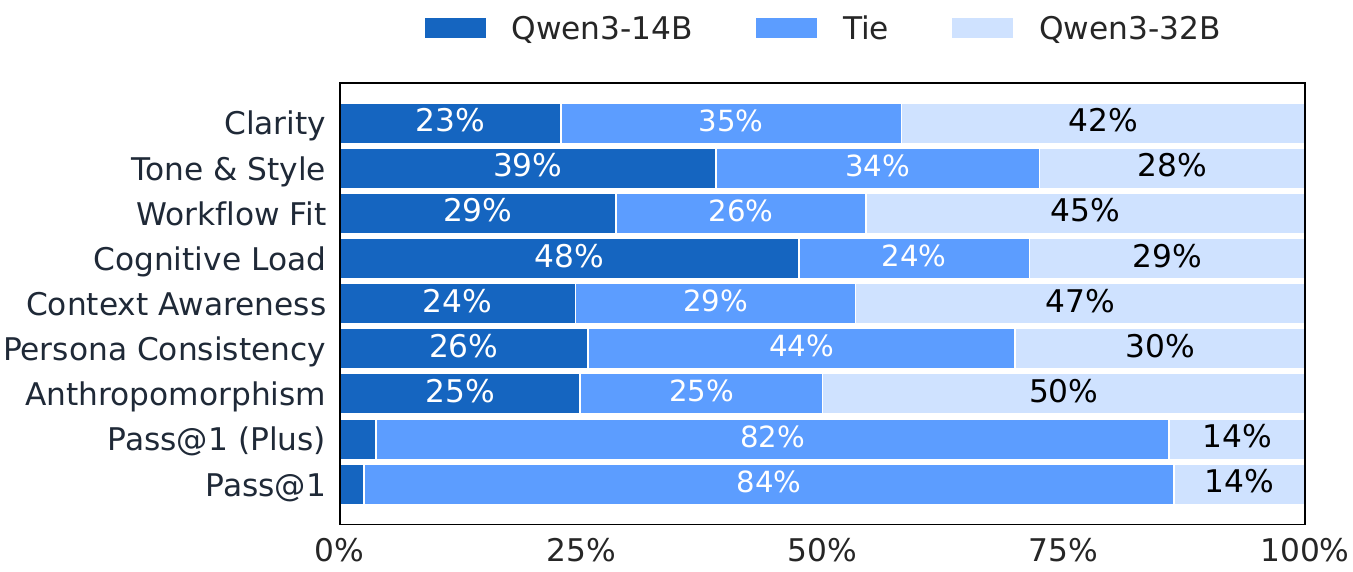}
    \caption{\textbf{Personalized}}
    \label{fig:pairwise_results_personlized_qwen}
  \end{subfigure}
   \caption{\textbf{Personalization changes model preferences for Qwen3-32B vs.\ Qwen3-14B.} Head-to-head win rates on MBPP+, broken down by dimensions. \textbf{Left:} original benchmark prompts. \textbf{Right:} persona-specific rewrites averaged over four personas. Several dimensions favor different models depending on the prompt form, showing that benchmark prompts can mask user-relevant differences beyond correctness.}
  \label{fig:head_to_head_results_qwen}
\end{figure*}





\begin{figure}
\centering
\begin{tcolorbox}[
  enhanced,
  breakable,
  width=\textwidth,
  colback=surveybg,
  colframe=surveyframe,
  fonttitle=\bfseries\small,
  title={\textcolor{white}{\strut Survey Preamble Shown to Participants}},
  colbacktitle=surveyheader,
  boxrule=0.5pt,
  arc=2pt,
  top=5pt, bottom=5pt, left=7pt, right=7pt,
]
\small
\textbf{Vibe-Testing AI Systems}

\medskip
Welcome to the Vibe-Testing AI Study!

Thank you for participating in our research on how people informally evaluate AI systems (``vibe-testing''). Your responses will help us understand real-world interaction patterns that are not captured by standard benchmarks.

\medskip
\noindent\textbf{Estimated Time:} ($\sim$)7--9 minutes

\noindent\textbf{Voluntary Participation:} Your participation is completely voluntary. You may stop at any time.

\noindent\textbf{No Compensation:} This survey does not include monetary compensation.

\noindent\textbf{Privacy:} Your responses will remain anonymous and used only for research purposes. We will not collect identifying information unless you choose to provide it at the end.

\noindent\textbf{Stay Informed:} If you would like to receive an email when the study results are published, please enter your email address at the end of this form. Your email, if provided, will be stored separately from your survey answers to protect your privacy.

\medskip
By continuing, you acknowledge that you understand this information and consent to participate.

\medskip
\noindent\textbf{What we're asking you to do}

This survey asks about how you interact with and evaluate AI systems when you use them. There are no right or wrong answers---we are interested in your honest experiences and opinions.

\medskip
\noindent\textbf{Vibe-Testing Definition (for this Survey)}

For the purposes of this study, vibe-testing means informally evaluating an AI model through direct interaction, using your own prompts or tasks to judge how the model performs in practice. This includes noticing qualities such as:
Usefulness for your workflow;
Clarity and structure of responses;
Tone or style;
How it handles ambiguity;
Whether it ``feels'' good to work with.

You may do this consciously (on purpose) or unconsciously (as part of how you explore models). We want to understand all such behaviors. Please answer candidly based on this definition and how you actually use AI models in your daily life.
\end{tcolorbox}

\caption{Full preamble and informed-consent text shown to survey participants before the main questionnaire.}
\label{fig:survey-preamble}
\end{figure}

\section{In-the-wild examples labeling}\label{appendix:full_in_the_wild_analysis}

We provide a detailed description of the four-stage procedure summarized in Section~\ref{sec:understanding_vibe_testing}, including the prompting setup used for extraction and labeling, the dimension consolidation process, and the final corpus-level analysis.

\subsection{Goal and scope}

The goal of this analysis is to empirically ground our definition of \emph{vibe-testing} by documenting how practitioners qualitatively evaluate LLMs in real-world settings.
Accordingly, the corpus is curated for concreteness and diversity rather than representativeness, and the analysis is qualitative and exploratory.

\subsection{Stage (1): source retrieval and selection}

We manually collected 40 public model comparison reports from four source types: YouTube review transcripts, Reddit threads, blog posts, and technology news articles.
Sources were identified via targeted searches combining model names (e.g., ``GPT'', ``Claude'', ``Gemini''), comparison terms (e.g., ``vs'', ``review'', ``comparison''), and subjective language (e.g., ``feels'', ``vibes'', ``I prefer'', ``works better for me'').
See source list in Table~\ref{tab:vibe_wild_sources_a} and Table~\ref{tab:vibe_wild_sources_b}.

We included sources that satisfy the three criteria stated in the main text:
(i) reference at least one specific model,
(ii) describe at least one concrete test input (prompt, task, or scenario),
and (iii) include qualitative judgments and subjective claims about model behavior or output quality.
We excluded sources that were purely promotional, focused exclusively on benchmark scores without qualitative discussion, or lacked concrete test inputs.
The final set of 40 sources was selected to maximize diversity of task types and interaction settings.

\subsection{Stage (2): vibe-test extraction and structured labeling}


\begin{figure*}[t]
\centering
\begin{tcolorbox}[
  enhanced,
  breakable,
  width=\textwidth,
  colback=promptbg,
  colframe=promptframe,
  fonttitle=\bfseries\small,
  title={\textcolor{white}{\strut Prompt: Vibe-Test Extraction and Labeling}},
  colbacktitle=promptheader,
  boxrule=0.4pt,
  arc=2pt,
  top=5pt, bottom=5pt, left=7pt, right=7pt,
]
\footnotesize
You are an NLP researcher analyzing transcripts to identify vibe-testing instances. A vibe-testing instance occurs when someone is:

(1)~evaluating one or more AI models or systems,
(2)~describing a specific task they used to judge the models,
and (3)~using subjective, qualitative impressions rather than quantitative metrics.

Subjective language includes terms like: ``vibes'', ``feels better/worse'', ``I prefer'', ``seems more'', clarity, tone, workflow fit, reliability, consistency, verbosity, creativity, naturalness, hallucination patterns, ``sounds like'', ``comes across as'', or any other qualitative judgment about model behavior or output quality.

\medskip
First, read the attached text.

Your task is to extract all vibe-testing instances from this text and return them as a JSON array.

\medskip
\texttt{<scratchpad>}\\
Before creating your final JSON output, use this space to:
\begin{itemize}[nosep, leftmargin=1.2em, topsep=2pt]
  \item identify potential vibe-testing instances,
  \item check each candidate against the criteria (evaluation + specific task + subjective language),
  \item note which subjective descriptors are used,
  \item verify that the instance describes an actual evaluation task, not just general discussion,
  \item exclude any mentions that are purely about benchmarks or quantitative metrics.
\end{itemize}
\texttt{</scratchpad>}

\medskip
Each item in your JSON array must include these fields:
\texttt{quote}, \texttt{task\_type}, \texttt{models\_mentioned}, \texttt{vibe\_language}, \texttt{why\_this\_is\_vibe\_testing}, \texttt{benchmark\_mention}, and \texttt{timestamp\_range} (if available) or comment \texttt{metadata} (if available).

\medskip
\noindent Important exclusions:
\begin{itemize}[nosep, leftmargin=1.2em, topsep=2pt]
  \item Do NOT include instances where someone only discusses benchmark scores or quantitative metrics without subjective evaluation on a specific task.
  \item Do NOT include technical discussions about architecture or training without evaluation context.
\end{itemize}

\medskip
Your output should consist only of the scratchpad followed by the final JSON array. The JSON should be properly formatted and valid.
\end{tcolorbox}

\caption{\textbf{LLM prompt used for vibe-test extraction and labeling from YouTube transcripts and Reddit threads.} Minor formatting adjustments were made between YouTube and Reddit to reflect available metadata.}
\label{fig:prompt-extraction}
\end{figure*}


\begin{figure*}[t]
\centering
\begin{tcolorbox}[
  enhanced,
  breakable,
  width=\textwidth,
  colback=promptbg,
  colframe=promptframe,
  fonttitle=\bfseries\small,
  title={\textcolor{white}{\strut Prompt: Dimension-Based Re-Annotation}},
  colbacktitle=promptheader,
  boxrule=0.4pt,
  arc=2pt,
  top=5pt, bottom=5pt, left=7pt, right=7pt,
]
\footnotesize
You are an expert NLP researcher annotating empirical examples for an ACL paper.
Your job is to carefully annotate vibe-testing examples based only on the information explicitly present in the input JSON.

\medskip
\noindent Critical rules:
\begin{itemize}[nosep, leftmargin=1.2em, topsep=2pt]
  \item Do not infer facts that are not directly stated.
  \item Do not guess the user's intent, persona, or evaluation criteria unless clearly supported.
  \item If something is unclear or missing, mark it as \texttt{uncertain} or \texttt{not stated}.
  \item Do not add benchmark claims unless the quote explicitly mentions benchmarks.
  \item Prefer under-annotation over speculation.
\end{itemize}

\medskip
Return a single JSON object following the provided schema, selecting vibe dimensions only from the closed set and justifying each assigned dimension using exact quoted language.

\smallskip
\hfill\textit{[Schema and closed dimension list as in the prompt shown in the main conversation.]}
\end{tcolorbox}

\caption{\textbf{LLM prompt used for dimension-based re-annotation.}  Each vibe-test instance re-annotated with the fixed dimension set.}
\label{fig:prompt-reannotation}
\end{figure*}


\begin{figure*}[t]
\centering
\begin{tcolorbox}[
  enhanced,
  breakable,
  width=\textwidth,
  colback=promptbg,
  colframe=promptframe,
  fonttitle=\bfseries\small,
  title={\textcolor{white}{\strut Prompt: Consistency Check and Gap Analysis}},
  colbacktitle=promptheader,
  boxrule=0.4pt,
  arc=2pt,
  top=5pt, bottom=5pt, left=7pt, right=7pt,
]
\footnotesize
You will be provided with (1)~the draft paper definitions and (2)~a JSON of labeled vibe-testing instances.
Your objective is to conduct a rigorous consistency check and gap analysis between the theoretical framework and the empirical data:
verify that the definitions encompass the subjective language used, verify coverage of all distinct vibes, identify instances that do not fit the proposed dimensions, and identify paper dimensions that lack supporting evidence.
Be critical and use direct quotes from the instances to support every critique.
\end{tcolorbox}

\caption{\textbf{LLM prompt used for the final consistency check and gap analysis.} Provided together with the current definitions and the consolidated JSON as inputs.}
\label{fig:prompt-consistency}
\end{figure*}

For each source, we used LLMs (GPT-5.2 and Gemini 3 Pro) to extract and label \emph{vibe-tests}, defined as localized \emph{instances} where an author evaluates one or more models on a specific task using subjective criteria.
For both YouTube transcripts and Reddit threads, the LLM was prompted to (a) produce a brief analysis identifying candidate vibe-testing instances and checking them against the definition, and then (b) output a JSON array of extracted vibe-tests with structured fields.

Each extracted vibe-test included:
\begin{itemize}
    \item \texttt{quote}: a direct excerpt capturing the instance (1--5 sentences);
    \item \texttt{task\_type}: a brief description of the tested task or scenario;
    \item \texttt{models\_mentioned}: a list of model names mentioned (empty if not specified);
    \item \texttt{vibe\_language}: the subjective descriptors used in the quote;
    \item \texttt{why\_this\_is\_vibe\_testing}: a short justification;
    \item \texttt{benchmark\_mention}: whether benchmarks were mentioned (\texttt{Yes}/\texttt{No});
    \item \texttt{timestamp\_range} (YouTube) or \texttt{metadata} when available (Reddit).
\end{itemize}

The authors manually verified the sampled extracted vibe-tests and iterated over the prompt to remove false positives (e.g., general opinions without a concrete task, or purely benchmark-driven comparisons), corrected extraction errors (e.g., truncated quotes, misattributed models), and retained only instances that clearly satisfy the operational definition.

\paragraph{LLM prompt for vibe-test extraction and labeling.}
We used the prompt in Figure~\ref{fig:prompt-extraction} (with minor formatting adjustments between YouTube and Reddit to reflect available metadata):

\subsection{Stage (3): trend analysis and common dimensions}

Stage (3) produces a consolidated, closed set of recurring subjective dimensions and applies it consistently across all vibe-tests.

\paragraph{(3a) Proposing and reconciling common dimensions.}
Starting from the open-ended \texttt{vibe\_language} and justifications produced in Stage (2), we asked the LLMs to propose lists of repeated subjective dimensions appearing across vibe-tests, grouping similar criteria under shared labels.
In parallel, we independently compiled candidate lists from manual review.
We then iteratively reconciled and refined these lists, merging overlapping categories, splitting overly broad ones, and discarding idiosyncratic or weakly supported dimensions, to derive a final closed set of common dimensions.

\paragraph{(3b) Dimension-based re-annotation of each vibe-test.}
After fixing the closed set of dimensions, we performed a second annotation pass in which each vibe-test was re-annotated using only this dimension list.
We prompted the model to output a single JSON object per vibe-test, selecting dimensions only when clearly supported by the text and providing per-dimension justifications citing exact phrases.
The model was explicitly instructed to avoid inferring missing information and to mark missing evidence as \texttt{not stated} or \texttt{uncertain}.
We compiled the resulting per-instance annotations into a single consolidated JSON file representing the full in-the-wild corpus.

\paragraph{Prompt for dimension-based re-annotation.}
We used the prompt in Figure~\ref{fig:prompt-reannotation} to re-annotate each vibe-test with the fixed dimension set:

\subsection{Stage (4): framework consistency check and gap analysis}

Finally, we provided (i) the draft paper definitions (including the proposed input/output dimensions) and (ii) the consolidated JSON of dimension-annotated vibe-tests to an LLM, and asked it to conduct a critical consistency check between the framework and the empirical data.

The model was instructed to:
\begin{itemize}
    \item verify that the paper definitions encompass the subjective language used in the vibe-tests,
    \item verify that the dimension inventory covers all distinct ``vibes'' found in the corpus,
    \item identify \textbf{unmapped instances} whose language does not fit the proposed dimensions,
    \item identify \textbf{unsupported theory} (dimensions defined in the paper but absent from the corpus),
    \item highlight \textbf{mismatches} where our definitions conflict with practitioner usage.
\end{itemize}

We used this output as a gap analysis to refine dimension definitions and ensure that the final framework closely reflects the empirical examples rather than purely theoretical assumptions.

\paragraph{Prompt for consistency check and gap analysis.}
For the final analysis, we used the prompt in Figure~\ref{fig:prompt-consistency} (with the paper definitions and the consolidated JSON provided as inputs):

\begin{quote}
\small
You will be provided with (1) the draft paper definitions and (2) a JSON of labeled vibe-testing instances.
Your objective is to conduct a rigorous consistency check and gap analysis between the theoretical framework and the empirical data:
verify that the definitions encompass the subjective language used, verify coverage of all distinct vibes, identify instances that do not fit the proposed dimensions, and identify paper dimensions that lack supporting evidence.
Be critical and use direct quotes from the instances to support every critique.
\end{quote}



\subsection{Vibe dimensions details}

\begin{table*}[t]
\centering
\setlength{\tabcolsep}{6pt}
\renewcommand{\arraystretch}{1.15}
\resizebox{\textwidth}{!}{%
\begin{tabular}{l|p{6.7cm}|p{5.8cm}}
\toprule
\textbf{Input Dimension} & \textbf{Operational Meaning} & \textbf{Typical Cues}  \\
\midrule


  \textbf{Task type} &
The kind of task used as the test, ranging from short, isolated prompts to multi-step scenarios (e.g., single-turn coding, debugging, planning, writing, workflow tasks). &
Mentions of ``my daily workflow,'' ``I tried debugging,'' ``I asked it to plan,'' or explicit task categories (code generation, refactor, summarization, reasoning). \\

\midrule

  \textbf{Task complexity} &
How demanding the test is in length, number of constraints, required steps, and need for long-range coherence. &
Short one-shot prompts vs.\ long prompts with many requirements, multi-step instructions, or tasks requiring sustained structure. \\

\midrule

  \textbf{Real-world context} &
Whether the prompt situates the task in a concrete domain or scenario (personal, professional, or everyday), rather than an abstract benchmark-like query. &
``In my codebase,'' ``for my project,'' ``as a product manager,'' ``explain like cooking,'' domain-specific details, constraints, or artifacts. \\

\midrule

  \textbf{Persona-based framing} &
How the user frames the interaction style, role, or audience (tutor vs.\ peer, novice vs.\ expert, tone requirements). &
``Explain to a beginner,'' ``be concise,'' ``act as a senior engineer,'' ``write like a legal memo,'' explicit tone or audience constraints. \\

\midrule

  \textbf{Underspecification level} &
How much the prompt leaves implicit, to test the model's ability to ask clarifying questions, make reasonable assumptions, or handle ambiguity. &
Vague instructions on purpose (``do something better''), missing details, open-ended tasks, or explicit stress-tests of ambiguity handling. \\

\midrule

  \textbf{Constraint tightness} &
How strictly the prompt constrains format, tools, style, or rules (schemas, word limits, citation rules, refusal constraints). &
``Return JSON with this schema,'' ``no markdown,'' ``must cite sources,'' ``do not use X,'' strict formatting or policy constraints. \\

\midrule

  \textbf{Reference availability} &
Whether the prompt provides grounding material (code, document excerpt, table) vs.\ asking from scratch. &
Presence of pasted context, files, or citations, vs.\ prompts that rely on the model's background knowledge alone. \\

\bottomrule
\end{tabular}%
}
\caption{\textbf{Taxonomy of vibe-testing dimensions.} We organize recurring axes of variation into input-oriented dimensions (what users choose to test and how they frame it) and output-oriented dimensions (how users compare outputs and what qualities they prioritize). Each dimension is intended to be actionable: it can be instantiated when constructing tests and referenced when judging model responses.}
\label{tab:vibe_taxonomy_input}
\end{table*}

\begin{table*}[t]
\centering
\setlength{\tabcolsep}{6pt}
\renewcommand{\arraystretch}{1.15}
\resizebox{\textwidth}{!}{%
\begin{tabular}{l|p{6.7cm}|p{5.8cm}}
\toprule
\textbf{Output
Dimension} & \textbf{Operational Meaning} & \textbf{Typical Cues} \\
\midrule

   \textbf{Comparison setup} &
How the user elicits and compares outputs, including side-by-side comparisons, reruns, and multi-prompt probing. &
``I compared A vs.\ B,'' ``ran it twice,'' ``tried a few variants,'' ``it was inconsistent across runs.'' \\
\midrule

   \textbf{Correctness} &
Whether the output is correct with respect to the task requirements and stated constraints. &
Explicit correctness judgments (``it works,'' ``it is wrong,'' ``hallucinated''), tests with executable code, or fact-check claims. \\
\midrule

   \textbf{Clarity} &
How readable and well organized the response is for the intended user, including logical flow and formatting. &
Mentions of ``easy to follow,'' ``well structured,'' ``confusing,'' ``jumps around,'' ``good explanation.'' \\
\midrule

   \textbf{Cognitive load} &
How mentally taxing the response is, given chunking, ordering, and amount of new concepts introduced at once. &
``Too much at once,'' ``overwhelming,'' ``nice step-by-step,'' ``dumped details,'' ``hard to track.'' \\
\midrule

   \textbf{Style/Tone fit} &
Alignment with the requested or expected tone, persona, and communication style for the user context. &
``Too formal,'' ``felt friendly,'' ``matches my style,'' ``sounds like marketing,'' ``too robotic.'' \\
\midrule

   \textbf{Workflow fit} &
Whether the output integrates into the user’s actual workflow, helping them complete the task with minimal friction. &
``I can paste this into my codebase,'' ``saved me time,'' ``usable,'' ``does not match how I work,'' ``not actionable.'' \\
\midrule

   \textbf{Friction} &
Whether the model resists the user’s intent in frustrating ways (fighting constraints, derailing from role, refusing incorrectly, or being hard to steer). &
``It kept insisting,'' ``ignored constraints,'' ``would not do what I asked,'' ``felt gaslighting,'' ``hard to steer.'' \\
\midrule

   \textbf{Ambiguity handling} &
How the model behaves under underspecified inputs, including making reasonable assumptions and asking clarifying questions. &
``Asked the right questions,'' ``made bad assumptions,'' ``filled in details incorrectly,'' ``handled vagueness well.'' \\
\midrule

   \textbf{Reliability} &
Consistency across reruns and robustness to small prompt changes, including sensitivity and variance. &
``Sometimes it works,'' ``unstable,'' ``high variance,'' ``breaks with tiny changes,'' ``consistent across attempts.'' \\
\midrule

   \textbf{Trustworthiness} &
Whether the model behaves safely and predictably, including refusal correctness and avoidance of harmful or risky guidance. &
Mentions of unsafe suggestions, unnecessary refusals, policy overreach, or trustworthy caution (especially on sensitive tasks). \\
\midrule

   \textbf{Anthropomorphism} &
Whether the interaction feels human-like vs.\ robotic, including naturalness, conversational coherence, and perceived ``personality.'' &
``Felt human,'' ``felt robotic,'' ``sounds like a real assistant,'' ``too template-y,'' ``has a vibe.'' \\

\bottomrule
\end{tabular}%
}
\caption{\textbf{Taxonomy of vibe-testing output dimensions: how users compare outputs and what they value in responses.} We organize recurring axes of variation into input-oriented dimensions (what users choose to test and how they frame it) and output-oriented dimensions (how users compare outputs and what qualities they prioritize). Each dimension is intended to be actionable: it can be instantiated when constructing tests and referenced when judging model responses.}\label{tab:vibe_taxonomy_output}
\end{table*}

The detailed list of input and output vibe dimensions is shown in Tables~\ref{tab:vibe_taxonomy_input} and Table~\ref{tab:vibe_taxonomy_output}.

\section{Experimental Details}\label{appendix:exp_details}

We provide full reproducibility details, including model inference settings, persona specifications, prompt templates, the evaluated dimension subset with guidance, and the judging and debiasing protocol.

\paragraph{Model inference settings.}
For each model, we report: (i) system prompt, (ii) decoding parameters (e.g., temperature/top-$p$/top-$k$ or greedy), (iii) token limits, and any provider-specific options.
For GPT-5.1, we use model ID \texttt{gpt-5.1-2025-11-13} and for Gemini-3-Pro \texttt{gemini-3-pro-preview}.
All GPT models run with thinking enabled when relevant (low effort) and with a $ 5,000$ token limit, and Qwen3 models use a $ 15,000$-token limit to accommodate longer reasoning traces.

\paragraph{Persona profiles.}
We provide the full persona specifications (YAML/JSON): persona description (Also in Table~\ref{tab:user-profiles}), input-dimension settings, output-dimension weights (1--5), and expressive-style instructions used by the prompt composer and judge.

\paragraph{Prompt variants.}
We include the exact templates for: (i) the original dataset prompt formatting, (ii) the personalized prompt rewriting prompts used by GPT-5.1 and Qwen3-32B, and (iii) the control prompt generation procedure and templates for \textsc{PromptSuite}.
We also document how variants are constructed for each persona.

\paragraph{Vibe dimensions used for coding.}
We list the final subset of vibe dimensions evaluated in coding assistance, with brief guidance text for each dimension as used by the judge.

\paragraph{Judging protocol and de-biasing.}
We provide the full judge prompt, the required output format (\textsc{A}/\textsc{B}/\textsc{Tie} plus rationale and confidence), and the position-swap procedure.
We specify the confidence-based conflict resolution rule and the alternative agreement-only rule used as a robustness check.

Regarding LLM judge model choice -- pilots with Llama3.3-70B~\citep{meta_llama33_modelcard_2024} and Gemini 3 Flash~\citep{googledeepmind_gemini3flash_modelcard_2025} yielded similar or lower-quality judgments and are omitted.

\subsection{Prompts and personas used in the pipeline}\label{app:prompts_personas}


\begin{figure*}[t]
\centering
\begin{tcolorbox}[
  enhanced,
  breakable,
  width=\textwidth,
  colback=promptbg,
  colframe=promptframe,
  title={\textcolor{white}{\strut Prompt: Persona Parsing}},
  colbacktitle=promptheader,
  boxrule=0.4pt,
  arc=2pt,
  top=5pt, bottom=5pt, left=7pt, right=7pt,
]
\footnotesize
You are an expert user experience researcher. Your task is to analyze the following user description and generate a structured JSON profile based on it.

\medskip
\noindent\textbf{User Description:}\\
``\texttt{\{description\}}''

\medskip
Based on this description, create a JSON object that captures the user's persona, their needs regarding coding tasks, and their preferences for the output.

\medskip
Your output MUST be a single, valid JSON object. Use the following structure and explanations as a guide.

\medskip
\noindent\textbf{JSON Template:}\\
\texttt{\{ \ldots \}}

\medskip
Generate the JSON object now.
\end{tcolorbox}

\caption{\textbf{Persona-parsing prompt.} Given a short natural-language user description, the LLM produces a structured JSON profile describing input and output preferences. The model must output a single JSON object and nothing else.}
\label{fig:prompt-persona-parsing}
\end{figure*}


\begin{figure*}[t]
\centering
\begin{tcolorbox}[
  enhanced,
  breakable,
  width=\textwidth,
  colback=promptbg,
  colframe=promptframe,
  title={\textcolor{white}{\strut Prompt: Change Identification (Profile $\rightarrow$ Concrete Modifications)}},
  colbacktitle=promptheader,
  boxrule=0.4pt,
  arc=2pt,
  top=5pt, bottom=5pt, left=7pt, right=7pt,
]
\footnotesize
You are an expert in personalizing programming tasks.

\medskip
\noindent\textbf{Task:}\\
Given the following user profile, identify 2--3 concrete, distinct ways to modify a programming problem prompt for each field of the profile.
The modifications should instruct how to adapt the prompt's phrasing, tone, and instructions to match the user's persona and preferences, without changing the core programming task.
\ldots

\medskip
\noindent\textbf{User Profile:}\\
---\\
\texttt{\{user\_profile\_json\}}\\
---

\medskip
\noindent\textbf{Output rules (MUST FOLLOW):}
\begin{enumerate}[nosep, leftmargin=1.5em, topsep=2pt]
  \item Output must be a single valid JSON object. No explanations, no comments, no markdown fences.
  \item Root object has exactly one key: \texttt{"changes\_by\_field"}.\\
  \ldots
\end{enumerate}

\medskip
Now, produce the JSON for all fields of the given user profile in exactly this format.
Verify the output is a valid JSON object!
\end{tcolorbox}

\caption{\textbf{Change-identification prompt.} To operationalize a persona profile into actionable prompt edits, the LLM proposes 2--3 concrete modification options for a fixed set of fields, while explicitly disallowing changes that alter the task itself. This stage outputs a single JSON object with a list of changes keyed by profile fields.}
\label{fig:prompt-change-identification}
\end{figure*}


\begin{figure*}[t]
\centering
\begin{tcolorbox}[
  enhanced,
  breakable,
  width=\textwidth,
  colback=promptbg,
  colframe=promptframe,
  title={\textcolor{white}{\strut Prompt: Personalized Prompt Composition}},
  colbacktitle=promptheader,
  boxrule=0.4pt,
  arc=2pt,
  top=5pt, bottom=5pt, left=7pt, right=7pt,
]
\footnotesize
You are an expert in rewriting programming tasks as a persona.
You will be given an original programming problem prompt and a list of modifications to apply.
Your task is to rewrite the prompt to incorporate all the specified changes while ensuring the underlying programming problem remains identical.
\ldots

\medskip
\noindent\textbf{Original Prompt:}\\
---\\
\texttt{\{original\_prompt\}}\\
---

\medskip
\noindent\textbf{Modifications to apply:}\\
---\\
\texttt{\{changes\_description\}}\\
---

\medskip
\noindent\textbf{Instructions:}
\begin{enumerate}[nosep, leftmargin=1.5em, topsep=2pt]
  \item Carefully read the original prompt and the list of changes.
  \item Rewrite the prompt to apply all changes cohesively.
  \item DO NOT alter the core requirements of the programming task. The new prompt must lead to the same solution.\\
  \ldots
  \item[11.] The length should not be longer than 2--4 short sentences.
  \item[12.] Your output MUST only contain the prompt text and nothing else.
\end{enumerate}
\end{tcolorbox}

\caption{\textbf{Personalized prompt composition.} Given an original benchmark prompt and the selected modifications, the LLM generates a personalized version that preserves the underlying programming task. The prompt is written in the persona voice (first person), avoids explicit references to the profile schema, and is constrained to a short length.}
\label{fig:prompt-composition}
\end{figure*}


\begin{figure*}[t]
\centering
\begin{tcolorbox}[
  enhanced,
  breakable,
  width=\textwidth,
  colback=promptbg,
  colframe=promptframe,
  title={\textcolor{white}{\strut Prompt: HumanEval+ Prefix Composition}},
  colbacktitle=promptheader,
  boxrule=0.4pt,
  arc=2pt,
  top=5pt, bottom=5pt, left=7pt, right=7pt,
]
\footnotesize
You are an expert in rewriting programming tasks as a persona.
\ldots\\
Your task is to write a prefix for the prompt to incorporate all the specified changes while ensuring the underlying programming problem remains identical.
\ldots

\medskip
\noindent\textbf{Original Prompt:}\\
---\\
\texttt{\{original\_prompt\}}\\
---

\medskip
\noindent\textbf{Modifications to apply:}\\
---\\
\texttt{\{changes\_description\}}\\
---

\medskip
\noindent\textbf{Instructions:}\\
\ldots
\begin{enumerate}[nosep, leftmargin=1.5em, topsep=2pt]
  \item[10.] The length should not be longer than 2--4 short sentences.
  \item[11.] Your output MUST only contain the prefix text and nothing else.
\end{enumerate}
\end{tcolorbox}

\caption{\textbf{Prompt for HumanEval+ prefix composition.} For HumanEval+ style prompts that include code context and docstrings, only a short persona prefix is produced and concatenated to the original prompt, avoiding perturbation of code formatting while still injecting persona-relevant framing.}
\label{fig:prompt-humaneval-prefix}
\end{figure*}


\begin{figure*}[t]
\centering
\begin{tcolorbox}[
  enhanced,
  breakable,
  width=\textwidth,
  colback=promptbg,
  colframe=promptframe,
  title={\textcolor{white}{\strut Prompt: Semantic-Preservation Verification}},
  colbacktitle=promptheader,
  boxrule=0.4pt,
  arc=2pt,
  top=5pt, bottom=5pt, left=7pt, right=7pt,
]
\footnotesize
You are an expert in verifying programming problem statements.
You will be given an original prompt and a modified version of it.
Your task is to determine if the modified prompt preserves the original's core task and would result in the same ground-truth solution.
\ldots

\medskip
\noindent\textbf{Original Prompt:}\\
---\\
\texttt{\{original\_prompt\}}\\
---

\medskip
\noindent\textbf{Modified Prompt:}\\
---\\
\texttt{\{modified\_prompt\}}\\
---

\medskip
\noindent\textbf{Decide:}
\begin{itemize}[nosep, leftmargin=1.2em, topsep=2pt]
  \item \texttt{"same\_end\_goal"}: Is the fundamental task identical in both prompts?
  \item \texttt{"same\_ground\_truth"}: Would the correct solution for the modified prompt also be correct for the original prompt?
\end{itemize}
\ldots

\medskip
Output MUST be a single valid JSON object, nothing else.
Use exactly these keys: \texttt{same\_end\_goal}, \texttt{same\_ground\_truth}, \texttt{reason\_if\_failed}.
\end{tcolorbox}

\caption{\textbf{Semantic-preservation verifier prompt.} To ensure personalized prompts remain faithful to the original benchmark intent, the verifier checks (i)~whether the end goal is identical and (ii)~whether the ground-truth solution set is preserved. The verifier returns two booleans and an error string if either check fails.}
\label{fig:prompt-verification}
\end{figure*}

\begin{figure*}[t]
\centering
\begin{tcolorbox}[
  enhanced,
  breakable,
  width=\textwidth,
  colback=promptbg,
  colframe=promptframe,
  title={\textcolor{white}{\strut Prompt: Persona Creator}},
  colbacktitle=promptheader,
  boxrule=0.4pt,
  arc=2pt,
  top=5pt, bottom=5pt, left=7pt, right=7pt,
]
\footnotesize
You will be writing a persona description for a programmer. A persona description is a brief first-person statement that captures the programmer's context (their role, experience level, work environment, etc.) and what they value when writing code (e.g., readability, performance, simplicity, etc.).

\medskip
First, here are some example persona descriptions to use as style references:

\medskip
\noindent\texttt{<example\_personas>}\\
\texttt{<1>}\\
\texttt{\{example\_persona\_1\}}\\
\texttt{</1>}

\medskip
\noindent\texttt{<2>}\\
\texttt{\{example\_persona\_2\}}\\
\texttt{</2>}\\
\texttt{</example\_personas>}

\medskip
Now, here are some Reddit posts written by the new user for whom you need to write a persona description:

\medskip
\noindent\texttt{<user\_info>}\\
\texttt{\{user\_info\}}\\
\texttt{</user\_info>}

\medskip
\noindent\textbf{Your task is to write a persona description for this new user that:}
\begin{itemize}[nosep, leftmargin=1.5em, topsep=2pt]
  \item Is written in first person (using "I", "my", etc.).
  \item Matches the style, tone, and approximate length of the example personas.
  \item Includes a brief description of the user's context (their role, background, work situation, or relevant circumstances).
  \item Describes what they value or prioritize in their code writing.
  \item Feels natural and authentic based on the information provided.
\end{itemize}

\medskip
Study the example personas carefully to understand the typical structure and level of detail. Then craft a new persona description tailored to the new user's specific information.

\medskip
Write your persona description inside \texttt{<persona>} tags.
\end{tcolorbox}

\caption{\textbf{Persona creation.} Given example personas and public posts from a target user, the LLM generates a brief first-person persona description that captures the user's programming context and coding preferences. The generated description follows the style and approximate length of the provided examples.}
\label{fig:persona-creator}
\end{figure*}

\begin{figure*}[t]
\centering
\begin{tcolorbox}[
  enhanced,
  breakable,
  width=\textwidth,
  colback=promptbg,
  colframe=promptframe,
  title={\textcolor{white}{\strut Prompt: Short Minimal Personalization}},
  colbacktitle=promptheader,
  boxrule=0.4pt,
  arc=2pt,
  top=5pt, bottom=5pt, left=7pt, right=7pt,
]
\footnotesize
You are an expert in subtly adapting programming task prompts to match a user persona.
You will be given an original programming problem prompt and a brief user persona summary.
Your task is to make ONLY minimal phrasing adjustments to the prompt that subtly reflect the user's background or style.

\medskip
\noindent\textbf{STRICT CONSTRAINTS:}
\begin{itemize}[nosep, leftmargin=1.5em, topsep=2pt]
  \item The modified prompt must NOT be longer than the original by more than 3--4 words.
  \item Do NOT add any new requirements, constraints, or instructions that were not in the original.
  \item Do NOT add explanations, examples, or clarifications that were not in the original.
  \item The modification should be limited to subtle word choice, phrasing, or contextual cues (e.g., adding "I need a function for my homework" for a student user).
  \item It must contain at least one user-specific detail for that persona from the user context or workflow mentioned in the possible modifications, without adding any new requirements, constraints, or instructions.
  \item The modified prompt should be human-like and natural, and not longer than the original beyond 3--4 words.
  \item Avoid phrasing such as "As a [PERSONA\_DETAIL]\ldots". Better phrasing is "I need to write\ldots because I'm a [PERSONA\_DETAIL]" or "I'm a [PERSONA\_DETAIL], write\ldots".
  \item The core programming task and its solution MUST remain identical.
\end{itemize}

\medskip
\noindent\textbf{User Persona Summary:}\\
---\\
\texttt{\{persona\_summary\}}\\
---

\medskip
\noindent\textbf{Possible modifications to apply (choose one):}\\
---\\
\texttt{\{changes\_description\}}\\
---

\medskip
\noindent\textbf{Original Prompt:}\\
---\\
\texttt{\{original\_prompt\}}\\
---

\medskip
\noindent\textbf{Instructions:}
\begin{enumerate}[nosep, leftmargin=1.5em, topsep=2pt]
  \item Read the original prompt and the persona summary.
  \item Make only subtle phrasing changes (word substitutions, minor rewordings) that reflect the persona.
  \item Keep the prompt nearly identical in length and structure to the original.
  \item Do NOT add constraints, instructions, or any new content, including output format.
  \item Your output MUST be a single valid JSON object with exactly one key \texttt{"modified\_prompt"}.
  \item No extra keys, no trailing commas, no markdown fences, no commentary.
\end{enumerate}

\medskip
\noindent\textbf{Correct Examples:}

\medskip
\noindent\textbf{Novice user example}\\
\textbf{Original:} \texttt{"Write a function sum\_list(nums) that returns the sum of a list of integers."}\\
\textbf{Persona:} \texttt{"A data science beginner learning Python"}

\medskip
\noindent
\texttt{\{\{"modified\_prompt": "Help me write a function sum\_list(nums) for my stat course, it should return the sum of a list of integers."\}\}}

\medskip
\noindent\textbf{Advanced developer example}\\
\textbf{Original:} \texttt{"Write a function sum\_list(nums) that returns the sum of a list of integers."}\\
\textbf{Persona:} \texttt{"An experienced developer"}

\medskip
\noindent
\texttt{\{\{"modified\_prompt": "Write a highly optimized function sum\_list(nums) that returns the sum of a list of integers."\}\}}
\end{tcolorbox}

\caption{\textbf{Short minimal personalization.} Given an original programming prompt, a user persona, and a set of possible modifications, the LLM makes a minimal persona-relevant phrasing change. The rewritten prompt is limited to an increase of at most three to four words and must preserve the original task, requirements, and solution.}
\label{fig:short-minimal-personalization}
\end{figure*}

We release the full prompts used across all stages of the pipeline: persona parsing, prompt-change proposal, personalized prompt generation and verification, and subjective pairwise judging.

\subsubsection{Personas}
\label{app:prompts_personas:personas}

The pipeline evaluates models under four fixed user personas that span a novice-to-expert spectrum (Section~\ref{sec:experiments_and_results}). Each persona is represented as a structured profile with (i) a short natural-language description and (ii) explicit preferences over the input and output dimensions. Profiles are used in two places: (1) to personalize benchmark prompts while preserving the underlying task, and (2) to condition pairwise subjective judging on what a given persona values.

In our release, persona profiles are stored as JSON/YAML files, one per persona, with a shared schema. Each profile includes a \texttt{persona\_description} field, an \texttt{input\_dimensions} object, and an \texttt{output\_dimensions} object whose entries are either categorical labels (for discrete dimensions) or importance weights on a 1--5 scale (for prioritized output criteria).
See the personas description in Table~\ref{tab:user-profiles}.

\paragraph{Persona Creator}

We include in our code a \textit{Persona Creator} - a tool that converts unstructured user information into the structured profile format used by the pipeline.
It accepts direct user descriptions, survey responses, public posts, and other collections of user-provided text.
The output follows the profile schema used in our experiments, including input dimensions for prompt personalization, output dimensions for evaluation, and optional importance weights over output dimensions.
The Persona Creator extracts only preferences supported by the source material. Unsupported fields are left unspecified to avoid making assumptions about the user.
Generated profiles can be manually reviewed and edited before use. See our code for the exact generation prompt, output schema, and example configurations.


\begin{table*}[t]
    \centering
    \small
    \renewcommand{\arraystretch}{1.15}
    \setlength{\tabcolsep}{6pt}
    \begin{tabular}{p{0.19\linewidth}|p{0.77\linewidth}}
    \hline
    \textbf{Profile} & \textbf{Description and Preference Weights} \\
    \hline
    Beginner Student &
    "I am an undergraduate student with limited Python experience, mainly using it for basic data analysis in my statistics courses. I need code that is very clear and well-explained."

    {\footnotesize \textit{Weights:} Clarity: 5, Style: 1, Workflow Fit: 2, Cognitive Load: 1, Context Awareness: 4, Persona Consistency: 4, Anthropomorphism: 5.} \\[0.5em]
    \hline
    Intermediate Learner &
    "I have moderate coding experience and typically work on debugging and small projects. I care most about getting the correct solution, and I appreciate helpful clarifications that support my learning as I go."

    {\footnotesize \textit{Weights:} Clarity: 5, Style: 1, Workflow Fit: 3, Cognitive Load: 1, Context Awareness: 4, Persona Consistency: 4, Anthropomorphism: 3.} \\[0.5em]
    \hline
    AI Researcher &
    "I am a researcher in the field of machine learning. I need code that is efficient and easy to modify and trace."

    {\footnotesize \textit{Weights:} Clarity: 4, Style: 1, Workflow Fit: 4, Cognitive Load: 1, Context Awareness: 4, Persona Consistency: 4, Anthropomorphism: 1.} \\[0.5em]
    \hline
    Advanced Developer &
    "I am an experienced programmer and I often work on optimization and complex debugging. I prefer efficient solutions with minimal guidance and concise outputs."

    {\footnotesize \textit{Weights:} Clarity: 3, Style: 1, Workflow Fit: 5, Cognitive Load: 1, Context Awareness: 4, Persona Consistency: 4, Anthropomorphism: 1.} \\
    \hline
    \end{tabular}
    \caption{Representative user persona used to instantiate personalized vibe-testing in coding assistance. Each persona includes a short natural-language description and importance weights for the output dimensions used in evaluation. These descriptions and weights were chosen before running experiments and were not tuned in any way.}
    \label{tab:user-profiles}
\end{table*}

\subsubsection{Prompts used in the pipeline}
\label{app:prompts_personas:prompts}

We release the full prompts used across all stages of the pipeline: persona parsing, prompt-change proposal, personalized prompt generation, verification of semantic preservation, and model evaluation and judging. Prompts are stored in configuration files, with model-specific wrappers (e.g., provider system and developer messages) and stage-specific user prompts. Below, we provide representative templates for the main stages. Throughout, we enforce strict output formatting to support reliable automation and downstream parsing.

\paragraph{Provider wrapper.}
For models that support developer messages, we prepend a minimal wrapper to encourage strict adherence to instructions.

{\footnotesize
\begin{quote}
\textbf{Developer message:} Follow the instructions strictly.
\end{quote}
}

For the Qwen3 models, we used the recommended default instruction with thinking enabled:

{\footnotesize
\begin{quote}
\textbf{Developer message:} You are a helpful assistant. Please first think about the question thoroughly. Consider multiple approaches and show your reasoning. Wrap your thinking in $<$think$>$ and $<$/think$>$ tags and then return your final answer.
\end{quote}
}

\paragraph{Persona parsing.}
Given a short natural-language user description, the persona-parsing prompt asks an LLM to produce a structured JSON profile describing input and output preferences. The model must output a single JSON object and nothing else (Prompt is in Figure~\ref{fig:prompt-persona-parsing}).






\paragraph{Change identification (profile $\rightarrow$ concrete prompt modifications).}
To operationalize a persona profile into actionable prompt edits, we ask an LLM to propose 2--3 concrete modification options for a fixed set of fields, while explicitly disallowing changes that alter the task itself. This stage outputs a single JSON object with a list of changes keyed by profile fields (prompt is in Figure~\ref{fig:prompt-change-identification}).






\paragraph{Personalized prompt composition.}
Given an original benchmark prompt and the selected modifications, we generate a personalized version that preserves the underlying programming task. The prompt is written in the persona voice (first person), avoids explicit references to the profile schema, and is constrained to a short length (prompt is in Figure~\ref{fig:prompt-composition}).





\paragraph{HumanEval+ prefix composition.}
For HumanEval+ style prompts that include code context and docstrings, we produce only a short persona prefix that is concatenated to the original prompt (Figure~\ref{fig:prompt-humaneval-prefix}). This avoids perturbing code formatting while still injecting persona-relevant framing.





\paragraph{Semantic-preservation verification.}
To ensure personalized prompts remain faithful to the original benchmark intent, we use a verifier prompt that checks (i) whether the end goal is identical and (ii) whether the ground-truth solution set is preserved. The verifier returns two booleans and an error string if either check fails (prompt is in Figure~\ref{fig:prompt-verification}).

\paragraph{Notes on implementation and release.}
All prompt templates above are parameterized using placeholders (e.g., \texttt{\{original\_prompt\}}) and are instantiated deterministically by the pipeline. We release the full, exact prompt texts (including system and developer wrappers), together with the JSON schemas used for validation, in the accompanying repository to support faithful reproduction.

\section{Additional results}\label{appendix:more_results}

\begin{table*}[t]
  \centering
  \setlength{\tabcolsep}{6pt}
  \renewcommand{\arraystretch}{1.15}
  \begin{tabular}{c l| c c c c}
    \toprule
     &  &
    \multicolumn{4}{c}{\textbf{Win-rate (Tie-rate)}} \\
    \cmidrule(lr){3-6}
    \textbf{Model Pair} & \textbf{Prompt Type} & \textbf{Beginner} & \textbf{Intermediate} & \textbf{AI Researcher} & \textbf{Advanced} \\
    \midrule
    \multirow{3}{*}{%
      \begin{tabular}[c]{@{}c@{}}\texttt{GPT-5.1} \\\textit{vs.} \\\texttt{GPT-OSS-20B}\end{tabular}%
    }
      & Original
        & 0.03\textsuperscript{*} (0.01) & 0.02\textsuperscript{*} (0.01) & 0.12\textsuperscript{*} (0.02) & \textbf{0.85}\textsuperscript{*} (0.00) \\
      & Personalized
        & \textbf{0.60}\textsuperscript{*} (0.05) & \textbf{0.63}\textsuperscript{*} (0.03) & 0.39\textsuperscript{*} (0.01) & \textbf{0.63}\textsuperscript{*} (0.03) \\
      & Control
        & 0.12\textsuperscript{*} (0.01) & 0.10\textsuperscript{*} (0.01) & 0.20\textsuperscript{*} (0.02) & \textbf{0.87}\textsuperscript{*} (0.01) \\
    \midrule
    \multirow{3}{*}{%
      \begin{tabular}[c]{@{}c@{}}\texttt{GPT-5.1} \\\textit{vs.} \\\texttt{GPT-4o}\end{tabular}%
    }
      & Original
        & 0.13\textsuperscript{*} (0.01) & 0.22\textsuperscript{*} (0.03) & \textbf{0.68}\textsuperscript{*} (0.03) & \textbf{0.92}\textsuperscript{*} (0.01) \\
      & Personalized
        & \textbf{0.90}\textsuperscript{*} (0.01) & \textbf{0.90}\textsuperscript{*} (0.02) & \textbf{0.85}\textsuperscript{*} (0.01) & \textbf{0.85}\textsuperscript{*} (0.02) \\
      & Control
        & 0.11\textsuperscript{*} (0.01) & 0.12\textsuperscript{*} (0.04) & \textbf{0.66}\textsuperscript{*} (0.03) & \textbf{0.92}\textsuperscript{*} (0.01) \\

    \bottomrule

  \end{tabular}
  \caption{
    \textbf{Personalization shifts win-rates (HumanEval+).}
    Per-sample win rates for each model pair are shown by persona and prompt type; win-rates are reported from the perspective of the first model in each pair. Preference shifts mirror the ones observed in MBPP+:
    GPT-5.1 underperforms on the original Beginner/Intermediate prompt, but wins on personalized prompts. GPT-5.1 remains stronger for the Advanced persona in all prompts. $^{*}$ denotes statistical significance (two-sided binomial test).
  }
  \label{tab:humaneval_results}
\end{table*}

\begin{table*}[t]
  \centering
  \setlength{\tabcolsep}{6pt}
  \renewcommand{\arraystretch}{1.15}
  \begin{tabular}{c l| c c c c}
    \toprule
     &  &
    \multicolumn{4}{c}{\textbf{Win-rate (Tie-rate)}} \\
    \cmidrule(lr){3-6}
    \textbf{Model Pair} & \textbf{Prompt Type} & \textbf{Beginner} & \textbf{Intermediate} & \textbf{AI Researcher} & \textbf{Advanced} \\
    \midrule
    \multirow{3}{*}{%
      \begin{tabular}[c]{@{}c@{}}\texttt{GPT-5.1} \\\textit{vs.} \\\texttt{GPT-OSS-20B}\end{tabular}%
    }
      & Original
        & 0.04\textsuperscript{*} (0.00) & 0.01\textsuperscript{*} (0.01) & 0.03\textsuperscript{*} (0.02) & \textbf{0.82}\textsuperscript{*} (0.00) \\
      & Personalized
        & 0.40\textsuperscript{*} (0.03) & \textbf{0.55}\textsuperscript{*} (0.05) & 0.48 (0.12) & 0.45\textsuperscript{*} (0.10) \\
      & Control
        & 0.04\textsuperscript{*} (0.00) & 0.02\textsuperscript{*} (0.01) & 0.03\textsuperscript{*} (0.02) & \textbf{0.84}\textsuperscript{*} (0.01) \\
    \midrule
    \multirow{3}{*}{%
      \begin{tabular}[c]{@{}c@{}}\texttt{Qwen3-32B} \\\textit{vs.} \\\texttt{Qwen3-14B}\end{tabular}%
    }
      & Original
        & 0.56 (0.10) & \textbf{0.58}\textsuperscript{*} (0.08) & \textbf{0.61}\textsuperscript{*} (0.09) & 0.46 (0.07) \\
      & Personalized
        & \textbf{0.64}\textsuperscript{*} (0.01) & \textbf{0.63}\textsuperscript{*} (0.07) & \textbf{0.61}\textsuperscript{*} (0.08) & 0.41 (0.17) \\
      & Control
        & 0.55 (0.04) & 0.54 (0.07) & 0.45 (0.12) & 0.44\textsuperscript{*} (0.08) \\
    \bottomrule
  \end{tabular}
  \caption{ \textbf{Personalization shifts remain similar with Qwen3-32B as generator.}
    Per-sample win rates on MBPP+ by user and prompt type, reported for the first model in each pair (tie rate in parentheses), when personalized rewrites are generated by \texttt{Qwen3-32B} instead of \texttt{GPT-5.1}. Overall trends largely align with the main results, suggesting that the observed preference shifts are not specific to any single generator. One difference is a smaller increase in win-rate for \texttt{GPT-5.1} on the Beginner and Advanced personas, possibly due to different judge sets. Judges are \texttt{Qwen3-14B} and \texttt{GPT-OSS-20B} for \texttt{GPT-5.1} \textit{vs.} \texttt{GPT-OSS-20B}, and \texttt{GPT-OSS-20B} for \texttt{Qwen3-32B} \textit{vs.} \texttt{Qwen3-14B}. \textsuperscript{*} denotes statistical significance on a two-sided binomial test.
  }
  \label{tab:pairwise-joint-preference-qwen32}
\end{table*}

\begin{table*}[t]
\centering
\setlength{\tabcolsep}{7pt}
\renewcommand{\arraystretch}{1.15}
\begin{tabular}{@{}c|c|cc@{}}
\toprule
\textbf{Grouping} & \textbf{Subset} & \textbf{Agreement (\%)} & \textbf{Fleiss's $\kappa$} \\
\midrule

\multirow{4}{*}{\textbf{Model pair}} 
& \texttt{GPT-5.1} \textit{vs.} \texttt{GPT-OSS-20B} & \pmstd{0.82}{0.14} & \pmstd{0.38}{0.21} \\
& \texttt{GPT-5.1} \textit{vs.} \texttt{GPT-4o} & \pmstd{0.85}{0.10} & \pmstd{0.38}{0.12} \\
& \texttt{Gemini-3-Pro} \textit{vs.} \texttt{Gemma-3-4B} & \pmstd{0.82}{0.07} & \pmstd{0.49}{0.13} \\
& \texttt{Qwen3-32B} \textit{vs.} \texttt{Qwen3-14B} & \pmstd{0.64}{0.07} & \pmstd{0.30}{0.12} \\
\midrule

\multirow{4}{*}{\textbf{Persona}} 
& Beginner & \pmstd{0.81}{0.13} & \pmstd{0.36}{0.10} \\
& Intermediate & \pmstd{0.78}{0.13} & \pmstd{0.45}{0.18} \\
& Researcher & \pmstd{0.74}{0.14} & \pmstd{0.28}{0.17} \\
& Advanced & \pmstd{0.79}{0.11} & \pmstd{0.45}{0.13} \\
\midrule

\multirow{3}{*}{\textbf{Prompt type}} 
& Original & \pmstd{0.79}{0.14} & \pmstd{0.43}{0.22} \\
& Control & \pmstd{0.78}{0.13} & \pmstd{0.41}{0.18} \\
& Personalized & \pmstd{0.77}{0.12} & \pmstd{0.34}{0.09} \\
\midrule

\multirow{3}{*}{\textbf{Judge pair}} 
& \texttt{GPT-5.1} \textit{vs.} \texttt{GPT-OSS-20B} & \pmstd{0.78}{0.14} & \pmstd{0.38}{0.18} \\
& \texttt{GPT-5.1} \textit{vs.} \texttt{Qwen3-14B} & \pmstd{0.79}{0.14} & \pmstd{0.40}{0.20} \\
& \texttt{Qwen3-14B} \textit{vs.} \texttt{GPT-OSS-20B} & \pmstd{0.77}{0.13} & \pmstd{0.38}{0.19} \\
\midrule
\midrule
\textbf{Overall} & All Samples &\pmstd{\textbf{0.78}}{0.13} & \pmstd{\textbf{0.39}}{0.16} \\
\bottomrule
\end{tabular}
\caption{LLM-judge agreement for the subjective pairwise preference labels, reported over multiple pooled slices of the evaluation. For each subset, we aggregate all its pairwise decisions and compute (i) \textbf{raw agreement} as the mean percentage of items on which two judges output the same label, and (ii) \textbf{Fleiss's $\kappa$}, which adjusts agreement for chance given the judges' marginal label distributions (note that when one model is selected as the winner in most items, the resulting label imbalance can lower $\kappa$ despite high raw agreement). Values are reported as \(\text{mean} \pm \text{std}\), where the mean and standard deviation are computed across the judges' pairs when available.}
\label{tab:llm_judges_agreement}
\end{table*}

\begin{table*}[t]
  \centering
  \setlength{\tabcolsep}{6pt}
  \renewcommand{\arraystretch}{1.15}
  \begin{tabular}{c l| c c c c}
    \toprule
     &  &
    \multicolumn{4}{c}{\textbf{Win-rate (Tie-rate)}} \\
    \cmidrule(lr){3-6}
    \textbf{Model Pair} & \textbf{Prompt Type} & \textbf{Beginner} & \textbf{Intermediate} & \textbf{AI Researcher} & \textbf{Advanced} \\
    \midrule
    \multirow{3}{*}{%
      \begin{tabular}[c]{@{}c@{}}\texttt{GPT-5.1} \\\textit{vs.} \\\texttt{GPT-4o}\end{tabular}%
    }
      & Original
        & 0.09\textsuperscript{*} (0.02) & 0.15\textsuperscript{*} (0.07) & \textbf{0.61}\textsuperscript{*} (0.06) & \textbf{0.87}\textsuperscript{*} (0.04) \\
      & Personalized
        & \textbf{0.92}\textsuperscript{*} (0.03) & \textbf{0.75}\textsuperscript{*} (0.07) & \textbf{0.93}\textsuperscript{*} (0.04) & \textbf{0.74}\textsuperscript{*} (0.06) \\
      & Control
        & 0.08\textsuperscript{*} (0.03) & 0.18\textsuperscript{*} (0.07) & \textbf{0.69}\textsuperscript{*} (0.06) & \textbf{0.93}\textsuperscript{*} (0.03) \\
    \midrule
    \multirow{3}{*}{%
      \begin{tabular}[c]{@{}c@{}}\texttt{GPT-5.1} \\\textit{vs.} \\\texttt{GPT-OSS-20B}\end{tabular}%
    }
      & Original
        & 0.04\textsuperscript{*} (0.02) & 0.01\textsuperscript{*} (0.02) & 0.08\textsuperscript{*} (0.05) & \textbf{0.91}\textsuperscript{*} (0.02) \\
      & Personalized
        & \textbf{0.73}\textsuperscript{*} (0.06) & 0.53 (0.10) & 0.38\textsuperscript{*} (0.09) & 0.53 (0.07) \\
      & Control
        & 0.06\textsuperscript{*} (0.04) & 0.03\textsuperscript{*} (0.02) & 0.09\textsuperscript{*} (0.08) & \textbf{0.93}\textsuperscript{*} (0.03) \\
    \midrule
    \multirow{3}{*}{%
      \begin{tabular}[c]{@{}c@{}}\texttt{Gemini-3-Pro} \\\textit{vs.} \\\texttt{Gemma-3-4B}\end{tabular}%
    }
      & Original
        & 0.45 (0.07) & 0.50 (0.08) & 0.46 (0.07) & \textbf{0.65}\textsuperscript{*} (0.03) \\
      & Personalized
        & \textbf{0.92}\textsuperscript{*} (0.02) & \textbf{0.85}\textsuperscript{*} (0.05) & \textbf{0.95}\textsuperscript{*} (0.02) & \textbf{0.56}\textsuperscript{*} (0.09) \\
      & Control
        & 0.39\textsuperscript{*} (0.07) & 0.44\textsuperscript{*} (0.06) & 0.45\textsuperscript{*} (0.08) & \textbf{0.67}\textsuperscript{*} (0.05) \\
    \midrule
    \multirow{3}{*}{%
      \begin{tabular}[c]{@{}c@{}}\texttt{Qwen3-32B} \\\textit{vs.} \\\texttt{Qwen3-14B}\end{tabular}%
    }
      & Original
        & \textbf{0.44}\textsuperscript{*} (0.13) & 0.47 (0.17) & \textbf{0.44}\textsuperscript{*} (0.16) & 0.35\textsuperscript{*} (0.09) \\
      & Personalized
        & 0.49 (0.09) & \textbf{0.60}\textsuperscript{*} (0.11) & 0.52 (0.13) & \textbf{0.46}\textsuperscript{*} (0.10) \\
      & Control
        & \textbf{0.45}\textsuperscript{*} (0.15) & 0.50 (0.14) & 0.46 (0.15) & 0.37\textsuperscript{*} (0.11) \\
    \bottomrule
  \end{tabular}
  \caption{\textbf{Similar trends with unweighted aggregation.}
    Per-sample win rates on the same MBPP+ comparisons as in the main results, but with equal weight assigned to all output dimensions. Results remain close to the main findings, suggesting that the observed preference shifts are already strong without persona-specific weighting. \textsuperscript{*} denotes statistical significance on a two-sided binomial test.
  }
  \label{tab:pairwise-joint-preference-not-weighted}
\end{table*}

\begin{table*}[t]
  \centering
  \setlength{\tabcolsep}{6pt}
  \renewcommand{\arraystretch}{1.15}
  \begin{tabular}{c l| c c c c}
    \toprule
     &  &
    \multicolumn{4}{c}{\textbf{Win-rate (Tie-rate)}} \\
    \cmidrule(lr){3-6}
    \textbf{Model Pair} & \textbf{Prompt Type} & \textbf{Beginner} & \textbf{Intermediate} & \textbf{AI Researcher} & \textbf{Advanced} \\
    \midrule
    \multirow{3}{*}{%
      \begin{tabular}[c]{@{}c@{}}\texttt{GPT-5.1} \\\textit{vs.} \\\texttt{GPT-4o}\end{tabular}%
    }
      & Original
        & 0.09\textsuperscript{*} (0.00) & 0.16\textsuperscript{*} (0.02) & \textbf{0.63}\textsuperscript{*} (0.02) & \textbf{0.88}\textsuperscript{*} (0.00) \\
      & Personalized
        & \textbf{0.94}\textsuperscript{*} (0.01) & \textbf{0.77}\textsuperscript{*} (0.02) & \textbf{0.97}\textsuperscript{*} (0.00) & \textbf{0.82}\textsuperscript{*} (0.02) \\
      & Control
        & 0.08\textsuperscript{*} (0.01) & 0.19\textsuperscript{*} (0.03) & \textbf{0.70}\textsuperscript{*} (0.02) & \textbf{0.95}\textsuperscript{*} (0.01) \\
    \midrule
    \multirow{3}{*}{%
      \begin{tabular}[c]{@{}c@{}}\texttt{GPT-5.1} \\\textit{vs.} \\\texttt{GPT-OSS-20B}\end{tabular}%
    }
      & Original
        & 0.03\textsuperscript{*} (0.00) & 0.01\textsuperscript{*} (0.00) & 0.08\textsuperscript{*} (0.02) & \textbf{0.91}\textsuperscript{*} (0.01) \\
      & Personalized
        & \textbf{0.77}\textsuperscript{*} (0.02) & \textbf{0.55}\textsuperscript{*} (0.02) & 0.43\textsuperscript{*} (0.01) & \textbf{0.58}\textsuperscript{*} (0.03) \\
      & Control
        & 0.06\textsuperscript{*} (0.00) & 0.02\textsuperscript{*} (0.00) & 0.10\textsuperscript{*} (0.02) & \textbf{0.93}\textsuperscript{*} (0.01) \\
    \midrule
    \multirow{3}{*}{%
      \begin{tabular}[c]{@{}c@{}}\texttt{Gemini-3-Pro} \\\textit{vs.} \\\texttt{Gemma-3-4B}\end{tabular}%
    }
      & Original
        & 0.48 (0.02) & 0.53 (0.01) & 0.48 (0.03) & \textbf{0.66}\textsuperscript{*} (0.02) \\
      & Personalized
        & \textbf{0.93}\textsuperscript{*} (0.01) & \textbf{0.90}\textsuperscript{*} (0.01) & \textbf{0.97}\textsuperscript{*} (0.01) & \textbf{0.70}\textsuperscript{*} (0.02) \\
      & Control
        & 0.44\textsuperscript{*} (0.02) & 0.45\textsuperscript{*} (0.02) & 0.48 (0.02) & \textbf{0.69}\textsuperscript{*} (0.02) \\
    \midrule
    \multirow{3}{*}{%
      \begin{tabular}[c]{@{}c@{}}\texttt{Qwen3-32B} \\\textit{vs.} \\\texttt{Qwen3-14B}\end{tabular}%
    }
      & Original
        & 0.54 (0.04) & \textbf{0.62}\textsuperscript{*} (0.03) & 0.53 (0.04) & 0.38\textsuperscript{*} (0.05) \\
      & Personalized
        & 0.54 (0.02) & \textbf{0.67}\textsuperscript{*} (0.04) & \textbf{0.62}\textsuperscript{*} (0.03) & \textbf{0.54}\textsuperscript{*} (0.04) \\
      & Control
        & 0.54 (0.05) & \textbf{0.59}\textsuperscript{*} (0.04) & \textbf{0.55}\textsuperscript{*} (0.06) & 0.40\textsuperscript{*} (0.05) \\
    \bottomrule
  \end{tabular}
  \caption{\textbf{Similar trends when correctness is not used to determine the sample-level winner.}
    Per-sample win rates on the same MBPP+ comparisons as in the main results, but with winners determined only from dimension-level judgments, ignoring correctness. Results remain similar, suggesting that the main preference patterns are not driven primarily by the correctness gate. \textsuperscript{*} denotes statistical significance on a two-sided binomial test.
  }
  \label{tab:pairwise-joint-preference-no-correctness-gate}
\end{table*}

\begin{table*}[t]
  \centering
  \setlength{\tabcolsep}{6pt}
  \renewcommand{\arraystretch}{1.15}
  \begin{tabular}{c l| c c c c}
    \toprule
     &  &
    \multicolumn{4}{c}{\textbf{Win-rate (Tie-rate)}} \\
    \cmidrule(lr){3-6}
    \textbf{Model Pair} & \textbf{Prompt Type} & \textbf{Beginner} & \textbf{Intermediate} & \textbf{AI Researcher} & \textbf{Advanced} \\
    \midrule
    \multirow{3}{*}{%
      \begin{tabular}[c]{@{}c@{}}\texttt{GPT-5.1} \\\textit{vs.} \\\texttt{GPT-4o}\end{tabular}%
    }
      & Original
        & 0.08\textsuperscript{*} (0.01) & 0.16\textsuperscript{*} (0.03) & \textbf{0.61}\textsuperscript{*} (0.04) & \textbf{0.86}\textsuperscript{*} (0.01) \\
      & Personalized
        & \textbf{0.93}\textsuperscript{*} (0.01) & \textbf{0.75}\textsuperscript{*} (0.03) & \textbf{0.95}\textsuperscript{*} (0.01) & \textbf{0.82}\textsuperscript{*} (0.03) \\
      & Control
        & 0.07\textsuperscript{*} (0.01) & 0.17\textsuperscript{*} (0.04) & \textbf{0.68}\textsuperscript{*} (0.04) & \textbf{0.94}\textsuperscript{*} (0.01) \\
    \midrule
    \multirow{3}{*}{%
      \begin{tabular}[c]{@{}c@{}}\texttt{GPT-5.1} \\\textit{vs.} \\\texttt{GPT-OSS-20B}\end{tabular}%
    }
      & Original
        & 0.04\textsuperscript{*} (0.00) & 0.01\textsuperscript{*} (0.00) & 0.07\textsuperscript{*} (0.06) & \textbf{0.92}\textsuperscript{*} (0.01) \\
      & Personalized
        & \textbf{0.76}\textsuperscript{*} (0.03) & \textbf{0.57}\textsuperscript{*} (0.04) & 0.37\textsuperscript{*} (0.03) & \textbf{0.56}\textsuperscript{*} (0.04) \\
      & Control
        & 0.06\textsuperscript{*} (0.01) & 0.02\textsuperscript{*} (0.01) & 0.10\textsuperscript{*} (0.07) & \textbf{0.94}\textsuperscript{*} (0.01) \\
    \midrule
    \multirow{3}{*}{%
      \begin{tabular}[c]{@{}c@{}}\texttt{Gemini-3-Pro} \\\textit{vs.} \\\texttt{Gemma-3-4B}\end{tabular}%
    }
      & Original
        & 0.48 (0.03) & 0.53 (0.02) & 0.49 (0.03) & \textbf{0.68}\textsuperscript{*} (0.02) \\
      & Personalized
        & \textbf{0.93}\textsuperscript{*} (0.01) & \textbf{0.88}\textsuperscript{*} (0.02) & \textbf{0.97}\textsuperscript{*} (0.01) & \textbf{0.70}\textsuperscript{*} (0.02) \\
      & Control
        & 0.44\textsuperscript{*} (0.03) & 0.45\textsuperscript{*} (0.03) & 0.47 (0.04) & \textbf{0.69}\textsuperscript{*} (0.04) \\
    \midrule
    \multirow{3}{*}{%
      \begin{tabular}[c]{@{}c@{}}\texttt{Qwen3-32B} \\\textit{vs.} \\\texttt{Qwen3-14B}\end{tabular}%
    }
      & Original
        & 0.48 (0.10) & \textbf{0.56}\textsuperscript{*} (0.09) & \textbf{0.42}\textsuperscript{*} (0.20) & 0.35\textsuperscript{*} (0.18) \\
      & Personalized
        & 0.53 (0.05) & \textbf{0.68}\textsuperscript{*} (0.08) & \textbf{0.61}\textsuperscript{*} (0.04) & 0.53 (0.08) \\
      & Control
        & 0.50 (0.10) & 0.52 (0.12) & \textbf{0.42}\textsuperscript{*} (0.17) & 0.36\textsuperscript{*} (0.18) \\
    \bottomrule
  \end{tabular}
  \caption{\textbf{Similar results without confidence-based tie breaking.}
    Per-sample win rates on the same MBPP+ comparisons as in the main results, but treating swapped-order disagreements as ties rather than resolving them with judge confidence. Trends remain similar, indicating that position effects have only a small impact on the overall results. \textsuperscript{*} denotes statistical significance on a two-sided binomial test.
  }
  \label{tab:pairwise-joint-preference-strict-tie-breaker}
\end{table*}

We report additional results to test the robustness of the main findings along three axes: a second benchmark (HumanEval+), a different personalized prompt generator, and alternative aggregation rules. Figures~\ref{fig:head_to_head_results_gpt4o}, \ref{fig:head_to_head_results_gemini}, and \ref{fig:head_to_head_results_qwen} show the same per-dimension breakdown across personas as Figure~\ref{fig:head_to_head_results}, for the remaining three model pairs on the main results.
Across all settings, the main qualitative pattern remains the same: personalized rewrites often change model preferences, while neutral control paraphrases largely preserve the original pattern.

We report additional results to test the robustness of the main findings along three axes: a second benchmark (HumanEval+), a different personalized prompt generator, and alternative aggregation rules. Across all settings, the main qualitative pattern remains the same: personalized rewrites often change model preferences, while neutral control paraphrases largely preserve the original pattern. 

\paragraph{HumanEval+ shows the same qualitative trend.}
Table~\ref{tab:humaneval_results} reports head-to-head results on HumanEval+ for two model pairs, \texttt{GPT-5.1} \textit{vs.} \texttt{GPT-4o} and \texttt{GPT-5.1} \textit{vs.} \texttt{GPT-OSS-20B}, under the same evaluation protocol used in the main experiments. The results broadly mirror MBPP+. For \texttt{GPT-5.1} \textit{vs.} \texttt{GPT-4o}, \texttt{GPT-5.1} underperforms on original prompts for the Beginner and Intermediate personas, but becomes strongly preferred under personalized prompts across all personas. Control paraphrases remain close to the original pattern. For \texttt{GPT-5.1} \textit{vs.} \texttt{GPT-OSS-20B}, \texttt{GPT-5.1} is strongly disfavored on original prompts for the Beginner, Intermediate, and Researcher personas, but personalized rewrites substantially improve its win rate, making the comparison much more balanced for Beginner, Intermediate, and Advanced. Due to inference cost, HumanEval+ was restricted to these two model pairs, so we treat it as a supporting robustness check rather than a comprehensive replication across all model families. 

\paragraph{Using Qwen3-32B as the prompt generator yields mostly similar shifts.}
Table~\ref{tab:pairwise-joint-preference-qwen32} repeats the MBPP+ analysis with personalized rewrites generated by \texttt{Qwen3-32B} instead of \texttt{GPT-5.1}. The overall trends remain similar to the main results. For \texttt{GPT-5.1} \textit{vs.} \texttt{GPT-OSS-20B}, personalization still substantially improves \texttt{GPT-5.1}'s standing relative to original and control prompts, especially for Intermediate, and shifts \texttt{Qwen3-32B} \textit{vs.} \texttt{Qwen3-14B} preferences toward the larger model except for the Advanced persona. One difference is that the personalized advantage for \texttt{GPT-5.1} over \texttt{GPT-OSS-20B} is weaker for the Beginner and Advanced personas, where win rates fall below $0.50$. A likely reason is that, unlike the main setup, this study does not include \texttt{GPT-5.1} as a judge due to cost; in the main results, \texttt{GPT-5.1} judgments were more favorable to \texttt{GPT-5.1} on these personas. This suggests that the size of the shift is somewhat sensitive to the judge set, even though its overall direction remains similar.

\paragraph{Unweighted aggregation yields similar results.}
Table~\ref{tab:pairwise-joint-preference-not-weighted} reports results when all output dimensions are given equal weight, rather than persona-specific importance weights. The main qualitative trends remain intact. Personalized prompts still strongly improve \texttt{GPT-5.1}'s win rate against both \texttt{GPT-4o} and \texttt{GPT-OSS-20B}, and similarly strengthen \texttt{Gemini-3-Pro} relative to \texttt{Gemma-3-4B}. For \texttt{Qwen3-32B} \textit{vs.} \texttt{Qwen3-14B}, the shifts remain smaller but generally point in the same direction. This suggests that the preference changes we observe are already strong at the level of the underlying dimension judgments, and do not depend critically on persona-specific weighting, though weighting may still matter in closer comparisons or for users with more extreme priorities. 

\paragraph{Ignoring correctness also yields similar trends.}
Table~\ref{tab:pairwise-joint-preference-no-correctness-gate} reports results when sample-level winners are determined directly from dimension judgments, without first using correctness as a gate. Again, the main patterns remain similar. Personalized prompts continue to sharply improve \texttt{GPT-5.1}'s standing against \texttt{GPT-4o} and \texttt{GPT-OSS-20B}, and strongly favor \texttt{Gemini-3-Pro} over \texttt{Gemma-3-4B}. For \texttt{Qwen3-32B} \textit{vs.} \texttt{Qwen3-14B}, personalization still shifts preferences toward the larger model for some personas, while leaving others relatively balanced. The similarity to the main results suggests that the observed preference shifts are not driven primarily by the correctness gate, either because most samples are solved correctly or because models that better fit the user also tend to perform better overall. 

\paragraph{Treating jointly correct samples as ties.}

To test whether the observed preference shifts are driven by samples where both models are correct, we repeat the analysis while assigning a tie whenever both models pass the benchmark tests. The remaining samples, where at most one model passes, are evaluated using the same persona-conditioned preference procedure.
As expected, this produces high tie rates because both models solve many samples (Table~\ref{tab:pairwise-joint-preference-both-models-correct-is-tie}). Personalized prompts still reduce ties and shift win rates for the GPT and Qwen comparisons. For \texttt{Gemini-3-Pro} vs.\ \texttt{Gemma-3-4B}, Gemini's win rate also increases from approximately 0.13--0.14 on original prompts to 0.23--0.41 on personalized prompts, but the comparison remains dominated by ties and the shift is weaker than in the main results. The control therefore shows that the personalization effect extends beyond jointly correct samples in several matchups, although it does not appear uniformly across models.

\paragraph{Strict tie handling has little effect.}
Table~\ref{tab:pairwise-joint-preference-strict-tie-breaker} reports results when disagreements between swapped response orders are treated as ties, rather than resolved with judge confidence. The resulting win rates remain close to the main results. Personalized prompts still reverse or substantially shift the original preference pattern in the \texttt{GPT-5.1} and \texttt{Gemini-3-Pro} comparisons, while the \texttt{Qwen} comparison remains weaker and more balanced. This suggests that the main findings are not driven by the confidence-based soft tie-breaker and that residual position effects have only a limited impact on the overall conclusions. 

\begin{table*}[t]
  \centering
  \setlength{\tabcolsep}{6pt}
  \renewcommand{\arraystretch}{1.15}
  \begin{tabular}{c l| c c c c}
    \toprule
     &  &
    \multicolumn{4}{c}{\textbf{Win-rate (Tie-rate)}} \\
    \cmidrule(lr){3-6}
    \textbf{Model Pair} & \textbf{Prompt Type} & \textbf{Beginner} & \textbf{Intermediate} & \textbf{AI Researcher} & \textbf{Advanced} \\
    \midrule
    \multirow{3}{*}{%
      \begin{tabular}[c]{@{}c@{}}\texttt{GPT-5.1} \\\textit{vs.} \\\texttt{GPT-4o}\end{tabular}%
    }
      & Original
        & 0.09\textsuperscript{*} (0.00) & 0.16\textsuperscript{*} (0.02) & \textbf{0.72}\textsuperscript{*} (0.02) & \textbf{0.91}\textsuperscript{*} (0.00) \\
      & Personalized
        & \textbf{0.96}\textsuperscript{*} (0.00) & \textbf{0.78}\textsuperscript{*} (0.02) & \textbf{0.98}\textsuperscript{*} (0.00) & \textbf{0.84}\textsuperscript{*} (0.02) \\
      & Control
        & 0.06\textsuperscript{*} (0.01) & 0.15\textsuperscript{*} (0.02) & \textbf{0.79}\textsuperscript{*} (0.02) & \textbf{0.94}\textsuperscript{*} (0.01) \\
    \midrule
    \multirow{3}{*}{%
      \begin{tabular}[c]{@{}c@{}}\texttt{GPT-5.1} \\\textit{vs.} \\\texttt{GPT-OSS-20B}\end{tabular}%
    }
      & Original
        & 0.05\textsuperscript{*} (0.00) & 0.04\textsuperscript{*} (0.00) & 0.06\textsuperscript{*} (0.00) & \textbf{0.94}\textsuperscript{*} (0.00) \\
      & Personalized
        & \textbf{0.80}\textsuperscript{*} (0.06) & 0.55 (0.01) & 0.32\textsuperscript{*} (0.15) & \textbf{0.58}\textsuperscript{*} (0.04) \\
      & Control
        & 0.04\textsuperscript{*} (0.00) & 0.02\textsuperscript{*} (0.00) & 0.05\textsuperscript{*} (0.01) & \textbf{0.94}\textsuperscript{*} (0.02) \\
    \midrule
    \multirow{3}{*}{%
      \begin{tabular}[c]{@{}c@{}}\texttt{Gemini-3-Pro} \\\textit{vs.} \\\texttt{Gemma-3-4B}\end{tabular}%
    }
      & Original
        & 0.48 (0.01) & 0.58 (0.00) & 0.52 (0.00) & \textbf{0.70}\textsuperscript{*} (0.02) \\
      & Personalized
        & \textbf{0.94}\textsuperscript{*} (0.01) & \textbf{0.93}\textsuperscript{*} (0.01) & \textbf{0.99}\textsuperscript{*} (0.00) & \textbf{0.72}\textsuperscript{*} (0.02) \\
      & Control
        & 0.42\textsuperscript{*} (0.02) & 0.47 (0.01) & 0.48 (0.01) & \textbf{0.69}\textsuperscript{*} (0.01) \\
    \midrule
    \multirow{3}{*}{%
      \begin{tabular}[c]{@{}c@{}}\texttt{Qwen3-32B} \\\textit{vs.} \\\texttt{Qwen3-14B}\end{tabular}%
    }
      & Original
        & 0.59 (0.02) & \textbf{0.64}\textsuperscript{*} (0.03) & 0.54 (0.02) & 0.40 (0.04) \\
      & Personalized
        & 0.54 (0.01) & \textbf{0.70}\textsuperscript{*} (0.03) & \textbf{0.65}\textsuperscript{*} (0.01) & 0.56 (0.05) \\
      & Control
        & 0.53 (0.05) & \textbf{0.62}\textsuperscript{*} (0.04) & 0.54 (0.06) & 0.43 (0.04) \\
    \bottomrule
  \end{tabular}
  \caption{\textbf{Similar results with majority-vote aggregation across judges.} Per-sample win rates on the same MBPP+ comparisons as in the main results, but using a single majority-vote label per sample instead of counting each judge's vote separately. Trends remain similar, indicating that the main findings are not driven by the original judge aggregation rule. \textsuperscript{*} denotes statistical significance on a two-sided binomial test.}
  \label{tab:pairwise-joint-preference-sample-majority}
\end{table*}

\begin{table*}[t]
  \centering
  \setlength{\tabcolsep}{6pt}
  \renewcommand{\arraystretch}{1.15}
  \begin{tabular}{c l| c c c c}
    \toprule
     &  &
    \multicolumn{4}{c}{\textbf{Win-rate (Tie-rate)}} \\
    \cmidrule(lr){3-6}
    \textbf{Model Pair} & \textbf{Prompt Type} & \textbf{Beginner} & \textbf{Intermediate} & \textbf{AI Researcher} & \textbf{Advanced} \\
    \midrule
    \multirow{3}{*}{%
      \begin{tabular}[c]{@{}c@{}}\texttt{GPT-5.1} \\\textit{vs.} \\\texttt{GPT-4o}\end{tabular}%
    }
      & Original
        & 0.14\textsuperscript{*} (0.00) & 0.21\textsuperscript{*} (0.01) & \textbf{0.60}\textsuperscript{*} (0.01) & \textbf{0.87}\textsuperscript{*} (0.00) \\
      & Personalized
        & \textbf{0.93}\textsuperscript{*} (0.01) & \textbf{0.76}\textsuperscript{*} (0.01) & \textbf{0.96}\textsuperscript{*} (0.00) & \textbf{0.75}\textsuperscript{*} (0.03) \\
      & Control
        & 0.09\textsuperscript{*} (0.01) & 0.22\textsuperscript{*} (0.04) & \textbf{0.65}\textsuperscript{*} (0.03) & \textbf{0.94}\textsuperscript{*} (0.00) \\
    \midrule
    \multirow{3}{*}{%
      \begin{tabular}[c]{@{}c@{}}\texttt{GPT-5.1} \\\textit{vs.} \\\texttt{GPT-OSS-20B}\end{tabular}%
    }
      & Original
        & 0.07\textsuperscript{*} (0.00) & 0.04\textsuperscript{*} (0.00) & 0.04\textsuperscript{*} (0.00) & \textbf{0.88}\textsuperscript{*} (0.02) \\
      & Personalized
        & \textbf{0.72}\textsuperscript{*} (0.04) & 0.53 (0.03) & 0.20\textsuperscript{*} (0.03) & 0.40\textsuperscript{*} (0.04) \\
      & Control
        & 0.09\textsuperscript{*} (0.00) & 0.02\textsuperscript{*} (0.00) & 0.02\textsuperscript{*} (0.00) & \textbf{0.90}\textsuperscript{*} (0.01) \\
    \midrule
    \multirow{3}{*}{%
      \begin{tabular}[c]{@{}c@{}}\texttt{Qwen3-32B} \\\textit{vs.} \\\texttt{Qwen3-14B}\end{tabular}%
    }
      & Original
        & 0.55 (0.02) & \textbf{0.62}\textsuperscript{*} (0.03) & 0.49 (0.03) & 0.40\textsuperscript{*} (0.02) \\
      & Personalized
        & \textbf{0.56}\textsuperscript{*} (0.01) & \textbf{0.66}\textsuperscript{*} (0.04) & \textbf{0.60}\textsuperscript{*} (0.03) & 0.52 (0.04) \\
      & Control
        & \textbf{0.56}\textsuperscript{*} (0.05) & \textbf{0.60}\textsuperscript{*} (0.03) & 0.53 (0.06) & 0.41\textsuperscript{*} (0.03) \\
    \bottomrule
  \end{tabular}
    \caption{\textbf{Similar results with disjoint judges only.}
    Per-sample win rates on the same MBPP+ comparisons as in the main results, but excluding votes from any judge that is also one of the compared models. Trends remain mostly similar, with the main difference being a lower win rate for the Advanced persona ($0.40$), suggesting limited sensitivity to the judge set. \textsuperscript{*} denotes statistical significance on a two-sided binomial test.}
      \label{tab:pairwise-joint-preference-disjoint-judges}
\end{table*}

\begin{table*}[t]
  \centering
  \setlength{\tabcolsep}{6pt}
  \renewcommand{\arraystretch}{1.15}
  \begin{tabular}{c l| c c c c}
    \toprule
     &  &
    \multicolumn{4}{c}{\textbf{Win-rate (Tie-rate)}} \\
    \cmidrule(lr){3-6}
    \textbf{Model Pair} & \textbf{Prompt Type} & \textbf{Beginner} & \textbf{Intermediate} & \textbf{AI Researcher} & \textbf{Advanced} \\
    \midrule
    \multirow{3}{*}{%
      \begin{tabular}[c]{@{}c@{}}\texttt{GPT-5.1} \\\textit{vs.} \\\texttt{GPT-4o}\end{tabular}%
    }
      & Original
        & 0.02\textsuperscript{*} (0.93) & 0.02\textsuperscript{*} (0.93) & 0.03\textsuperscript{*} (0.93) & 0.04 (0.93) \\
      & Personalized
        & \textbf{0.15}\textsuperscript{*} (0.84) & \textbf{0.20}\textsuperscript{*} (0.73) & \textbf{0.31}\textsuperscript{*} (0.68) & \textbf{0.14}\textsuperscript{*} (0.84) \\
      & Control
        & 0.02\textsuperscript{*} (0.94) & 0.02\textsuperscript{*} (0.94) & \textbf{0.04}\textsuperscript{*} (0.94) & \textbf{0.04}\textsuperscript{*} (0.94) \\
    \midrule
    \multirow{3}{*}{%
      \begin{tabular}[c]{@{}c@{}}\texttt{GPT-5.1} \\\textit{vs.} \\\texttt{GPT-OSS-20B}\end{tabular}%
    }
      & Original
        & 0.02\textsuperscript{*} (0.94) & 0.01\textsuperscript{*} (0.94) & 0.01\textsuperscript{*} (0.94) & 0.03 (0.94) \\
      & Personalized
        & \textbf{0.10}\textsuperscript{*} (0.86) & 0.15\textsuperscript{*} (0.68) & 0.18\textsuperscript{*} (0.63) & \textbf{0.11}\textsuperscript{*} (0.85) \\
      & Control
        & 0.02\textsuperscript{*} (0.95) & 0.01\textsuperscript{*} (0.95) & 0.01\textsuperscript{*} (0.95) & \textbf{0.03}\textsuperscript{*} (0.95) \\
    \midrule
    \multirow{3}{*}{%
      \begin{tabular}[c]{@{}c@{}}\texttt{Gemini-3-Pro} \\\textit{vs.} \\\texttt{Gemma-3-4B}\end{tabular}%
    }
        & Original
        & 0.13\textsuperscript{*} (0.82) & 0.14\textsuperscript{*} (0.82) & 0.14\textsuperscript{*} (0.82) & 0.14\textsuperscript{*} (0.83) \\
        & Personalized
        & \textbf{0.23}\textsuperscript{*} (0.75) & \textbf{0.30}\textsuperscript{*} (0.66) & \textbf{0.41}\textsuperscript{*} (0.58) & \textbf{0.26}\textsuperscript{*} (0.68) \\
        & Control
        & 0.13\textsuperscript{*} (0.82) & 0.14\textsuperscript{*} (0.82) & 0.14\textsuperscript{*} (0.82) & 0.14\textsuperscript{*} (0.82) \\
    \midrule
    \multirow{3}{*}{%
      \begin{tabular}[c]{@{}c@{}}\texttt{Qwen3-32B} \\\textit{vs.} \\\texttt{Qwen3-14B}\end{tabular}%
    }
      & Original
        & 0.03\textsuperscript{*} (0.96) & 0.04\textsuperscript{*} (0.95) & 0.03\textsuperscript{*} (0.95) & 0.03\textsuperscript{*} (0.96) \\
      & Personalized
        & \textbf{0.10}\textsuperscript{*} (0.80) & \textbf{0.33}\textsuperscript{*} (0.53) & \textbf{0.25}\textsuperscript{*} (0.60) & \textbf{0.10}\textsuperscript{*} (0.83) \\
      & Control
        & 0.03\textsuperscript{*} (0.94) & 0.03\textsuperscript{*} (0.94) & 0.02\textsuperscript{*} (0.94) & 0.03\textsuperscript{*} (0.94) \\
    \bottomrule
  \end{tabular}
  \caption{\textbf{Similar results when jointly correct samples are treated as ties.} Per-sample win rates on the same MBPP+ comparisons as in the main results, but samples solved by both models are counted as ties. Personalized prompts still shift outcomes for the GPT and Qwen comparisons, while \texttt{Gemini-3-Pro} vs.\ \texttt{Gemma-3-4B} remains tie-dominated and shows a weaker effect. Therefore, this suggests that the observed preference shifts are mostly not driven by cases where both models are correct. \textsuperscript{*} denotes statistical significance on a two-sided binomial test.}\label{tab:pairwise-joint-preference-both-models-correct-is-tie}
\end{table*}

\paragraph{Restricting to disjoint judges preserves the main pattern.}
To reduce possible self-preference in model judging, we recompute the results using only disjoint judges, excluding votes from judges that are also one of the compared models. The overall trends remain similar to the main results (Table~\ref{tab:pairwise-joint-preference-disjoint-judges}).
The main exception is the Advanced persona, where the win rate falls to $0.40$. This indicates some sensitivity to the judge set, and suggests that majority voting or larger disjoint judge pools would improve robustness.

\paragraph{Majority-vote aggregation also preserves the main pattern.}
Instead of counting each judge vote separately, we also compute one majority-vote label per sample across judges, with mixed cases resolved conservatively toward ties.
The resulting trends are again similar to the main results (Table~\ref{tab:pairwise-joint-preference-sample-majority}). This suggests that the overall findings are stable to the choice of judge aggregation.

Taken together, these additional analyses strengthen the main conclusion of the paper. The preference shifts induced by personalized prompts generalize beyond MBPP+ to HumanEval+, remain visible when using a different prompt generator, and are robust to several alternative aggregation choices. Across all these checks, the same broad pattern remains: personalization can materially change which model is preferred, whereas neutral paraphrases usually do not.

\subsection{Data-Derived User Profiles}\label{appendix:real_personas}

We use our Persona Creator that converts user information, such as public profiles, posts, or direct descriptions, into the structured profile format used by our pipeline.
To provide a limited grounding check for our controlled archetypes, we apply it to two users represented in our in-the-wild corpus: a beginner with public posts in \texttt{r/learnpython} and the advanced coding YouTube channel \emph{Fireship}. We extract only coding-related task and response preferences and manually review both profiles before use.

The Reddit-derived profile largely matches the controlled Beginner archetype (Table~\ref{tab:data-derived-personas}). The YouTube-derived profile reflects advanced coding expertise while also prioritizing simple, tutorial-style explanations for its audience. It therefore combines preferences associated with both advanced and beginner users.

We evaluate \texttt{GPT-5.1} against \texttt{GPT-OSS-20B} on 25 samples. For the Reddit-derived beginner profile, the \texttt{GPT-5.1} win rate increases from 0.04 on original prompts to 0.68 on personalized prompts, following the same direction as the controlled Beginner archetype, which increases from 0.03 to 0.77. For the YouTube-derived advanced profile, the win rate increases from 0.28 to 0.53. Its lower original win rate compared with the controlled Advanced archetype, which changes from 0.91 to 0.58, likely reflects its preference for simple tutorial explanations.

\begin{table}[t]
  \centering
  \setlength{\tabcolsep}{8pt}
  \renewcommand{\arraystretch}{1.15}
  \begin{tabular}{l|cc}
    \toprule
    & \multicolumn{2}{c}{\textbf{Win-rate}} \\
    \cmidrule(lr){2-3}
    \textbf{User Profile} & \textbf{Original} & \textbf{Personalized} \\
    \midrule
    Beginner Student   & 0.03 & \textbf{0.77} \\
    Advanced Developer & 0.91 & \textbf{0.58} \\
    \midrule
    Reddit Beginner    & 0.04 & \textbf{0.68} \\
    YouTube Advanced   & 0.28 & \textbf{0.53} \\
    \bottomrule
  \end{tabular}
  \caption{\textbf{Data-derived profiles show similar directional preference shifts.}
  Win rates for \texttt{GPT-5.1} against \texttt{GPT-OSS-20B} on 25 MBPP+ samples, reported for the controlled archetypes and corresponding data-derived profiles. Personalized prompts increase \texttt{GPT-5.1}'s win rate for both beginner profiles and shift both advanced profiles in the same direction.}\label{tab:data-derived-personas}
\end{table}

These results show similar directional personalization shifts for profiles derived from public user information.
Although limited to two coarse profiles, this experiment provides initial evidence that the pipeline can be extended to real user data and motivates broader validation at both the population and individual levels.

\subsection{Minimal-Personalization Control}\label{appendix:short_personalization}

Personalization often adds user context, constraints, and preferences, which naturally increases prompt length. To test whether length alone explains the observed preference shifts, we create minimal-personalization variants that add only one to three persona-relevant words to each prompt.
For example, ``Write a function...'' becomes ``Write an \textbf{efficient} function...'' for the Advanced persona and ``Write a \textbf{simple} function...'' for the Beginner persona. These variants are also mostly shorter than the neutral paraphrase controls.

We evaluate \texttt{GPT-5.1} against \texttt{GPT-OSS-20B} under four prompt conditions. For the Beginner persona, the \texttt{GPT-5.1} win rate increases from 0.03 on original prompts and 0.06 on control prompts to 0.16 under minimal personalization, compared with 0.77 under full personalization. For the Advanced persona, it decreases from 0.91 on original prompts and 0.93 on control prompts to 0.84 with minimal personalization, approaching the full-personalization rate of 0.58.

For both personas, minimal personalization shifts preferences away from the original and control conditions and toward full personalization.
The smaller shifts are expected because the intervention captures only a small part of each profile.
Full personalization still combines several factors, including context, constraints, specificity, and requested response style.
These results suggest that increased prompt length alone does not explain the main effect.

\section{LLM judge agreement analysis}\label{appendix:judge_agreements}

To evaluate the reliability of the automated pairwise evaluation, we measure agreement between LLM judges on the final per-sample preference labels. Unless stated otherwise, we use \texttt{GPT-5.1}, \texttt{GPT-OSS-20B}, and \texttt{Qwen3-14B} as judges, with \texttt{GPT-5.1} omitted for Gemini and Qwen comparisons due to cost.

For each evaluated sample, each judge produces a final pairwise label for the preferred model (\textsc{A wins}/\textsc{B wins}/\textsc{Tie}) after resolving swapped-order comparisons as described in Section~\ref{sec:experiments_and_results}.
We then compare judges on these final labels using two agreement measures: (1) raw agreement, the percentage of samples on which judges assign the same label, and (2) Fleiss's $\kappa$, which adjusts for chance agreement given the marginal label distribution (equivalent to Cohen's  $\kappa$ for two judges).
To summarize agreement over a subset, we first split it into finer-grained conditions, compute judge-pair agreement within each condition, reduce each condition to a single mean agreement score, and then report the mean and standard deviation across conditions, weighted by the number of items in each condition. Since some subsets are label-imbalanced, for example when one model is preferred on most samples, $\kappa$ can be lower even when raw agreement is relatively high.

\subsection{LLM agreement results}

Table~\ref{tab:llm_judges_agreement} reports agreement across several pooled slices of the evaluation: by model pair, persona, prompt type, and judge pair. For each slice, we summarize condition-level agreement scores computed from the final preference labels. Values are reported as mean $\pm$ standard deviation across conditions, with each condition weighted by its number of items.

Overall, LLM judges show reasonably consistent preferences, with mean raw agreement of $78\% \pm 0.13$ and Fleiss's $\kappa = 0.39 \pm 0.16$. Agreement is fairly stable across most model pairs, personas, and prompt types. By model pair, agreement is highest for \texttt{GPT-5.1} \textit{vs.} \texttt{GPT-4o} and Gemini-3-Pro \textit{vs.} Gemma-3-4B, and lowest for \texttt{Qwen3-32B} \textit{vs.} \texttt{Qwen3-14B}, suggesting that this comparison is harder to judge consistently. Across personas, agreement is somewhat lower for the Researcher persona than for the others. Across prompt types, original, control, and personalized prompts yield broadly similar agreement, with only a small drop for personalized prompts. Finally, agreement is also similar across judge pairs, indicating that no single judge pair is driving the overall pattern.

These results support the use of LLM judges for subjective pairwise comparisons in our pipeline and show that agreement varies across subsets, with lower levels in some harder settings.

\section{Human validation of automated judgments}\label{appendix:human_validation}

To validate whether our automated LLM-based pairwise judgments align with human assessments, we conducted a human preference annotation study. Annotators were presented with pairs of coding-assistant responses and asked to judge which response better fit a given user persona, both overall and across multiple quality dimensions. The study was designed to directly validate the persona-conditioned pairwise evaluation used in our pipeline.

\subsection{Study setting}

The study was grounded in the same evaluation framework used in the main experiments. We considered two prompt types: \emph{original prompts}, taken directly from coding benchmarks, and \emph{personalized prompts}, which rewrite the same tasks to reflect a specific user persona. We used two personas throughout the study: a \emph{Novice User}, who values clarity and detailed explanations, and an \emph{Advanced Developer}, who prefers concise, technical, high-signal responses. Each item compared the outputs of one of two model pairs: GPT-4o vs.\ GPT-5.1, and Gemini-3-Pro-Preview vs.\ Gemma-3-4B-IT. Model identities were hidden from annotators.

\paragraph{Pre-selection of items.}
Before the human study, candidate items were judged automatically by three LLM judges: GPT-5.1, GPT-OSS-20B, and Qwen3-14B. Judges compared model outputs from the perspective of a given persona and selected an overall winner, as well as winners for individual quality dimensions. Starting from a pool of 1,196 source items, filtering produced 667 items eligible for sampling. Filtering excluded items with overly long responses, unanimous overall ties, or insufficient judge agreement on the overall winner. Full rejection counts and filtering details are described in the study manifest files.

\paragraph{Sampling and design.}
The final study followed a 2 x 2 x 2 factorial design over persona, prompt type, and model pair. This yielded 8 condition cells. We sampled 32 unique items in total, using approximately balanced allocation across conditions. Two items were designated as calibration items and shown to all annotators; the remaining regular items were each assigned to exactly two annotators. Sampling used a fixed random seed.

\subsection{Annotators and assignment}

Six annotators participated in the study. Each annotator completed 12 items: 10 regular items and 2 shared calibration items. The resulting dataset contains 72 total annotation tasks, including 60 regular assignments and 12 calibration assignments. Regular-item assignment was balanced so that each annotator saw both personas, both model pairs, and both prompt types.

\paragraph{Annotation interface and questions.}
For each item, annotators were shown three elements: (1) a natural-language persona description, (2) the coding prompt, either original or personalized, and (3) two responses labeled A and B. Response order was randomized for half of the assignments to reduce position bias, and results were later mapped back to the canonical model order. Annotators selected \emph{dimension-level preferences} between Response A / Tie / Responses B for seven evaluated dimensions: Clarity, Tone/Style Fit, Workflow Fit, Cognitive Load, Context Awareness, Persona Consistency, and Anthropomorphism. Finally, annotators reported overall response preferences and a confidence level of Low, Medium, or High. No free-text rationale was collected. Importantly, annotators were explicitly instructed that correctness was not the criterion of evaluation; instead, they were asked to judge which response better served the target persona.

\subsection{Annotation difficulty and study scope}
This annotation task proved demanding. It required annotators to understand code, compare often lengthy responses, and make judgments across seven dimensions from the perspective of a user persona that was not their own.

In post-task interviews, annotators described the task as ``difficult'' and ``exhaustive.'' Several reported that, when comparisons contained especially long responses, they sometimes skimmed parts of the text and relied on heuristics to judge which answer was better. They also noted that it could be hard to assess whether a response truly satisfied persona-specific requests, especially when those requests involved technical preferences they did not fully share or understand, such as a particular complexity analysis or algorithmic explanation. 

These difficulties were especially pronounced for personalized prompts, which tend to be longer because they add user-specific requirements and often elicit longer responses. This is not a flaw of the personalized prompts themselves: as our survey and in-the-wild analysis suggest, real vibe-testing often involves long, detailed, user-specific prompts and responses. Rather, it makes external validation by human annotators substantially harder. For this reason, we kept the study small, sampled relatively few personalized items, and omit their detailed results from the main text.

Overall, the study provides a focused human validation of the automated persona-conditioned judging setup used in the main experiments. It covers two personas, two prompt types, and two model pairs, and evaluates whether human preferences follow the same pairwise comparison framework used by the automated judges. Additional reproducibility details, including configuration files, item assignment manifests, and survey generation settings, are documented in the study materials.

\subsection{Results}

\begin{table*}[t]
    \centering
    \setlength{\tabcolsep}{7pt}
    \renewcommand{\arraystretch}{1.15}
    \begin{tabular}{@{}c|c|cc|cc@{}}
        \toprule
        \textbf{Prompt Type} & \textbf{Judge Pair Type} & \textbf{Agreement (\%)} & \textbf{Cohen's $\kappa$} & \textbf{\#Pairs} & \textbf{\#Items} \\
        \midrule
            \multirow{3}{*}{\textbf{Original}} & LLM--LLM & \pmstd{90.9}{4.5} & \pmstd{0.81}{0.10} & 3 & 66 \\
        & Human--Human & \pmstd{94.4}{15.0} & \pmstd{0.80}{0.39} & 15 & 35 \\
        & Human--LLM & \pmstd{89.5}{15.6} & \pmstd{0.78}{0.33} & 18 & 141 \\
        \midrule
            \multirow{3}{*}{\textbf{Personalized}} & LLM--LLM & \pmstd{100.0}{0.0} & \pmstd{1.00}{0.00} & 3 & 30 \\
            & Human--Human & \pmstd{40.0}{43.1} & \pmstd{-0.17}{0.58} & 15 & 24 \\
            & Human--LLM & \pmstd{50.0}{21.0} & \pmstd{-0.06}{0.32} & 18 & 72 \\
        \midrule
        \bottomrule
    \end{tabular}
    \caption{Human judgment validation results across prompt types and compared grouped judge types. Agreement is reported as mean percentage agreement and Cohen's Kappa, with standard deviations across judge pairs. The reported Pairs counts judge pairs, and Items sums the per-pair overlap counts across those judge pairs.}
    \label{tab:human_validation_agreement}
\end{table*}

\begin{table}[t]
    \centering
    \setlength{\tabcolsep}{3pt}
    \renewcommand{\arraystretch}{1.12}
    \begin{tabular}{@{}c|c|cc|cc@{}}
        \toprule
        \textbf{Dimension} & \textbf{Pair Type} & \textbf{Agr. (\%)} & \textbf{$\kappa$} & \textbf{Agr. excl. ties (\%)} & \textbf{$\kappa$ excl. ties} \\
        \midrule
\multirow{4}{*}{\textbf{Clarity}} & Overall & \pmstd{54.4}{30.8} & \pmstd{0.27}{0.32} & \pmstd{73.6}{25.1} & \pmstd{0.40}{0.42} \\
 & LLM--LLM & \pmstd{69.7}{6.9} & \pmstd{0.43}{0.10} & \pmstd{88.1}{6.1} & \pmstd{0.70}{0.15} \\
 & Human--Human & \pmstd{48.7}{40.2} & \pmstd{0.20}{0.34} & \pmstd{75.0}{34.5} & \pmstd{0.34}{0.47} \\
 & Human--LLM & \pmstd{56.6}{23.2} & \pmstd{0.30}{0.33} & \pmstd{70.3}{19.1} & \pmstd{0.38}{0.42} \\
\cmidrule(lr){1-6}
\multirow{4}{*}{\textbf{Tone/Style Fit}} & Overall & \pmstd{92.3}{11.4} & \pmstd{0.81}{0.23} & \pmstd{97.6}{5.5} & \pmstd{0.92}{0.16} \\
 & LLM--LLM & \pmstd{90.9}{4.5} & \pmstd{0.78}{0.11} & \pmstd{100.0}{0.0} & \pmstd{1.00}{0.00} \\
 & Human--Human & \pmstd{97.8}{8.6} & \pmstd{0.94}{0.18} & \pmstd{100.0}{0.0} & \pmstd{1.00}{0.00} \\
 & Human--LLM & \pmstd{88.0}{12.5} & \pmstd{0.75}{0.25} & \pmstd{95.2}{7.0} & \pmstd{0.87}{0.19} \\
\cmidrule(lr){1-6}
\multirow{4}{*}{\textbf{Workflow Fit}} & Overall & \pmstd{54.1}{29.7} & \pmstd{0.24}{0.35} & \pmstd{75.0}{27.6} & \pmstd{0.47}{0.47} \\
 & LLM--LLM & \pmstd{59.1}{16.4} & \pmstd{0.32}{0.27} & \pmstd{77.8}{19.2} & \pmstd{0.53}{0.41} \\
 & Human--Human & \pmstd{55.7}{42.5} & \pmstd{0.29}{0.45} & \pmstd{75.0}{34.5} & \pmstd{0.42}{0.49} \\
 & Human--LLM & \pmstd{52.0}{17.0} & \pmstd{0.20}{0.30} & \pmstd{74.5}{24.8} & \pmstd{0.48}{0.50} \\
\cmidrule(lr){1-6}
\multirow{4}{*}{\textbf{Cognitive Load}} & Overall & \pmstd{50.0}{30.4} & \pmstd{0.18}{0.35} & \pmstd{65.9}{29.5} & \pmstd{0.28}{0.51} \\
 & LLM--LLM & \pmstd{57.6}{2.6} & \pmstd{0.33}{0.02} & \pmstd{78.2}{5.6} & \pmstd{0.55}{0.11} \\
 & Human--Human & \pmstd{42.9}{40.9} & \pmstd{0.01}{0.37} & \pmstd{58.5}{38.2} & \pmstd{0.06}{0.45} \\
 & Human--LLM & \pmstd{54.7}{21.3} & \pmstd{0.27}{0.34} & \pmstd{68.4}{25.6} & \pmstd{0.34}{0.55} \\
\cmidrule(lr){1-6}
\multirow{4}{*}{\textbf{Context Awareness}} & Overall & \pmstd{44.9}{36.2} & \pmstd{0.22}{0.34} & \pmstd{78.5}{32.7} & \pmstd{0.45}{0.56} \\
 & LLM--LLM & \pmstd{39.4}{22.9} & \pmstd{0.17}{0.22} & \pmstd{72.1}{10.9} & \pmstd{0.20}{0.35} \\
 & Human--Human & \pmstd{42.8}{46.4} & \pmstd{0.19}{0.40} & \pmstd{65.0}{47.4} & \pmstd{0.17}{0.75} \\
 & Human--LLM & \pmstd{47.6}{29.1} & \pmstd{0.24}{0.33} & \pmstd{89.5}{16.7} & \pmstd{0.70}{0.37} \\
\cmidrule(lr){1-6}
\multirow{4}{*}{\textbf{Persona Consistency}} & Overall & \pmstd{64.1}{36.9} & \pmstd{0.43}{0.45} & \pmstd{95.1}{18.6} & \pmstd{0.90}{0.25} \\
 & LLM--LLM & \pmstd{53.0}{21.5} & \pmstd{0.29}{0.26} & \pmstd{94.7}{9.1} & \pmstd{0.88}{0.20} \\
 & Human--Human & \pmstd{67.0}{46.1} & \pmstd{0.43}{0.55} & \pmstd{100.0}{0.0} & \pmstd{1.00}{0.00} \\
 & Human--LLM & \pmstd{63.5}{31.1} & \pmstd{0.46}{0.43} & \pmstd{92.2}{24.4} & \pmstd{0.86}{0.30} \\
\cmidrule(lr){1-6}
\multirow{4}{*}{\textbf{Anthropomorphism}} & Overall & \pmstd{38.8}{35.2} & \pmstd{0.05}{0.13} & \pmstd{59.7}{34.9} & \pmstd{0.10}{0.17} \\
 & LLM--LLM & \pmstd{45.5}{25.3} & \pmstd{0.18}{0.21} & \pmstd{62.7}{26.3} & \pmstd{0.27}{0.24} \\
 & Human--Human & \pmstd{26.7}{41.7} & \pmstd{0.03}{0.10} & \pmstd{56.2}{49.6} & \pmstd{0.00}{0.00} \\
 & Human--LLM & \pmstd{47.8}{28.7} & \pmstd{0.03}{0.13} & \pmstd{60.7}{30.0} & \pmstd{0.09}{0.16} \\
\midrule
\midrule
\multirow{4}{*}{\textbf{Pooled}} & Overall & \pmstd{56.9}{14.2} & \pmstd{0.27}{0.23} & \pmstd{79.4}{14.7} & \pmstd{0.51}{0.33} \\
 & LLM--LLM & \pmstd{59.3}{7.4} & \pmstd{0.34}{0.09} & \pmstd{83.1}{3.6} & \pmstd{0.64}{0.08} \\
 & Human--Human & \pmstd{54.5}{17.3} & \pmstd{0.16}{0.27} & \pmstd{78.9}{19.9} & \pmstd{0.39}{0.45} \\
 & Human--LLM & \pmstd{58.6}{12.5} & \pmstd{0.34}{0.17} & \pmstd{79.2}{10.5} & \pmstd{0.57}{0.23} \\
        \bottomrule
    \end{tabular}
    \caption{\textbf{Dimension-level human judgment validation results on the original prompts.} Agreement is reported as mean percentage agreement and Cohen's $\kappa$, with standard deviations across judge pairs. Pooled counts treat each sample-dimension pair as one item. Excluding ties (\texttt{excl. tie}) removes items marked as ties by either judge, since ties do not affect which model wins.}
    \label{tab:human_validation_dimension_agreement_original}
\end{table}

\begin{table}[t]
    \centering
    \setlength{\tabcolsep}{3pt}
    \renewcommand{\arraystretch}{1.12}
    \begin{tabular}{@{}c|c|cc|cc@{}}
    \toprule
    \textbf{Dimension} & \textbf{Pair Type} & \textbf{Agr. (\%)} & \textbf{$\kappa$} & \textbf{Agr. excl. ties (\%)} & \textbf{$\kappa$ excl. ties} \\
    \midrule
\multirow{4}{*}{\textbf{Clarity}} & Overall & \pmstd{28.5}{32.7} & \pmstd{-0.04}{0.35} & \pmstd{47.6}{38.4} & \pmstd{-0.13}{0.64} \\
 & LLM--LLM & \pmstd{83.3}{5.8} & \pmstd{0.71}{0.08} & \pmstd{96.7}{5.8} & \pmstd{0.93}{0.12} \\
 & Human--Human & \pmstd{20.0}{36.8} & \pmstd{-0.13}{0.32} & \pmstd{35.7}{47.6} & \pmstd{-0.20}{0.45} \\
 & Human--LLM & \pmstd{26.4}{21.8} & \pmstd{-0.11}{0.22} & \pmstd{43.3}{30.7} & \pmstd{-0.33}{0.53} \\
\cmidrule(lr){1-6}
\multirow{4}{*}{\textbf{Tone/Style Fit}} & Overall & \pmstd{39.7}{35.3} & \pmstd{0.08}{0.41} & \pmstd{69.6}{38.7} & \pmstd{0.28}{0.54} \\
 & LLM--LLM & \pmstd{76.7}{11.5} & \pmstd{0.64}{0.18} & \pmstd{95.2}{8.2} & \pmstd{0.91}{0.16} \\
 & Human--Human & \pmstd{23.3}{37.2} & \pmstd{-0.10}{0.32} & \pmstd{43.8}{49.6} & \pmstd{0.00}{0.00} \\
 & Human--LLM & \pmstd{47.2}{29.6} & \pmstd{0.12}{0.41} & \pmstd{78.3}{28.8} & \pmstd{0.22}{0.62} \\
\cmidrule(lr){1-6}
\multirow{4}{*}{\textbf{Workflow Fit}} & Overall & \pmstd{34.4}{31.9} & \pmstd{0.05}{0.41} & \pmstd{52.6}{42.2} & \pmstd{0.13}{0.61} \\
 & LLM--LLM & \pmstd{63.3}{5.8} & \pmstd{0.45}{0.05} & \pmstd{95.8}{7.2} & \pmstd{0.92}{0.14} \\
 & Human--Human & \pmstd{16.7}{36.2} & \pmstd{-0.14}{0.48} & \pmstd{27.8}{44.1} & \pmstd{-0.14}{0.38} \\
 & Human--LLM & \pmstd{44.4}{22.0} & \pmstd{0.13}{0.31} & \pmstd{57.9}{37.9} & \pmstd{0.10}{0.63} \\
\cmidrule(lr){1-6}
\multirow{4}{*}{\textbf{Cognitive Load}} & Overall & \pmstd{34.2}{27.5} & \pmstd{0.00}{0.30} & \pmstd{62.9}{34.4} & \pmstd{0.21}{0.46} \\
 & LLM--LLM & \pmstd{60.0}{10.0} & \pmstd{0.39}{0.15} & \pmstd{89.7}{9.0} & \pmstd{0.79}{0.18} \\
 & Human--Human & \pmstd{36.7}{35.2} & \pmstd{0.03}{0.21} & \pmstd{66.7}{43.3} & \pmstd{0.00}{0.00} \\
 & Human--LLM & \pmstd{27.8}{19.0} & \pmstd{-0.08}{0.32} & \pmstd{55.7}{30.2} & \pmstd{0.15}{0.48} \\
\cmidrule(lr){1-6}
\multirow{4}{*}{\textbf{Context Awareness}} & Overall & \pmstd{35.1}{31.2} & \pmstd{0.05}{0.37} & \pmstd{60.6}{41.8} & \pmstd{0.18}{0.55} \\
 & LLM--LLM & \pmstd{80.0}{0.0} & \pmstd{0.66}{0.00} & \pmstd{100.0}{0.0} & \pmstd{1.00}{0.00} \\
 & Human--Human & \pmstd{20.0}{31.6} & \pmstd{-0.10}{0.30} & \pmstd{44.4}{52.7} & \pmstd{-0.20}{0.45} \\
 & Human--LLM & \pmstd{40.3}{24.5} & \pmstd{0.05}{0.34} & \pmstd{62.3}{35.1} & \pmstd{0.14}{0.46} \\
\cmidrule(lr){1-6}
\multirow{4}{*}{\textbf{Persona Consistency}} & Overall & \pmstd{38.6}{31.4} & \pmstd{0.05}{0.31} & \pmstd{69.6}{40.1} & \pmstd{0.14}{0.69} \\
 & LLM--LLM & \pmstd{46.7}{15.3} & \pmstd{0.20}{0.22} & \pmstd{86.1}{12.7} & \pmstd{0.72}{0.25} \\
 & Human--Human & \pmstd{36.7}{39.9} & \pmstd{-0.08}{0.32} & \pmstd{33.3}{57.7} & \pmstd{-0.50}{0.71} \\
 & Human--LLM & \pmstd{38.9}{26.0} & \pmstd{0.12}{0.30} & \pmstd{73.8}{37.8} & \pmstd{0.07}{0.67} \\
\cmidrule(lr){1-6}
\multirow{4}{*}{\textbf{Anthropomorphism}} & Overall & \pmstd{33.6}{32.9} & \pmstd{0.01}{0.27} & \pmstd{47.0}{36.8} & \pmstd{-0.02}{0.33} \\
 & LLM--LLM & \pmstd{70.0}{10.0} & \pmstd{0.53}{0.15} & \pmstd{76.8}{9.3} & \pmstd{0.53}{0.19} \\
 & Human--Human & \pmstd{40.0}{43.1} & \pmstd{0.06}{0.13} & \pmstd{68.8}{45.8} & \pmstd{0.00}{0.00} \\
 & Human--LLM & \pmstd{22.2}{16.9} & \pmstd{-0.11}{0.24} & \pmstd{29.4}{23.8} & \pmstd{-0.13}{0.27} \\
\midrule
\midrule
\multirow{4}{*}{\textbf{Pooled}} & Overall & \pmstd{34.9}{16.2} & \pmstd{0.03}{0.21} & \pmstd{61.3}{26.4} & \pmstd{0.09}{0.39} \\
 & LLM--LLM & \pmstd{68.6}{1.4} & \pmstd{0.51}{0.03} & \pmstd{91.7}{2.1} & \pmstd{0.83}{0.04} \\
 & Human--Human & \pmstd{27.6}{14.3} & \pmstd{-0.05}{0.14} & \pmstd{58.0}{34.0} & \pmstd{-0.16}{0.23} \\
 & Human--LLM & \pmstd{35.3}{11.1} & \pmstd{0.01}{0.16} & \pmstd{58.9}{18.1} & \pmstd{0.10}{0.33} \\
\bottomrule
    \end{tabular}
    \caption{\textbf{Dimension-level human judgment validation results on personalized prompts.} Agreement is reported as mean percentage agreement and Cohen's $\kappa$, with standard deviations across judge pairs. Pooled counts treat each sample-dimension pair as one item. Excluding ties (\texttt{excl. tie}) removes items marked as ties by either judge, since ties do not affect which model wins.}
    \label{tab:human_validation_dimension_agreement_personalized}
\end{table}

Table~\ref{tab:human_validation_agreement} summarizes agreement on the overall preference labels.
On \textbf{original prompts}, agreement is high across all judge-pair types: LLM--LLM agreement is ($90.9\% \pm 4.5$) with ($\kappa=0.81 \pm 0.10$), Human--Human agreement is ($94.4\% \pm 15.0$) with ($\kappa=0.80\pm0.39$), and Human--LLM agreement is ($89.5\%\pm15.6$) with ($\kappa=0.78\pm0.33$).
This indicates that, in the simpler original-prompt setting, automated persona-conditioned judgments align closely with human overall preferences.
By contrast, agreement drops sharply on \textbf{personalized prompts}: while LLM--LLM agreement remains perfect on this small subset (note these samples were pre-sampled with high consensus), ($100.0\%, \kappa=1.00$), Human--Human and Human--LLM agreement fall to ($40.0\%\pm43.1$) and ($50.0\%\pm21.0$), respectively, with near-zero or negative ($\kappa$). Given the small number of personalized items and the much greater annotation difficulty discussed above, we view these personalized-prompt results as noisy and inconclusive rather than as evidence against the automated setup. 

\paragraph{Dimension-level agreement}
Tables~\ref{tab:human_validation_dimension_agreement_original} and~\ref{tab:human_validation_dimension_agreement_personalized} break the agreement down by evaluation dimension. On \textbf{original prompts}, the strongest agreement appears on \emph{Tone/Style Fit}, where pooled agreement reaches ($92.3\%\pm11.4$) ($\kappa=0.81\pm0.23$), and on \emph{Persona Consistency} after excluding ties, where pooled agreement reaches ($95.1\%\pm18.6$) ($\kappa=0.90\pm0.25$). Several other dimensions show moderate agreement once ties are excluded, including \emph{Clarity} ($73.6\%\pm25.1$), \emph{Workflow Fit} ($75.0\%\pm27.6$), and \emph{Context Awareness} ($78.5\%\pm32.7$). In contrast, \emph{Anthropomorphism} is the least reliable dimension, with low pooled agreement ($38.8\%\pm35.2$), ($\kappa=0.05\pm0.13$). Overall, pooling all original-prompt sample-dimension decisions yields ($56.9\%\pm14.2$) agreement and ($\kappa=0.27\pm0.23$), rising to ($79.4\%\pm14.7$) and ($\kappa=0.51\pm0.33$) when ties are excluded. This suggests that much of the disagreement on original prompts comes from tie decisions rather than direct winner conflicts. 

On \textbf{personalized prompts}, dimension-level agreement is substantially lower and much more variable. The pooled score across all dimensions is only ($34.9\%\pm16.2$) ($\kappa=0.03\pm0.21$), rising to ($61.3\%\pm26.4$) ($\kappa=0.09\pm0.39$) after excluding ties. LLM--LLM agreement remains noticeably higher than Human--Human or Human--LLM agreement, with pooled tie-excluded agreement of ($91.7\%\pm2.1$) for LLM--LLM versus ($58.0\%\pm34.0$) for Human--Human and ($58.9\%\pm18.1$) for Human--LLM. Among dimensions, \emph{Tone/Style Fit} and \emph{Anthropomorphism} are relatively more stable after excluding ties, while dimensions such as \emph{Workflow Fit}, \emph{Context Awareness}, and \emph{Persona Consistency} remain highly inconsistent. Together with the annotator feedback, these results suggest that personalized prompts are harder to validate externally: they often require reading longer responses, tracking more user-specific constraints, and judging from the perspective of a user whose preferences the annotator may not share. 

Overall, the human validation results mostly support the automated judging setup in the original-prompt setting, where human and LLM judgments align closely at the overall level and reasonably well across several key dimensions.
The personalized-prompt results are too small and inconsistent to validate the corresponding automated judgments, and annotation difficulty alone cannot rule out systematic judge bias. Stronger validation will require larger samples, annotators matched more closely to the target profiles, and annotation protocols that reduce response length and cognitive burden.

\section{Practitioner's Guide}
\label{appendix:practitioner_guide}

\begin{table*}[t]
\centering

\begin{tcolorbox}[
  width=\textwidth,
  colback=checkGreenLight,
  colframe=checkGreenLine,
  colbacktitle=checkGreenDark,
  coltitle=white,
  fonttitle=\bfseries\small,
  title={\strut Vibe-Testing Evaluation Checklist},
  boxrule=0.5pt,
  arc=2pt,
  top=5pt,
  bottom=5pt,
  left=5pt,
  right=5pt
]

\footnotesize
\setlength{\tabcolsep}{5pt}
\renewcommand{\arraystretch}{1.28}

\begin{adjustbox}{max width=\linewidth}
\begin{tabular}{
  @{}
  L{3.8cm}
  |
  L{4.2cm}
  |
  L{3.6cm}
  |
  L{5.3cm}
  @{}
}
\toprule

\textcolor{checkGreenDark}{\textbf{Stage}} &
\textcolor{checkGreenDark}{\textbf{Required Decision}} &
\textcolor{checkGreenDark}{\textbf{Framework Elements}} &
\textcolor{checkGreenDark}{\textbf{Validation and Reporting}} \\

\midrule

\textbf{\textcolor{checkGreenDark}{1. Target definition}}
&
Define the target user, group, or community and the intended use of the evaluation.
&
User evidence, intended use, and evaluation scope.
&
Report the source of the user information, the target population, and the scope of the resulting profile. \\

\addlinespace[2pt]

\midrule

\textbf{\textcolor{checkGreenDark}{2. Profile construction}}
&
Convert the available user information into explicit input and output preferences.
&
Input dimensions, output dimensions, and optional importance weights.
&
Leave unsupported preferences unspecified and manually review automatically generated profiles. \\

\addlinespace[2pt]

\midrule

\textbf{\textcolor{checkGreenDark}{3. Task selection}}
&
Select tasks that reflect the target users' expected needs, goals, and workflows.
&
Task type, complexity, real-world context, constraints, and reference material.
&
Describe the domain and task coverage, including important settings that are not represented. \\

\addlinespace[2pt]

\midrule

\textbf{\textcolor{checkGreenDark}{4. Prompt personalization}}
&
Apply profile-relevant changes to each task prompt.
&
Context, persona framing, specificity, constraints, underspecification, and requested response style.
&
Verify task preservation and report the prompt conditions, generation procedure, and controls. \\

\addlinespace[2pt]

\midrule

\textbf{\textcolor{checkGreenDark}{5. Model generation}}
&
Run candidate models under a consistent comparison setup.
&
Prompt condition, model pair, number of generations, and comparison format.
&
Report model versions, system prompts, decoding settings, and generations per prompt. \\

\addlinespace[2pt]

\midrule

\textbf{\textcolor{checkGreenDark}{6. Correctness evaluation}}
&
Apply objective or task-specific checks when reliable measures are available.
&
Correctness, constraint satisfaction, or task success.
&
Describe the tests, report their coverage, and identify relevant qualities they do not measure. \\

\addlinespace[2pt]

\midrule

\textbf{\textcolor{checkGreenDark}{7. Subjective judging}}
&
Compare responses from the perspective of the target profile.
&
Selected output dimensions, profile preferences, and dimension-specific criteria.
&
Define each dimension clearly and mitigate position effects, judge bias, and model identity effects. \\

\addlinespace[2pt]

\midrule

\textbf{\textcolor{checkGreenDark}{8. Aggregation}}
&
Combine dimension-level judgments into user-level or community-level summaries.
&
Dimension weights, tie handling, comparison rules, and profile aggregation.
&
Report the aggregation rule, tie rates, profile weights, and variation across users or groups. \\

\addlinespace[2pt]

\midrule

\textbf{\textcolor{checkGreenDark}{9. Validation}}
&
Assess the reliability of the profiles, prompt transformations, judgments, and conclusions.
&
All pipeline components used in the evaluation.
&
Include appropriate human validation, agreement analysis, ablations, robustness checks, or manual review. \\

\bottomrule
\end{tabular}
\end{adjustbox}

\end{tcolorbox}

\caption{\textbf{Evaluation checklist for applying the vibe-testing framework.}
Each stage specifies a practical decision, the relevant framework elements, and the minimum validation or reporting needed to construct a clear and reproducible user-conditioned evaluation.}\label{tab:vibe_testing_checklist}
\end{table*}

The framework can be used as an evaluation scaffold for adapting existing tasks to a target user or community. The taxonomy provides a set of input dimensions for deciding what to test and output dimensions for deciding how responses should be judged. Practitioners can select the dimensions relevant to their setting, define their values from available user information, and implement each pipeline component with methods that fit their data and evaluation goals. The evaluation checklist shown in Table~\ref{tab:vibe_testing_checklist} summarizes the main decisions and validation steps involved.

\paragraph{Defining the target user or community.}
A target profile can be constructed from a direct description, survey responses, public posts, interviews, support requests, or other user data. The source should provide evidence about the user's tasks, context, constraints, and response preferences. For example, a developer evaluating coding assistants for beginner Python students could use course surveys and forum questions to identify common needs, such as simple explanations, limited use of advanced syntax, and step-by-step debugging guidance. These observations can be mapped to input dimensions that shape the prompts and output dimensions such as clarity, cognitive load, and workflow fit. Multiple sources can be combined to represent a broader community, while individual profiles can be retained when the goal is to study variation between users. Profiles should be manually reviewed before use, especially when they are generated automatically from unstructured text.

\paragraph{Adapting an existing evaluation set.}
Practitioners can begin with an existing benchmark, internal task collection, or set of real user requests. For each task, they first identify which input dimensions are relevant to the target profile. They can then generate one or more prompt variants that add the corresponding context, constraints, or requested response style while preserving the underlying task. In the beginner Python example, a benchmark prompt asking for a function may be rewritten to request a simple implementation, avoid unfamiliar language features, and include a short explanation. Semantic-preservation checks, manual review, or executable tests can help verify that the rewritten prompt still evaluates the intended capability. Neutral paraphrases can also be included to separate persona-specific effects from ordinary changes in wording.

\paragraph{Evaluating model responses.}
Models are run on the original and personalized task variants, and their responses are compared using the output dimensions selected for the target profile. When executable tests are available, correctness can be reported alongside the subjective judgments. Pairwise evaluation is often useful because it asks which response better serves the target user on each dimension. Practitioners can use human annotators, LLM judges, or a combination of both, depending on the task and available resources. The evaluation rubric should define each dimension clearly and explain how it applies to the target profile.

\paragraph{Inspecting and aggregating results.}
Results can first be inspected at the dimension level. This analysis can reveal, for example, that one model produces clearer explanations while another follows constraints more reliably. Dimension-level judgments can then be combined using importance weights from the user profile to produce an overall preference score. For a community-level analysis, practitioners can aggregate results across profiles, report the distribution of preferences, or group users with similar priorities. These summaries should preserve meaningful variation between users. Two users may assign different importance to clarity and concision, or may interpret workflow fit differently, leading to different model preferences. The resulting evaluation therefore supports user-conditioned comparisons and does not define a universal model ranking.

\paragraph{Open-ended tasks.}
The same procedure can be applied when automatic correctness metrics are unavailable. For tasks such as visual design, SVG generation, or dynamic webpage creation, practitioners can replace executable tests with task-specific rubrics, human review, constraint checks, or pairwise preference judgments. For example, a design team could compare generated webpages for accessibility, visual hierarchy, adherence to a brand style, and ease of modification. Some properties may still support automatic checks, such as valid HTML or accessibility rules, while subjective qualities can be evaluated by users or trained annotators. Reporting the dimensions separately helps distinguish disagreements about style and usability from failures to satisfy explicit requirements.

In practice, the framework is most useful when the target users, task collection, and evaluation dimensions are defined together. The taxonomy helps make these choices explicit, the pipeline applies them consistently, and the resulting analysis shows how model preferences depend on the users and settings being evaluated.


\begin{table}[t]
\centering
\small
\begin{tabular}{r p{2.2cm} p{10.6cm}}
\toprule
\# & Source & Title \\
\midrule
1 & Youtube & Google’s nano banana just killed Photoshop... let’s run it \\
2 & Youtube & GPT-5 is here... Can it win back programmers? \\
3 & Youtube & Google’s Genie model makes realistic worlds in realtime… \\
4 & Youtube & Is Elon’s Grok 3 the new AI king? \\
5 & Youtube & I Tested Gemini 3 Pro (it did really well) \\
6 & Youtube & Claude 4 vs GPT-5: Which AI is actually better? \\
7 & Youtube & Claude vs GPT-5: The REAL Winner (Coding Test) \\
8 & Youtube & Can AI Really Write Good Code? (GPT-5 vs Claude) \\
9 & Youtube & How Well Do AI Models Actually Work? (Claude, GPT-5, Gemini) \\
10 & Youtube & I Tried GPT-5 For A Week... Here’s What Happened \\
11 & Youtube & Claude 4 is here: first impressions and testing \\
12 & Youtube & Gemini 3 Pro vs GPT-5: Real World Tests \\
13 & Youtube & Best AI for writing in 2025? (Claude vs GPT) \\
14 & Youtube & The AI model that finally feels human (testing) \\
15 & Youtube & I compared AI chatbots for daily work (what surprised me) \\
16 & Youtube & I Tested the BEST AI Coding Assistants (Shocking Results) \\
17 & Youtube & GPT-5 vs Claude 4: 10 Real Use Cases \\
18 & Youtube & Claude 4 Is Better Than GPT-5 (Here’s Why) \\
19 & Youtube & Gemini 3 Pro vs Claude 4: Which is better? \\
20 & Youtube & I tested AI agents: what actually works \\
\bottomrule
\end{tabular}
\caption{Full list of in-the-wild sources used to construct the 40-report corpus (Part 1/2). URLs and additional source metadata are provided in the supplied CSV .}
\label{tab:vibe_wild_sources_a}
\end{table}

\begin{table}[t]
\centering
\small
\begin{tabular}{r p{2.2cm} p{10.6cm}}
\toprule
\# & Source & Title \\
\midrule
21 & Reddit & GPT-5 vs Claude 4 for coding: which feels better? \\
22 & Reddit & GPT-4o feels different, anyone else? \\
23 & Reddit & What model do you prefer for writing and why? \\
24 & Reddit & Benchmarks say X, but I keep using Y (reasons) \\
25 & Reddit & Llama 3 vs Claude: vibes and reliability \\
26 & Reddit & Which model has better tone for tutoring? \\
27 & Reddit & Claude feels safer but more constrained, thoughts? \\
28 & Reddit & What is the least frustrating model for daily work? \\
29 & Reddit & Which model do you trust most and why? \\
30 & Reddit & Why does this model “feel smarter” than the leaderboard? \\
31 & Blog / News & Choosing an LLM: what benchmarks miss in practice \\
32 & Blog / News & Real-world LLM eval: clarity, tone, and friction \\
33 & Blog / News & I switched from GPT to Claude for my workflow \\
34 & Blog / News & The best model is the one that fits your brain \\
35 & Blog / News & AI leaderboards cannot predict user satisfaction \\
36 & Blog / News & LLM adoption is driven by “feel,” not scores \\
37 & Blog / News & Why some models “feel smarter” despite lower benchmarks \\
38 & Blog / News & Prompting for vibe: how I judge assistants \\
39 & Blog / News & The model that feels most reliable for production use \\
40 & Blog / News & DeepSeek vs. ChatGPT vs. Qwen 2.5: Here's the winner \\
\bottomrule
\end{tabular}
\caption{Full list of in-the-wild sources used to construct the 40-report corpus (Part 2/2). URLs and additional source metadata are provided in the supplied CSV.}
\label{tab:vibe_wild_sources_b}
\end{table}

\newpage

\begin{table}
\centering
\setlength{\tabcolsep}{7pt}
\renewcommand{\arraystretch}{1.18}
\begin{tabular}{@{}p{0.62\textwidth}p{0.34\textwidth}@{}}
\toprule
\textbf{Question  } & \textbf{Responses (\%)} \\
\midrule

\textbf{\textit{AI tool usage frequency.}}
How often do you use AI tools (e.g., ChatGPT, Claude, Gemini, Copilot)? &
\textbf{Daily:} 92.2\% \newline
\textbf{Several times per week:} 7.8\% \\

\addlinespace[2pt]\midrule\addlinespace[2pt]

\textbf{\textit{Primary use cases.}}
What do you mainly use AI systems for? \textit{(multi-select)} &
\textbf{Coding / debugging:} 88.2\% \newline
\textbf{Writing / communication:} 70.6\% \newline
\textbf{Learning / tutoring:} 56.9\% \newline
\textbf{Complex reasoning / problem-solving:} 37.3\% \newline
\textbf{Productivity:} 35.3\% \newline
\textbf{Creative tasks:} 33.3\% \newline
\textbf{Entertainment:} 23.5\% \newline
\textbf{Other (free-text):} 6.0\% \\

\addlinespace[2pt]\midrule\addlinespace[2pt]

\textbf{\textit{Technical background.}}
How would you describe your technical background? &
\textbf{Technical (e.g., engineering, data science, CS background):} 47.1\% \newline
\textbf{AI/ML expert:} 47.1\% \newline
\textbf{Somewhat technical:} 5.9\% \\

\bottomrule
\end{tabular}
\caption{Extended survey questions and results (Part A: \textbf{Usage and Background}). Percentages are computed over respondents who answered each question.}
\label{tab:survey_part_a}
\end{table}

\begin{table*}[t]
\centering
\setlength{\tabcolsep}{7pt}
\renewcommand{\arraystretch}{1.18}
\begin{tabular}{@{}p{0.62\textwidth}p{0.34\textwidth}@{}}
\toprule
\textbf{Question  } & \textbf{Responses (\%)} \\
\midrule

\textbf{\textit{Vibe-testing familiarity.}}
Have you ever vibe-tested an AI model? &
\textbf{Yes:} 82.4\% \newline
\textbf{Not sure:} 13.7\% \newline
\textbf{No:} 3.9\% \\

\addlinespace[2pt]\midrule\addlinespace[2pt]

\textbf{\textit{Exploration frequency.}}
When using a new AI model, how often do you experiment with prompts or tasks just to see how it behaves? \newline
{\footnotesize (1 = never, 7 = very often)} &
\textbf{1:} 2.0\% \quad
\textbf{2:} 4.1\% \quad
\textbf{3:} 6.1\% \quad
\textbf{4:} 8.2\% \quad
\textbf{5:} 32.7\% \quad
\textbf{6:} 22.4\% \quad
\textbf{7:} 24.5\% \\

\addlinespace[2pt]\midrule\addlinespace[2pt]

\textbf{\textit{Typical vibe-testing actions.}}
Which of the following do you typically do when vibe-testing? \textit{(multi-select)} &
\textbf{Try tasks from my own workflow:} 73.5\% \newline
\textbf{Compare outputs from different models:} 65.3\% \newline
\textbf{Check tone, style, or personality:} 49.0\% \newline
\textbf{Give vague or underspecified instructions (ambiguity stress-test):} 44.9\% \newline
\textbf{Use a small set of personal ``test prompts'':} 38.8\% \newline
\textbf{Re-run the same prompt (stability / consistency):} 38.8\% \newline
\textbf{Other (free-text):} 6.0\% \\

\addlinespace[2pt]\midrule\addlinespace[2pt]

\textbf{\textit{Vibe-testing domains.}}
In which domains do you most often vibe-test AI tools? \textit{(multi-select)} &
\textbf{Coding / software development:} 65.3\% \newline
\textbf{Writing / communication:} 51.0\% \newline
\textbf{Reasoning / problem-solving:} 44.9\% \newline
\textbf{Learning / tutoring:} 28.6\% \newline
\textbf{Creative tasks:} 28.6\% \newline
\textbf{Productivity:} 14.3\% \newline
\textbf{Other (free-text):} 6.0\% \\

\addlinespace[2pt]\midrule\addlinespace[2pt]

\textbf{\textit{Judgment criteria.}}
When evaluating a model’s responses, which aspects matter most to you? \textit{(multi-select)} &
\textbf{Correctness / accuracy:} 91.8\% \newline
\textbf{Clarity and structure:} 59.2\% \newline
\textbf{Fit for my personal workflow:} 40.8\% \newline
\textbf{Tone / style:} 32.7\% \newline
\textbf{Ability to resolve ambiguity:} 18.4\% \newline
\textbf{Efficiency / conciseness:} 16.3\% \newline
\textbf{Stability / consistency:} 20.4\% \newline
\textbf{Other (free-text):} 14.0\% \\

\bottomrule
\end{tabular}
\caption{Extended survey questions and results (Part B: \textbf{Vibe-testing Practices}). Percentages are computed over respondents who answered each question; multi-select questions report the percent selecting each option.}
\label{tab:survey_part_b}
\end{table*}

\begin{table*}[t]
\centering
\setlength{\tabcolsep}{7pt}
\renewcommand{\arraystretch}{1.18}
\begin{tabular}{@{}p{0.62\textwidth}p{0.34\textwidth}@{}}
\toprule
\textbf{Question  } & \textbf{Responses (\%)} \\
\midrule

\textbf{\textit{Semi-consistent prompt set.}}
Do you have a semi-consistent set of prompts or tasks that you use when testing a new model? (Even just one) &
\textbf{Improvised:} 51.1\% \newline
\textbf{Partially consistent:} 34.7\% \newline
\textbf{Consistent set:} 10.2\% \\

\addlinespace[2pt]\midrule\addlinespace[2pt]

\textbf{\textit{Golden prompt existence.}}
Do you have a ``Golden Prompt'' (a specific test you always use to quickly judge a model’s quality)? &
\textbf{Yes:} 18.4\% \newline
\textbf{No:} 73.5\% \newline
\textbf{Not sure:} 8.2\% \\

\addlinespace[2pt]\midrule\addlinespace[2pt]

\textbf{\textit{Golden prompt content.}}
If yes, please provide your ``Golden Prompt'' here. &
\textit{Open-ended; omitted for anonymization.} \\

\addlinespace[2pt]\midrule\addlinespace[2pt]

\textbf{\textit{Example prompt/task and what to look for.}}
Please give one example of a prompt or task you might use when testing a model. Briefly say what you look for in the response. &
\textit{Open-ended; omitted for anonymization.} \\

\addlinespace[2pt]\midrule\addlinespace[2pt]

\textbf{\textit{Consent to publish prompts.}}
May we include your prompts in our public report? &
\textbf{Yes (anonymously):} 46.8\% \newline
\textbf{Yes (with acknowledgement):} 4.3\% \newline
\textbf{No:} 48.9\% \\

\bottomrule
\end{tabular}
\caption{Extended survey questions and results (Part C: \textbf{Routines, Prompts, and Consent}). Percentages are computed over respondents who answered each question. Open-ended prompt fields were omitted for anonymization.}
\label{tab:survey_part_c}
\end{table*}

\begin{table*}[t]
\centering
\setlength{\tabcolsep}{7pt}
\renewcommand{\arraystretch}{1.18}
\begin{tabular}{@{}p{0.62\textwidth}p{0.34\textwidth}@{}}
\toprule
\textbf{Question  } & \textbf{Responses (\%)} \\
\midrule

\textbf{\textit{Benchmark awareness.}}
Are you aware of the published performance scores (benchmarks) of the models you use? &
\textbf{Yes:} 46.9\% \newline
\textbf{Somewhat:} 42.9\% \newline
\textbf{No:} 10.2\% \\

\addlinespace[2pt]\midrule\addlinespace[2pt]

\textbf{\textit{Alignment with benchmarks.}}
How closely do your personal impressions of models match their published benchmark results? \newline
{\footnotesize (1 = not at all, 7 = very closely)} &
\textbf{1:} 0.0\% \quad
\textbf{2:} 13.6\% \quad
\textbf{3:} 9.1\% \quad
\textbf{4:} 27.3\% \quad
\textbf{5:} 36.4\% \quad
\textbf{6:} 13.6\% \quad
\textbf{7:} 0.0\% \\

\addlinespace[2pt]\midrule\addlinespace[2pt]

\textbf{\textit{Benchmark mismatch.}}
Have you ever found that a model \emph{felt} significantly better or worse than its reported scores suggested? &
\textbf{Yes:} 86.4\% \newline
\textbf{Not sure:} 11.4\% \newline
\textbf{No:} 2.3\% \\

\addlinespace[2pt]\midrule\addlinespace[2pt]

\textbf{\textit{What benchmarks miss.}}
Which aspects do you feel benchmarks fail to measure well? \textit{(multi-select)} &
\textbf{Fit for my workflow:} 61.4\% \newline
\textbf{Style, tone, or personality:} 40.9\% \newline
\textbf{Handling ambiguity / underspecification:} 34.1\% \newline
\textbf{Stability and consistency:} 31.8\% \newline
\textbf{Trustworthiness / safety:} 27.3\% \newline
\textbf{Clarity and readability:} 27.3\% \newline
\textbf{Other:} 13.6\% \\

\addlinespace[2pt]\midrule\addlinespace[2pt]

\textbf{\textit{Value of vibe-testing.}}
How valuable is vibe-testing for evaluating AI systems? \newline
{\footnotesize (1 = not valuable, 7 = very valuable)} &
\textbf{1:} 0.0\% \quad
\textbf{2:} 0.0\% \quad
\textbf{3:} 2.0\% \quad
\textbf{4:} 22.4\% \quad
\textbf{5:} 38.8\% \quad
\textbf{6:} 18.4\% \quad
\textbf{7:} 18.4\% \\

\addlinespace[2pt]\midrule\addlinespace[2pt]

\textbf{\textit{Interest in automation.}}
If there were an automated tool that produced a personalized vibe-test for you, would you use it? &
\textbf{Yes:} 32.7\% \newline
\textbf{Maybe:} 51.0\% \newline
\textbf{No:} 8.2\% \newline
\textbf{Other (free-text):} 8.0\% \\

\bottomrule
\end{tabular}
\caption{Extended survey questions and results (Part D: \textbf{Benchmarks, Gaps, Value, and Automation}). Percentages are computed over respondents who answered each question; multi-select questions report the percent selecting each option.}
\label{tab:survey_part_d}
\end{table*}

\setlength{\LTpre}{6pt}
\setlength{\LTpost}{6pt}
\renewcommand{\arraystretch}{1.08}

\newpage

\end{document}